\documentclass[print,bleed,faculty=firw,department=elt,phddegree=elt]{adsphd}

\title{Multi-Task Learning for Visual Scene Understanding}

\author{Simon}{Vandenhende}

\supervisor{Prof.~Luc~Van~Gool}{}
\president{Prof. Paul Sas}
\jury{Prof.~Tinne~Tuytelaars\\
	  Prof.~Marie-Francine Moens\\
	  Prof.~Matthew~Blaschko\\
}
\externaljurymember{Prof.~Hakan~Bilen}{University of Edinburgh}
\externaljurymember{Prof.~Gemma~Roig}{Goethe University}

\researchgroup{ESAT - PSI}
\website{https://github.com/SimonVandenhende} 
\email{simon.vandenhende@esat.kuleuven.be} 

\address{Kasteelpark Arenberg 10 - box 2441}
\date{March 2022}
\copyyear{2022}

\usepackage{nomencl}   
\renewcommand{\nomname}{List of Symbols}

\makenomenclature%
\usepackage{amsmath}
\usepackage{amssymb}
\usepackage{algorithm}
\usepackage{algpseudocode}
\usepackage{booktabs}
\usepackage{color}
\usepackage{caption}
\usepackage{epsfig}
\usepackage{etoolbox}
\usepackage[inline]{enumitem}
\usepackage[linguistics]{forest}
\usepackage{float}
\usepackage{graphicx}
\usepackage{makecell}
\usepackage{mathtools}
\usepackage{multirow}
\usepackage{pdflscape}
\usepackage{pifont}   
\usepackage{physics}
\usepackage{tabularx}
\usepackage{textcomp} 
\usepackage{times}
\usepackage{standalone}
\usepackage{subcaption}
\usepackage[dvipsnames]{xcolor}
\usepackage{xspace}

\begin{document}

\makefrontcoverXII

\maketitleX

\frontmatter 

\includepreface{preface}
\includeabstract{abstract}
\includeabstractnl{abstractnl}

\tableofcontents
\listoffigures
\listoftables

\mainmatter 

\makeatletter
\AfterEndEnvironment{algorithm}{\let\@algcomment\relax}
\AtEndEnvironment{algorithm}{\kern2pt\hrule\relax\vskip3pt\@algcomment}
\let\@algcomment\relax
\newcommand\algcomment[1]{\def\@algcomment{\footnotesize#1}}
\renewcommand\fs@ruled{\def\@fs@cfont{\bfseries}\let\@fs@capt\floatc@ruled
  \def\@fs@pre{\hrule height.8pt depth0pt \kern2pt}%
  \def\@fs@post{}%
  \def\@fs@mid{\kern2pt\hrule\kern2pt}%
  \let\@fs@iftopcapt\iftrue}
\makeatother

\newcolumntype{x}[1]{>{\centering\arraybackslash}p{#1pt}}
\newcolumntype{y}[1]{>{\raggedright\arraybackslash}p{#1pt}}
\newcolumntype{z}[1]{>{\raggedleft\arraybackslash}p{#1pt}}
\newlength\savewidth\newcommand\shline{\noalign{\global\savewidth\arrayrulewidth  \global\arrayrulewidth 1pt}\hline\noalign{\global\arrayrulewidth\savewidth}}
\newcommand{\tablestyle}[2]{\setlength{\tabcolsep}{#1}\renewcommand{\arraystretch}{#2}\centering\footnotesize}

\definecolor{Highlight}{HTML}{39b54a}  
\definecolor{green}{HTML}{39b54a} 
\definecolor{red}{HTML}{cb4335} 

\newcommand{\xmark}{\ding{55}}

\colorlet{shadecolor}{gray!20}

\instructionschapters\cleardoublepage
  \graphicspath{{\chaptersdir/introduction/image/}}%
\chapter{Introduction}
\label{ch: introduction}
Machine learning has become a ubiquitous technology - it's applications range from making recommendations on e-commerce websites to recognizing crop types in agriculture. Machine learning algorithms build models based on training examples, in order to make predictions or decisions for new samples, avoiding the need to program the desired behavior explicitly. Many applications that use machine learning rely on a class of techniques named deep learning. 

Deep learning~\cite{goodfellow2016deep} uses models that consist of multiple processing layers to learn representations of data with multiple levels of abstraction. The representations automatically discover relevant structures in the data through the backpropagation algorithm, which tells how the internal parameters of the representations should be adjusted. This setup diverges from prior attempts that involved the use of handcrafted features. Deep learning has been successfully applied in text processing~\cite{pennington2014glove}, speech recognition~\cite{amodei2016deep} and visual object recognition~\cite{he2016deep}.

In computer vision, deep neural networks have reported state-of-the-art-results for semantic segmentation~\cite{long2015fully}, instance segmentation~\cite{he2017mask} and depth estimation~\cite{eigen2014depth}. Traditionally, these tasks are tackled in isolation, meaning a separate neural network is trained for each task. Yet, many real-world problems require to solve a multitude of tasks concurrently. For example, an autonomous vehicle should be able to detect all objects in the scene, localize them, estimate their distance and trajectory, etc., in order to safely navigate itself in it's surroundings. Similarly, an image recognition system for e-commerce applications should be able to recognize products, retrieve similar items, provide personalized content, etc., to provide customers with the best possible service. These problems have motivated researchers to develop multi-task learning models that jointly infer all desired task outputs. 

There is a biological argument for building multi-task learning models as well. Humans are remarkably good at solving many different tasks. It appears that biological data processing follows a multi-task learning strategy too: instead of separating tasks and tackling them in isolation, different processes seem to share the same early processing layers in the brain (e.g. V1 in macaques~\cite{gur2007direction}).

In the deep learning era, multi-task learning~\cite{caruana1997multitask,vandenhende2021multi} (MTL) translates to designing networks capable of learning shared representations from multi-task supervisory signals. Compared to the single-task case, where each individual task is solved separately by its own networks, such multi-task networks bring several advantages to the table. First, due to their inherent layer sharing, the resulting memory footprint is substantially reduced. Second, as they explicitly avoid to repeatedly calculate features in the shared layers, once for every task, they show increased inference speeds. Third, they have the potential for improved performance if the associated tasks share complementary information, or act as a regularizer for one another. 

In this thesis, we study the problem of multi-task learning within the context of visual scene understanding. We develop new techniques - including better model architectures and training recipes - that improve over the current state-of-the-art for jointly tackling multiple visual recognition tasks. The proposed methods have the potential to directly benefit the many practical applications that rely on models for visual scene understanding. 

\section{Challenges}
We identify the most important challenges for jointly learning multiple visual recognition tasks that we tackle in this thesis. 

\paragraph{Network Architecture} We need to develop novel neural network architectures that are capable of handling multiple tasks. Early attempts used either hard or soft parameter sharing for learning multiple tasks. In hard parameter sharing~\cite{kendall2018multi,kokkinos2017ubernet}, the input is first encoded through a shared network of layers after which tasks branch out into their own sequence of task-specific layers. Differently, in soft parameter sharing~\cite{misra2016cross,gao2019nddr}, a set of task-specific networks is used in combination with a feature sharing mechanism. For example, Misra et al.~\cite{misra2016cross} proposed to use cross-stitch units that take a linear combination of the features in the intermediate layers of the network to fuse information.  

The aforementioned works followed a common pattern: they directly predict all task outputs from the same input in one processing cycle. In particular, the predictions are generated once, in parallel or sequentially, and are not refined afterwards. By doing so, these works fail to capture commonalities and differences among tasks, that are likely fruitful for one another (e.g. depth discontinuities are usually aligned with semantic edges). To alleviate this issue, a few recent works~\cite{xu2018pad,zhang2018joint} first employed a multi-task network to make initial task predictions, and then leveraged features from these initial predictions in order to further improve each task output - in a one-off or recursive manner. 

These different methods each have their own advantages and disadvantages. Therefore, the main challenge will be to find elegant solutions that address the shortcomings of existing methods. Also, note that there are several aspects that need to be taken into account. We do not wish to optimize for performance alone, but would also like to consider the memory footprint and the number of computations that are needed. Often, there will be a trade-off between the performance and computational resources that are required. As a result, the method of choice can depend on the available computational resources. 

\paragraph{Optimization} Another important challenge of jointly learning multiple tasks is properly weighting the loss functions associated with the individual tasks. In particular, the network weight update can be suboptimal when the task gradients conflict, or dominated by one task when its gradient magnitude is much higher with respect to the other tasks. 

There have been several attempts already to address this problem. Early work~\cite{kendall2018multi} used the homoscedastic uncertainty of each task to weigh the losses. Gradient normalization~\cite{chen2018gradnorm} balances the learning of tasks by dynamically adapting the gradient magnitudes in the network. Liu~\emph{et~al.}~\cite{liu2019end} weigh the losses to match the pace at which different tasks are learned. Dynamic task prioritization~\cite{guo2018dynamic} prioritizes the learning of difficult tasks. Sener~\emph{et~al.}~\cite{sener2018multi} cast multi-task learning as a multi-objective optimization problem, with the overall objective of finding a Pareto optimal solution. 

The availability of all these different methods might give the false impression that the problem of setting the task-specific weights is solved. However, as we will show in this manuscript, many existing methods were designed with specific datasets or tasks in mind. In contrast, we aim to develop techniques that are more robust across different settings. 

\paragraph{Pretraining} Existing models for scene understanding often rely on supervised pretraining to obtain good performance. This builds upon the fact that the weights of convolutional neural networks transfer well between tasks. In this way, pretraining on a large dataset like ImageNet~\cite{deng2009imagenet} allows to perform well on tasks with only a modest amount of labelled data. This is the case for both single- and multi-task learning networks.

Unfortunately, supervised pretraining requires access to an annotated pretraining dataset which limits it's applicability. To resolve this problem, researchers have begun to explore self-supervised learning methods which learn representations by solving pretext tasks for which the supervisory signal comes for free. Self-supervised learning can leverage huge amounts of raw unlabeled images.

Although the pretraining aspect is not directly tied to multi-task learning itself, we study it's use as it proves a crucial component for existing scene understanding models described in the literature. Also, an interesting property of working on the pretraining aspect, is that any improvements directly translate to several downstream tasks like object detection, semantic segmentation, etc. 

\section{Overview \& Contributions}
This manuscript focuses on multi-task learning with deep neural networks in the context of visual scene understanding. The work presented in this thesis has been spread across several academic papers. We present an overview and list the contributions made in each chapter.

\textbf{Chapter~\ref{ch: background}} familiarizes the reader with the necessary background materials. We define the multi-task learning setup within the context of deep neural networks. Further, we introduce the datasets and evaluation metrics that are used throughout this thesis.

\textbf{Chapter~\ref{ch: related_work}} reviews recent methods for multi-task learning, including both architectural and optimization based strategies. For each method, we describe its key aspects, discuss the commonalities and differences with related works, and present the possible advantages or disadvantages. Further, we also provide a novel taxonomy to classify recent contributions to deep multi-task learning within the context of computer vision. Finally, starting from our literature review, we motivate the various lenses through which we study the multi-task learning problem in the chapters that follow. The content of this chapter is based upon the following publication (journal):

\begin{itemize}
    \item \emph{\textbf{Vandenhende, S.}, Georgoulis, S., Van Gansbeke, W., Proesmans, M., Dai, D., \& Van Gool, L. (2021). Multi-task learning for dense prediction tasks: A survey. IEEE Transactions on Pattern Analysis and Machine Intelligence.~\cite{vandenhende2021multi}}
\end{itemize}

\textbf{Chapter~\ref{ch: branched_mtl}} presents a novel architecture for multi-task learning. We study branched multi-task networks, where layers gradually grow more task-specific as we move to the deeper layers. Further, we introduce a principled approach to automatically construct branched multi-task networks. The construction process groups related tasks together while striking a trade-off between task similarity and network complexity. The task relatedness scores are based on the premise that similar tasks can be solved with similar features. Our experimental evaluation shows that our architectures outperform other competing models which require a comparable amount of parameters or computations. The content of this chapter is based upon the following publication (poster presentation): 

\begin{itemize}
     \item \emph{\textbf{Vandenhende, S.}, Georgoulis, S., De Brabandere, B., \& Van Gool, L. (2020). Branched Multi-Task Networks: Deciding What Layers To Share. The British Machine Vision Conference.~\cite{vandenhende2019branched}}
\end{itemize}

\textbf{Chapter~\ref{ch: mti_net}} explicitly focuses on devising a better architecture for tackling multiple pixel-level prediction tasks. We argue about the importance of modeling task interactions at multiple scales when distilling the task information under a multi-task learning setup. In contrast to common belief, we show that tasks with high affinity at a certain scale are not guaranteed to retain this behaviour at other scales, and vice versa. We propose a novel architecture, namely MTI-Net, that builds upon this finding in three ways. First, it explicitly models task interactions at every scale via a multi-scale multi-modal distillation unit. Second, it propagates distilled task information from lower to higher scales via a feature propagation module. Third, it aggregates the refined task features from all scales via a feature aggregation unit to produce the final per-task predictions. Experiments on two multi-task dense labeling datasets show that our model delivers on the full potential of multi-task learning, that is, smaller memory footprint, reduced number of calculations, and better performance w.r.t single-task learning. The content of this chapter is based upon the following publication  (spotlight presentation): 
\begin{itemize}
    \item \emph{\textbf{Vandenhende, S.}, Georgoulis, S., \& Van Gool, L. (2020). MTI-Net: Multi-scale task interaction networks for multi-task learning. European Conference on Computer Vision.}
\end{itemize}

\textbf{Chapter~\ref{ch: optimization}} considers the problem of balancing the learning of tasks. First, we analyze existing strategies and isolate the elements that are most successful. We observe that existing methods still suffer from several shortcomings and highlight the discrepancies between existing works. Finally, we propose a set of heuristics to balance the tasks. The use of heuristics results in more robust performance across different datasets. The content of this chapter is based upon the following publication (journal):

\begin{itemize}
    \item \emph{\textbf{Vandenhende, S.}, Georgoulis, S., Van Gansbeke, W., Proesmans, M., Dai, D., \& Van Gool, L. (2021). Multi-task learning for dense prediction tasks: A survey. IEEE Transactions on Pattern Analysis and Machine Intelligence.~\cite{vandenhende2021multi}}
\end{itemize}

\textbf{Chapter~\ref{ch: ssl}} studies contrastive self-supervised learning. First, we study how biases in the dataset affect an existing method like MoCo. We empirically show that an approach like MoCo works well across: (i) object- versus scene-centric, (ii) uniform versus long-tailed and (iii) general versus domain-specific datasets. Second, given the generality of the approach, we try to realize further gains with minor modifications. We show that learning additional invariances - through the use of multi-scale cropping, stronger augmentations and nearest neighbors - improves the representations. Further, we observe that MoCo learns spatially structured representations when trained with a multi-crop strategy. The representations can be used for semantic segment retrieval and video instance segmentation without finetuning. Moreover, the results are on par with specialized models. Finally, we show that MoCo can be used to improve an existing model for indoor scene understanding. The content of this chapter is based upon the following publication~\footnote{* denotes equal contribution.}: 

\begin{itemize}
    \item \emph{Van Gansbeke, W.*, \textbf{Vandenhende, S.}*, Georgoulis, S., \& Van Gool, L. (2021). Revisiting Contrastive Methods for Unsupervised Learning of Visual Representations. NeurIPS.}
\end{itemize}

\textbf{Chapter~\ref{ch: conclusion}} concludes this manuscript.

\textbf{Acknowledgment.} I would like to acknowledge support by Toyota via the TRACE project and MACCHINA (KU Leuven, C14/18/065). This work is also sponsored by the Flemish Government under the Flemish AI programme.

\cleardoublepage
  \graphicspath{{\chaptersdir/background/image/}}%
\chapter{Background}
\label{ch: background} 
This chapter introduces the necessary background materials related to our work. First, we briefly discuss how neural networks are utilized under the multi-task learning setup. Second, we cover the different datasets that will be used in this manuscript. Third, we describe our evaluation criteria.

\section{Multi-Task Learning in Neural Networks}
\label{sec: mtl_setup}
In machine learning, we try to tackle problems by building a model based on training examples. The model can then be used to make predictions or decisions for new examples, avoiding the need to program the desired behavior explicitly. More concretely, given a set of training examples $\mathcal{D}_{\mathcal{T}}$ for task $\mathcal{T}$, we try to obtain a model by optimizing a performance criterion $\mathcal{L}_\mathcal{T}$ for the task. In computer vision, the model is typically represented by a convolutional neural network~\cite{goodfellow2016deep,lecun1998convolutional,lecun1989backpropagation} $f_\theta$ with learnable parameters $\theta$ and trained with the backpropagation algorithm. 

\begin{figure}
\begin{minipage}[b]{.48\textwidth}
    \centering
    \includegraphics[width=\linewidth]{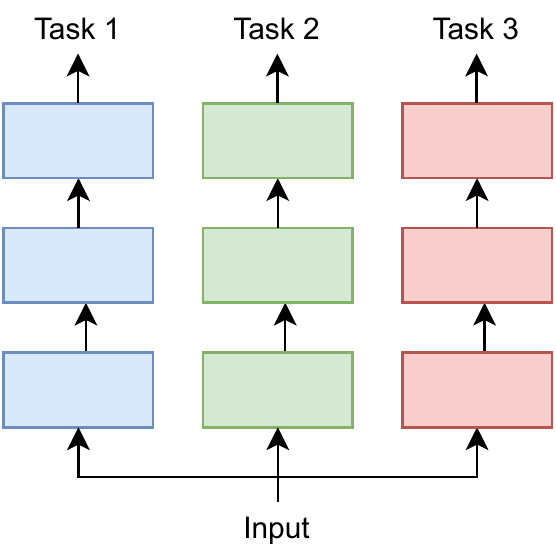}
    \caption[The single-task learning setup]{Single-task learning of three different tasks with the same inputs. Each of the neural networks is optimized independently.}
    \label{fig: single_task_setup}
\end{minipage}
\hfill
\begin{minipage}[b]{.48\textwidth}
    \centering
    \includegraphics[width=\linewidth]{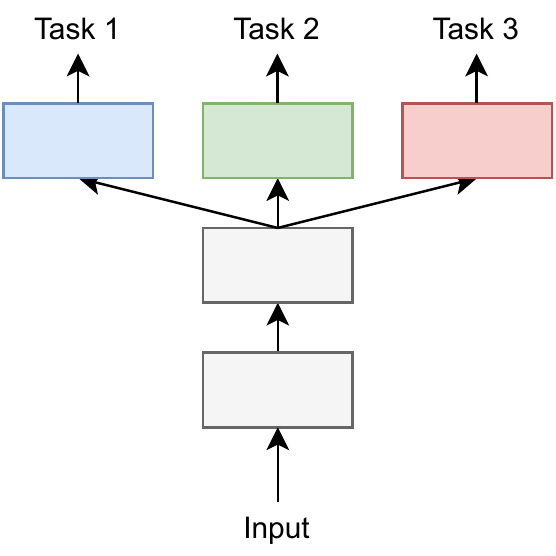}
    \caption[The multi-task learning setup]{Multi-task learning of three different tasks. We optimize all tasks in parallel, while sharing the learned representations.}
    \label{fig: multi_task_setup}
\end{minipage}
\end{figure}

The dominant paradigm is to learn one task at a time. For example, we train a separate neural network for detecting objects, estimating depth and predicting attributes. In this case, each neural network is a function of the same input image and produces one output. Since the neural networks are disconnected, it is not possible for tasks to utilize the - possibly relevant - information that is being extracted by other tasks. Figure~\ref{fig: single_task_setup} visualizes such a \emph{single-task learning} setup. 

Differently, a \emph{multi-task network}~\cite{caruana1997multitask} has the same inputs as the single-task networks, but infers all desired task outputs together. Figure~\ref{fig: multi_task_setup} shows a schematic diagram. The multi-task network is trained by applying backpropagation in parallel on all task outputs. The multi-task learning objective is defined as
\begin{equation}
\label{eq: mtl_objective}
\mathcal{L}_{MTL} = \sum_i w_i \cdot \mathcal{L}_i
\end{equation}
where $\mathcal{L}_i$ denotes the task-specific loss for task $i$ and $w_i$ is a task-specific weight. Since the tasks share a number of layers in the network, the internal representations that are being learned for one task can also be utilized by other tasks. Learning tasks in parallel while sharing the learned representations is the key idea in multi-task learning. 

Caruana~\cite{caruana1997multitask} showed that multi-task learning nets can improve generalization by using information from the training signals of related tasks. In particular, he identified several mechanisms that allow to improve generalization. We provide a brief overview below:
\begin{itemize}
    \item \textbf{Data amplification:} We effectively increase the size of the training data due to the additional information in the training signals of related tasks.
    \item \textbf{Eavesdropping:} Let $F$ be a feature, useful to the tasks $T$ and $T'$, that is easy to pick up when learning task $T$ but difficult to discover when learning $T'$. In a MTL net, task $T'$ could eavesdrop on the features learned for $T$, e.g. $F$, and obtain better performance.
    \item \textbf{Attributes selection:} Assume two tasks $T$ and $T'$ that use a common feature $F$. In this case, the MTL net will be better at selecting data attributes relevant to $F$ because there is a stronger training signal for these attributes.
    \item \textbf{Representation bias:} Representations that allow to solve multiple tasks better represent the data's regularities. 
    \item \textbf{Overfitting:} Tasks are less likely to overfit onto a feature $F$ if we train them together. 
\end{itemize}
We refer to the original work for a detailed analysis. 

The challenges for training multi-task learning models are two-fold. First, we need to design appropriate architectures that can handle multiple tasks. Second, we need to come up with effective ways of jointly optimizing multiple objectives. 

\section{Datasets}
\label{sec: datasets}
We perform our experimental evaluation on a variety of benchmarks. The included datasets are of varying size, cover different task dictionaries and contain annotations that were provided by human annotators or obtained through distillation. This diverse group of settings allows us to scrutinize the pros and cons of the methods under consideration. We describe the different benchmarks below.

The \textbf{PASCAL} dataset~\cite{everingham2010pascal} is a popular benchmark for dense prediction tasks. We use the split from PASCAL-Context~\cite{chen2014detect} which has annotations for semantic segmentation, human part segmentation and semantic edge detection. Additionally, we consider the tasks of surface normals prediction and saliency detection. The annotations were distilled by~\cite{maninis2019attentive} using pre-trained state-of-the-art models~\cite{bansal2017pixelnet,chen2018encoder}.

The \textbf{Cityscapes} dataset~\cite{cordts2016cityscapes} considers the scenario of urban scene understanding. The train, validation and test set contain 2975, 500 and 1525 images respectively, taken by driving a car in European cities. We consider a number of dense prediction tasks on the Cityscapes dataset: semantic segmentation, instance segmentation and monocular depth estimation. The depth maps were generated with SGM~\cite{hirschmuller2008stereo}. 

The \textbf{NYUD} dataset~\cite{silberman2012indoor} considers indoor scene understanding. The dataset contains 795 train and 654 test images annotated for semantic segmentation and monocular depth estimation. Other works have also considered surface normal prediction~\cite{xu2018pad,zhang2019pattern,maninis2019attentive} and semantic edge detection~\cite{xu2018pad,maninis2019attentive} on the NYUD-v2 dataset. The annotations for these tasks can be directly derived from the semantic and depth ground truth. In this manuscript we focus on the semantic segmentation and depth estimation tasks. 

The \textbf{Taskonomy} dataset~\cite{zamir2018taskonomy} contains semi-real images of indoor scenes, annotated for 26 (dense prediction, classification, etc.) tasks. Out of all available tasks, we select scene categorization, semantic segmentation, edge detection, monocular depth estimation and keypoint detection for our experiments. The task dictionary was selected to be as diverse as possible, while still keeping the total number of tasks reasonably low for all required computations. We use the tiny split of the dataset, containing 275k train, 52k validation and 54k test images. 

The \textbf{CelebA} dataset~\cite{liu2015faceattributes} contains over 200k images of celebrities labeled with 40 facial attribute categories. The training, validation and test set contain 160k, 20k and 20k images respectively. While this dataset is mainly suitable for multi-attribute learning, prior works~\cite{lu2017fully,sener2018multi,vandenhende2019branched} used it for multi-task learning by treating the prediction of each facial attribute as a separate binary classification task. This universal use of a binary cross-entropy loss across all tasks makes CelebA an interesting setting to compare the influence of different strategies for setting the task-specific weights in Equation~\ref{eq: mtl_objective}, thus we chose to also include it in our experiments. 

\begin{figure}
    \centering
    \includegraphics[width=.8\linewidth]{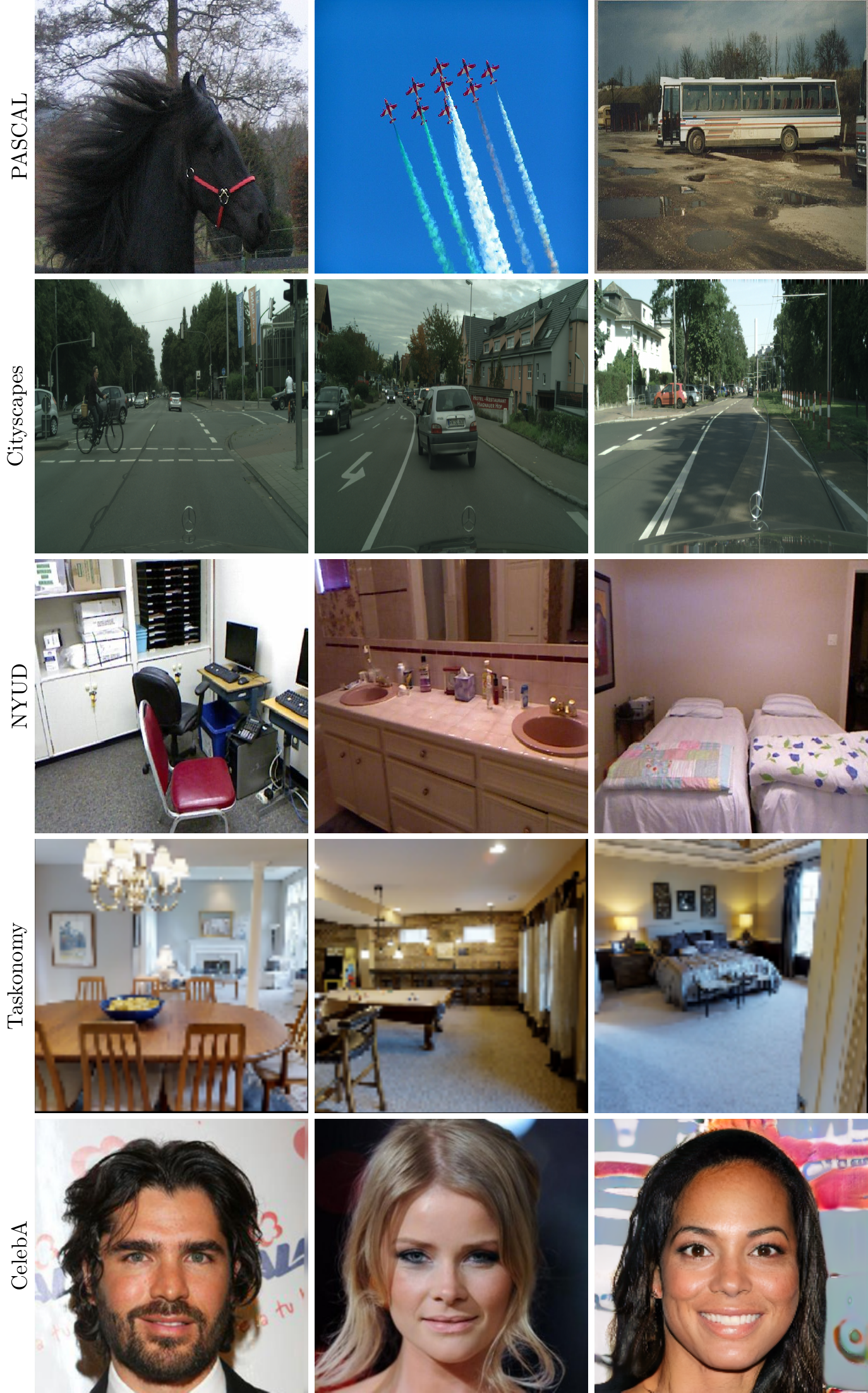}
    \caption[Dataset examples]{A few example images. The aspect ratios can differ from the originals due to the visualization.}
    \label{fig: datasets}
\end{figure}

Table~\ref{tab: datasets} shows an overview of the datasets. Figure~\ref{fig: datasets} shows a few example images from the different datasets. 

\begin{table}
\centering
\caption[Overview of the multi-task learning benchmarks.]{Overview of the multi-task learning benchmarks. Distilled task labels are marked with an asterisk.}
\label{tab: datasets}
\tiny{
\begin{tabular}{z{18}|x{14}x{14}x{14}|x{11}x{11}x{11}x{11}x{11}x{11}x{11}x{11}x{11}x{11}x{11}x{11}}
\toprule
Dataset & \#Train & \#Val & \#Test & Segm. & Depth & Inst. & Parts & Norm. & Sal. & Edges & Keyp. & Scene & Attr. \\ 
\midrule
PASCAL & 10,581 & 1,449 & - & $\checkmark$ &  & & $\checkmark$ & $\checkmark$* & $\checkmark$* & $\checkmark$ & & & \\
Cityscapes & 2,975 & 500 & 500 & $\checkmark$ & $\checkmark$* & $\checkmark$ & & & & & & & \\ 
NYUD & 795 & 654 & -  & $\checkmark$ & $\checkmark$ & & & $\checkmark$ & & $\checkmark$ & & & \\ 
Taskonomy & 275k & 52k & 54k & $\checkmark$* & $\checkmark$ & & & & & $\checkmark$* & $\checkmark$* & $\checkmark$* & \\
CelebA & 160k & 20k & 20k & & & & & & & & & & $\checkmark$ \\ 
\bottomrule
\multicolumn{14}{c}{Segm.: Semantic Segmentation, Inst.: Instance Segmentation, Parts: Human Parts Segmentation, Norm: Surface Normals Prediction, Sal.: Saliency Estimation}\\
\multicolumn{14}{c}{Edges: Semantic Edges or Boundaries, Keyp: 2D-Keypoint Estimation, Scene: Scene Categorization, Attr.: Person Attributes Classification.}
\end{tabular}}
\end{table}

\section{Evaluation Criteria}
\label{sec: evaluation}
In addition to reporting the performance on every individual task, we include a single-number performance metric for the multi-task models. We define the multi-task performance $\Delta_{MTL}$ of a multi-task learning model $m$ as the average per-task drop or increase in performance w.r.t. the single-task baseline $b$:
\begin{equation}
\label{eq: evaluation_metric}
\Delta_{MTL} = \frac{1}{T} \sum_{i=1}^{T} \left(-1\right)^{l_i} \left(M_{m,i} - M_{b,i} \right) / M_{b,i},
\end{equation}
where $l_i = 1$ if a lower value means better performance for metric $M_i$ of task $i$, and 0 otherwise. The single-task performance is measured for a fully-converged model that uses the same backbone network only to perform that task. To achieve a fair comparison, we report the results after performing a search on the most important hyperparameters: optimizer, initial learning rate, learning rate scheduler and weight decay. Hyperparameters that are method-specific are set in accordance with the setup advised by the authors. This practice ensures that every model is trained with comparable amounts of finetuning. 

The MTL performance metric does not account for the variance when different hyperparameters are used. To address this, we analyze the influence of the used hyperparameters with performance profiles for some experiments. Finally, in addition to a performance evaluation, we also include the model resources, i.e. number of parameters and FLOPS, when comparing the multi-task architectures. 

\section{Conclusion}
We have described the problem setup, the benchmarks and the evaluation criteria that will be used in this manuscript. The most important challenges for multi-task learning consist of designing appropriate architectures and finding effective training recipes. The next chapter summarizes prior works that considered these problems.

\cleardoublepage%

  \graphicspath{{\chaptersdir/relatedwork/image/}}%
\chapter{Related Work}
\label{ch: related_work} 
The objective of this chapter is two-fold. First, we aim to provide a structured overview of state-of-the-art MTL techniques for computer vision. Second, starting from the proposed taxonomy, we try to motivate the different lenses through which we study the multi-task learning problem in this thesis. 

\definecolor{level1}{RGB}{244,142,130}
\definecolor{level2}{RGB}{99,175,243}
\definecolor{level3}{RGB}{116,223,123}
\definecolor{level4}{RGB}{255,224,128}

\begin{figure*}[ht]
\centering
\medskip
\footnotesize
\begin{forest}
for tree={s sep=1mm, inner sep=2, l=1}
[
{Multi-Task Learning Methods },fill=level1
[Deep Multi-Task\\ Architectures \textbf{(Sec.~\ref{sec: related_architectures})},fill=level2
[Encoder-Focused\\ \textbf{(Sec.~\ref{subsec: related_encoder})},fill=level3
[{MTL Baseline}\\Cross-Stitch Networks~\cite{misra2016cross}\\ Sluice Networks~\cite{ruder2019latent}\\NDDR-CNN~\cite{gao2019nddr}\\MTAN~\cite{liu2019end}\\{\color{Purple} \textbf{Branched MTL~(Ch.~\ref{ch: branched_mtl})}},fill=level4]]
[Decoder-Focused\\ \textbf{(Sec.~\ref{subsec: related_decoder})},fill=level3
[PAD-Net~\cite{xu2018pad}\\PAP-Net~\cite{zhang2019pattern}\\JTRL~\cite{zhang2018joint}\\PSD~\cite{zhou2020pattern}\\{\color{Purple} \textbf{MTI-Net~(Ch.~\ref{ch: mti_net})}},fill=level4]]
]
[{Optimization Strategy \\ Methods \textbf{(Sec.~\ref{sec: related_optimization})}},fill=level2
[Task Balancing\\ \textbf{(Sec.~\ref{subsec: related_task_balancing})},fill=level3
[Fixed\\Uncertainty~\cite{kendall2018multi}\\GradNorm~\cite{chen2018gradnorm}\\DWA~\cite{liu2019end}\\DTP~\cite{guo2018dynamic}\\Multi-Obj. Opt.~\cite{sener2018multi,lin2019pareto}\\ {\color{Purple} \textbf{Heuristics~(Ch.~\ref{ch: optimization})}},fill=level4]
]
[{Other\\ \textbf{(Sec.~\ref{subsec: related_optimization_other})}},fill=level3
[Adversarial~\cite{liu2017adversarial,maninis2019attentive,sinha2018gradient}\\Modulation~\cite{zhao2018modulation}\\Heuristics~\cite{sanh2019hierarchical,raffel2019exploring}\\Grad Sign Dropout~\cite{chen2020just},fill=level4]
]
]
]
\end{forest}
\caption[A taxonomy of multi-task learning approaches for computer vision]{A taxonomy of recent deep learning approaches for jointly solving multiple visual recognition tasks. We indicate {\color{Purple}\textbf{methods}} that are proposed in this manuscript.}
\label{fig: taxonomy}
\end{figure*}
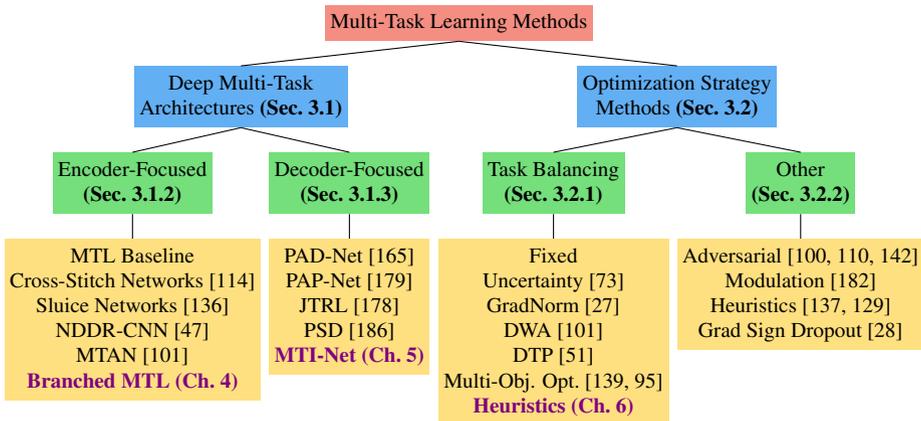

Figure~\ref{fig: taxonomy} shows an overview of the methods that we will review. Section~\ref{sec: related_architectures} considers different deep multi-task architectures, categorizing them into two main groups: encoder- and decoder-focused approaches. Section~\ref{sec: related_optimization} surveys various optimization techniques for balancing the influence of the tasks when updating the network's weights. We consider the majority of task balancing, adversarial and modulation techniques. Section~\ref{sec: related_conclusion} concludes our literature review.

The content of this chapter is based upon the following publication:
\begin{itemize}
    \item \emph{Vandenhende, S., Georgoulis, S., Van Gansbeke, W., Proesmans, M., Dai, D., \& Van Gool, L. (2021). Multi-task learning for dense prediction tasks: A survey. IEEE Transactions on Pattern Analysis and Machine Intelligence.}
\end{itemize}

\section{Deep Multi-Task Architectures}
\label{sec: related_architectures}
In this section, we review deep multi-task architectures used in computer vision. First, we give a brief historical overview of MTL approaches, before introducing a novel taxonomy to categorize different methods. Second, we discuss network designs from different groups of works, and analyze their advantages and disadvantages. Note that, as a detailed presentation of each architecture is beyond the scope of this work, in each case we refer the reader to the corresponding paper for further details that complement the following descriptions. 

\begin{figure*}
\begin{minipage}[b]{1.0\textwidth}
\centering
\begin{subfigure}[t]{.438\linewidth}
  \centering
  \includegraphics[width=\textwidth]{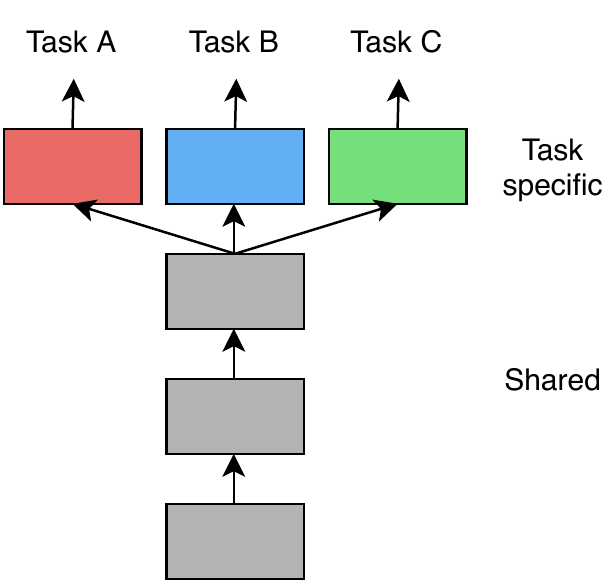}
  \caption{Hard parameter sharing}
  \label{fig: hard_parameter_sharing}
\end{subfigure}%
\hspace*{\fill}
\begin{subfigure}[t]{.523\linewidth}
  \centering
  \includegraphics[width=\textwidth]{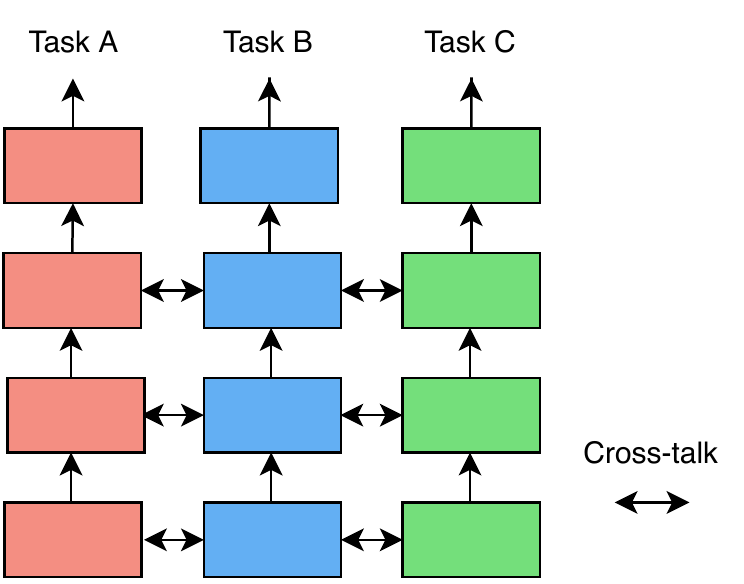}
  \caption{Soft parameter sharing}
  \label{fig: soft_parameter_sharing}
\end{subfigure}
\caption[Soft versus hard parameter sharing models]{Historically multi-task learning using deep neural networks has been subdivided into soft- and hard-parameter sharing schemes.}
\label{fig: multi_task_architectures}
\end{minipage}
\end{figure*}

\subsection{Historical Overview and Taxonomy}
\label{subsec: historical}

\subsubsection{Non-Deep Learning Methods}
Before the deep learning era, MTL works tried to model the common information among tasks in the hope that a joint task learning would result in better generalization performance. To achieve this, they placed assumptions on the task parameter space, such as: task parameters should lie close to each other w.r.t. some distance metric~\cite{evgeniou2004regularized,xue2007multi,jacob2009clustered,zhou2011clustered}, share a common probabilistic prior~\cite{bakker2003task,yu2005learning,lee2007learning,daume2009bayesian,kumar2012learning}, or reside in a low dimensional subspace~\cite{argyriou2008convex,liu2012multi,jalali2010dirty} or manifold~\cite{agarwal2010learning}. These assumptions work well when all tasks are related~\cite{evgeniou2004regularized,ando2005framework,argyriou2008convex,rai2010infinite}, but can lead to performance degradation if information sharing happens between unrelated tasks. The latter is a known problem in MTL, referred to as \emph{negative transfer}. To mitigate this problem, some of these works opted to cluster tasks into groups based on prior beliefs about their similarity or relatedness.

\subsubsection{Soft and Hard Parameter Sharing in Deep Learning} 
In the context of deep learning, MTL is performed by learning shared representations from multi-task supervisory signals. Historically, deep multi-task architectures were classified into hard or soft parameter sharing techniques. In \textit{hard parameter sharing}, the parameter set is divided into shared and task-specific parameters (see Figure~\ref{fig: hard_parameter_sharing}). MTL models using hard parameter sharing typically consist of a shared encoder that branches out into task-specific heads~\cite{neven2017fast,kendall2018multi,chen2018gradnorm,sener2018multi,teichmann2018multinet}. In \textit{soft parameter sharing}, each task is assigned its own set of parameters and a feature sharing mechanism handles the cross-task talk (see Figure~\ref{fig: soft_parameter_sharing}). We summarize representative works for both groups of works below. 

\paragraph{Hard Parameter Sharing} UberNet~\cite{kokkinos2017ubernet} was the first hard-parameter sharing model to jointly tackle a large number of low-, mid-, and high-level vision tasks. The model featured a multi-head design across different network layers and scales. Still, the most characteristic hard parameter sharing design consists of a shared encoder that branches out into task-specific decoding heads~\cite{neven2017fast,kendall2018multi,chen2018gradnorm,sener2018multi,teichmann2018multinet}. Multilinear relationship networks~\cite{long2017learning} extended this design by placing tensor normal priors on the parameter set of the fully connected layers. In a different vein, stochastic filter grouping~\cite{bragman2019stochastic} re-purposes the convolution kernels in each layer to support shared or task-specific behavior. 

\paragraph{Soft Parameter Sharing} Cross-stitch networks~\cite{misra2016cross} introduced soft-parameter sharing in deep MTL architectures. The model uses a linear combination of the activations in every layer of the task-specific networks as a means for soft feature fusion. Sluice networks~\cite{ruder2019latent} extended this idea by allowing to learn the selective sharing of layers, subspaces and skip connections. NDDR-CNN~\cite{gao2019nddr} also incorporated dimensionality reduction techniques into the feature fusion layers. Differently, MTAN~\cite{liu2019end} used an attention mechanism to share a general feature pool amongst the task-specific networks. A concern with soft parameter sharing approaches is scalability, as the size of the multi-task network tends to grow linearly with the number of tasks.

\subsubsection{Distilling Task Predictions in Deep Learning}
\label{subsubsec: distilling}
All deep learning works presented so far follow a common pattern: they \textit{directly} predict all task outputs from the same input in one processing cycle. In this case, we rely entirely on the backpropagation algorithm to find the right representation to solve the tasks. However, it's unlikely that such an approach will be able to discover all useful relationships between tasks. For example, the model might fail to align discontinuities for tasks like depth estimation and semantic segmentation. For this reason, a few recent works first employed a multi-task network to make initial task predictions, and then leveraged features from these initial predictions to further improve each task output -- in a one-off or recursive manner. In this case, the task relationships are modeled in a more explicit way, which makes it easier to exploit correlations between tasks that allow to improve the performance.

Before the deep learning era, Hoiem~\emph{et~al.}~\cite{hoiem2008closing} applied the idea of recursively predicting tasks to develop an integrated 3D scene understanding system with estimates of surface orientations, occlusion boundaries, objects, camera viewpoint, and relative depth. Later, Multi-Net~\cite{bilen2016integrator} re-introduced the concept in the context of deep neural networks. Opposed to encoder-focused approaches, Multi-Net does not only share the features between tasks, but also allows tasks to interact in a recursive manner by encoding the task predictions in a common shared representation. PAD-Net~\cite{xu2018pad} proposed to distill information from the initial task predictions of other tasks, by means of spatial attention, before adding it as a residual to the task of interest. JTRL~\cite{zhang2018joint} opted for sequentially predicting each task, with the intention to utilize information from the past predictions of one task to refine the features of another task at each iteration. PAP-Net~\cite{zhang2019pattern} extended upon this idea, and used a recursive procedure to propagate similar cross-task and task-specific patterns found in the initial task predictions. To do so, they operated on the affinity matrices of the initial predictions, and not on the features themselves, as was the case before~\cite{xu2018pad,zhang2018joint}. Zhou et al.~\cite{zhou2020pattern} refined the use of pixel affinities to distill the information by separating inter- and intra-task patterns from each other. 

\subsubsection{A New Taxonomy of MTL Approaches}
\begin{figure*}
\begin{minipage}[b]{1.0\textwidth}
\centering
\begin{subfigure}[t]{.438\linewidth}
  \centering
  \includegraphics[width=\textwidth]{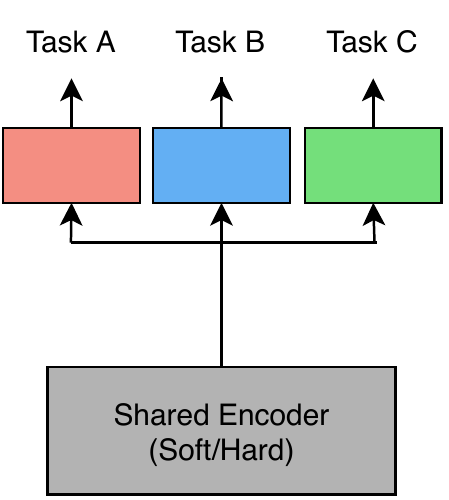}
  \caption{Encoder-focused model}
  \label{fig: encoder_focused_model}
\end{subfigure}%
\hspace*{\fill}
\begin{subfigure}[t]{.438\linewidth}
  \centering
  \includegraphics[width=\textwidth]{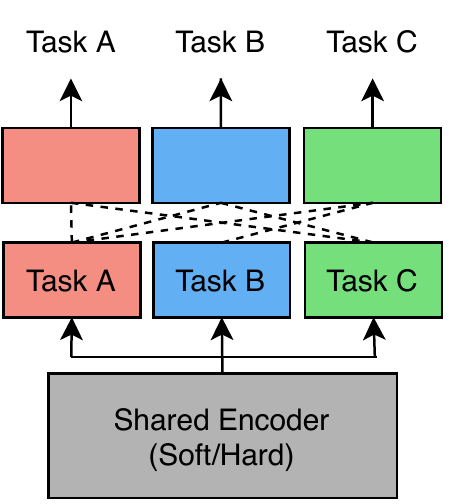}
  \caption{Decoder-focused model}
  \label{fig: decoder_focused_model}
\end{subfigure}
\caption[Encoder- versus decoder-focused models]{In this work we discriminate between encoder- and decoder-focused models depending on where the task interactions take place.}
\label{fig: multi_task_architectures_new}
\end{minipage}
\end{figure*}

As explained before, multi-task networks have historically been classified into soft or hard parameter sharing techniques. However, several recent works took inspiration from both groups of works to jointly solve multiple tasks. As a consequence, it is debatable whether the soft versus hard parameter sharing paradigm should still be used as the main framework for classifying MTL architectures. In this work, we propose an alternative taxonomy that discriminates between different architectures on the basis of where the \emph{task interactions} take place, i.e. locations in the network where information or features are exchanged or shared between tasks. The impetus for this framework was given in the previous subsection. Based on the proposed criterion, we distinguish between two types of models: \emph{encoder-focused} and \emph{decoder-focused} architectures. The encoder-focused architectures (see Figure~\ref{fig: encoder_focused_model}) only share information in the encoder, using either hard- or soft-parameter sharing, before decoding each task with an independent task-specific head. Differently, the decoder-focused architectures (see Figure~\ref{fig: decoder_focused_model}) also exchange information during the decoding stage. In particular, they first make initial task predictions, and then leverage features from these initial task predictions to improve each task output - in a sequential manner. Figure~\ref{fig: taxonomy} gives an overview of the proposed taxonomy, listing representative works in each case.

\subsection{Encoder-focused Architectures}
\label{subsec: related_encoder}
Encoder-focused architectures (see Figure~\ref{fig: encoder_focused_model}) share the task features in the encoding stage, before they process them with a set of independent task-specific heads. A number of works~\cite{neven2017fast,kendall2018multi,chen2018gradnorm,sener2018multi,teichmann2018multinet} followed an ad hoc strategy by sharing an off-the-shelf backbone network in combination with small task-specific heads (see Figure~\ref{fig: hard_parameter_sharing}). This model relies on the encoder (i.e. backbone network) to learn a generic representation of the scene. The features from the encoder are then used by the task-specific heads to get the predictions for every task. While this simple model shares the full encoder amongst all tasks, recent works have considered \textit{where} and \textit{how} the feature sharing should happen in the encoder. We discuss such sharing strategies in the following sections.

\begin{figure}
\begin{minipage}[t]{0.50\linewidth}
\centering
\includegraphics[width=\textwidth]{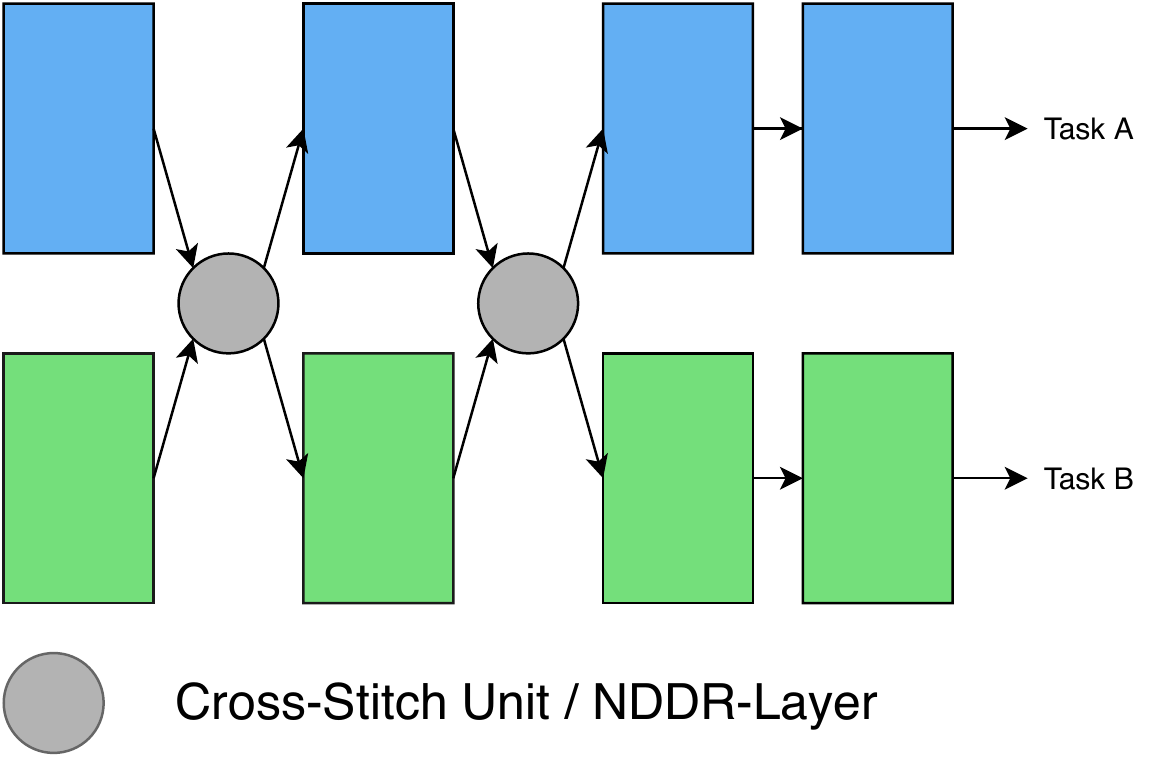}
\caption[Cross-stitch networks and NDDR-CNNs]{The architecture of cross-stitch networks~\cite{misra2016cross} and NDDR-CNNs~\cite{gao2019nddr}. The activations from all single-task networks are fused across several encoding layers. Different feature fusion mechanisms were used in each case.}
\label{fig: crossstitch_nddr}
\end{minipage}
\hspace*{\fill}
\begin{minipage}[t]{0.47\linewidth}
\centering
\includegraphics[width=\textwidth]{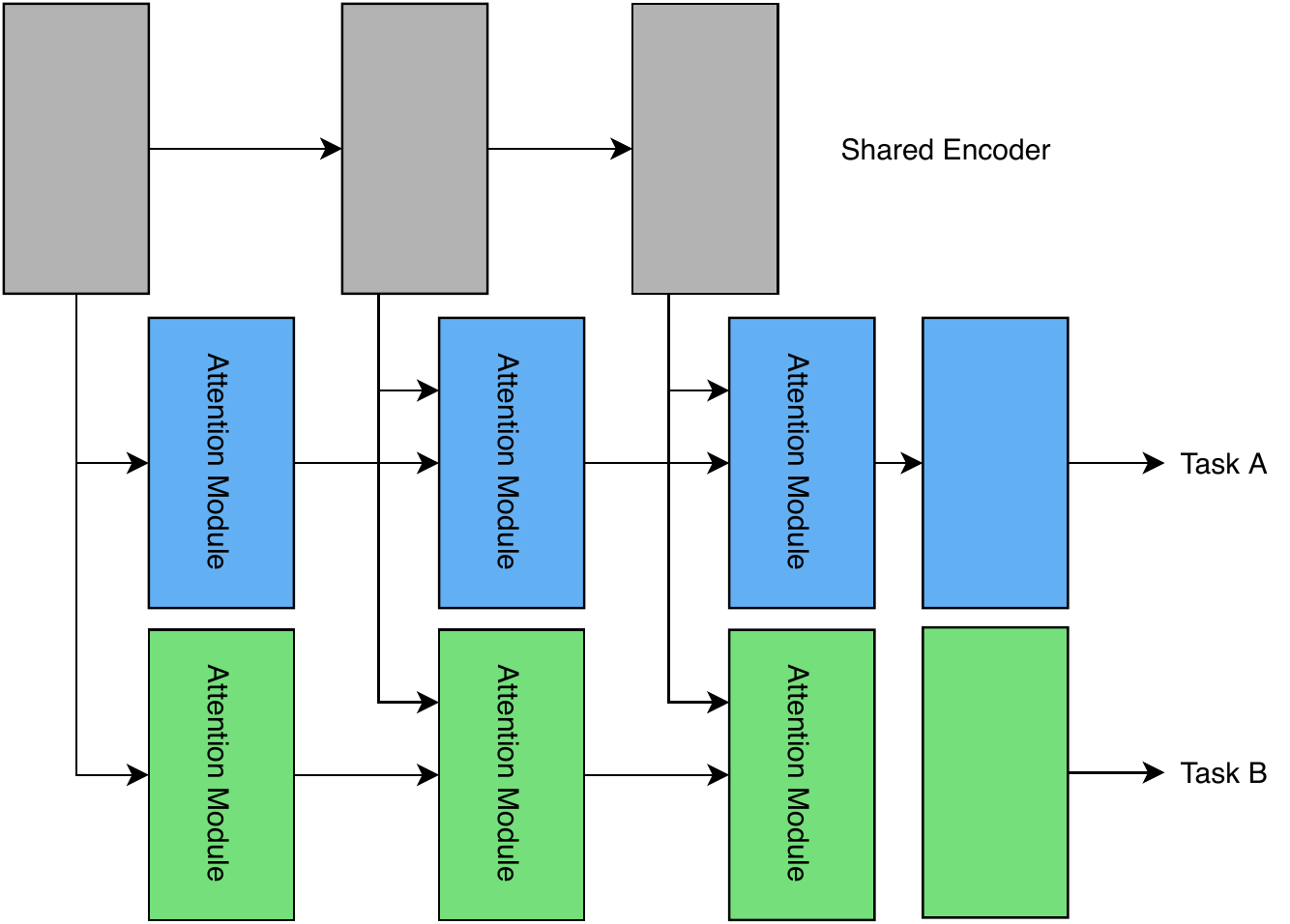}
\caption[Multi-Task Attention Networks]{The architecture of MTAN~\cite{liu2019end}. Task-specific attention modules select and refine features from several layers of a shared encoder.}
\label{fig: mtan}
\end{minipage}
\end{figure}

\subsubsection{Cross-Stitch Networks}
\label{subsubsec: related_cross_stitch_nets}

\textbf{Cross-stitch networks}~\cite{misra2016cross} shared the activations amongst all single-task networks in the encoder. Assume we are given two activation maps $x_A, x_B$ at a particular layer, that belong to tasks $A$ and $B$ respectively. A learnable linear combination of these activation maps is applied, before feeding the transformed result $\tilde{x}_A, \tilde{x}_B$ to the next layer in the single-task networks. The transformation is parameterized by learnable scalars $\alpha \in \mathbb{R}$, and can be expressed as

\begin{equation}
\begin{bmatrix}
\tilde{x}_A \\
\tilde{x}_B
\end{bmatrix}
= 
\begin{bmatrix}
\alpha_{AA} & \alpha_{AB} \\
\alpha_{BA} & \alpha_{BB} \\
\end{bmatrix}
\begin{bmatrix}
x_A \\
x_B
\end{bmatrix}.
\end{equation}

As illustrated in Figure~\ref{fig: crossstitch_nddr}, this procedure is repeated at multiple locations in the encoder. By learning the weights $\alpha$, the network can decide the degree to which the features are shared between tasks. In practice, we are required to pre-train the single-task networks, before stitching them together, in order to maximize the performance. A disadvantage of cross-stitch networks is that the size of the network increases linearly with the number of tasks. Furthermore, it is not clear where the cross-stitch units should be inserted in order to maximize their effectiveness. \textbf{Sluice networks}~\cite{ruder2019latent} extended this work by also supporting the selective sharing of subspaces and skip connections. 

\subsubsection{Neural Discriminative Dimensionality Reduction}
\textbf{Neural Discriminative Dimensionality Reduction CNNs} (NDDR-CNNs)~\cite{gao2019nddr} used a similar architecture with cross-stitch networks (see Figure~\ref{fig: crossstitch_nddr}). However, instead of utilizing a linear combination to fuse the activations from all single-task networks, a dimensionality reduction mechanism is employed. First, features with the same spatial resolution in the single-task networks are concatenated channel-wise. Second, the number of channels is reduced by processing the features with a 1 by 1 convolutional layer, before feeding the result to the next layer. The convolutional layer allows to fuse activations across all channels. Differently, cross-stitch networks only allow to fuse activations from channels that share the same index. The NDDR-CNN behaves as a cross-stitch network when the non-diagonal elements in the weight matrix of the convolutional layer are zero.

Due to their similarity with cross-stitch networks, NDDR-CNNs are prone to the same problems. First, there is a scalability concern when dealing with a large number of tasks. Second, NDDR-CNNs involve additional design choices, since we need to decide where to include the NDDR layers. Finally, both cross-stitch networks and NDDR-CNNs only allow to use limited local information (i.e. small receptive field) when fusing the activations from the different single-task networks. We hypothesize that this is suboptimal because the use of sufficient context is very important during encoding -- as already shown for the tasks of image classification~\cite{he2015spatial} and semantic segmentation~\cite{chen2017deeplab,zhao2017pyramid,yuan2018ocnet}. This is backed up by certain decoder-focused architectures in Section~\ref{subsec: related_decoder} that overcome the limited receptive field by predicting the tasks at multiple scales and by sharing the features repeatedly at every scale. 

\subsubsection{Multi-Task Attention Networks}
\textbf{Multi-Task Attention Networks} (MTAN)~\cite{liu2019end} used a shared backbone network in conjunction with task-specific attention modules in the encoder (see Figure~\ref{fig: mtan}). The shared backbone extracts a general pool of features. Then, each task-specific attention module selects features from the general pool by applying a soft attention mask. The attention mechanism is implemented using regular convolutional layers and a sigmoid non-linearity. Since the attention modules are small compared to the backbone network, the MTAN model does not suffer as severely from the scalability issues that are typically associated with cross-stitch networks and NDDR-CNNs. However, similar to the fusion mechanism in the latter works, the MTAN model can only use limited local information to produce the attention mask.  

\subsubsection{Towards Branched Multi-Task Networks (Chapter~\ref{ch: branched_mtl})}
Most of the encoder-focused architectures softly share the features amongst tasks during the encoding stage. These methods report good performance, but require a considerable amount of computational resources. In contrast, hard parameter sharing models consume fewer resources, but report lower performance compared to their soft parameter sharing equivalents. The latter can be explained as follows. Assuming a hard parameter sharing setting, the number of possible network configurations grows quickly with the number of tasks. Therefore, previous works on hard parameter sharing opted for the simple strategy of sharing the initial layers in the network, after which all tasks branch out simultaneously. The point at which the branching occurs is usually determined ad hoc~\cite{kendall2018multi,guo2018dynamic,sener2018multi}. This situation hurts performance, as a suboptimal grouping of tasks will lead to the sharing of information between unrelated tasks. 

\textbf{Branched Multi-Task Networks} present an approach to decide on the degree of layer sharing between tasks in order to eliminate the need for manual exploration. To this end, we base the layer sharing on measurable levels of \emph{task affinity} or \emph{task relatedness}: two tasks are strongly related, if their single task models rely on a similar set of features. Additionally, our method allows to trade network complexity against task similarity. We provide extensive empirical evaluation of our method, showing its superiority in terms of multi-task performance versus computational resources. 

\subsection{Decoder-Focused Architectures}
\label{subsec: related_decoder}
The encoder-focused architectures in Section~\ref{subsec: related_encoder} follow a common pattern: they \textit{directly} predict all task outputs from the same input in one processing cycle (i.e. all predictions are generated once, in parallel or sequentially, and are not refined afterwards). By doing so, they fail to capture commonalities and differences among tasks, that are likely fruitful for one another (e.g. depth discontinuities are usually aligned with semantic edges). Arguably, this might be the reason for the moderate only performance improvements achieved by the encoder-focused approaches to MTL. To alleviate this issue, a few recent works first employed a multi-task network to make initial task predictions, and then leveraged features from these initial predictions in order to further improve each task output -- in an one-off or recursive manner. As these MTL approaches also share or exchange information during the decoding stage, we refer to them as decoder-focused architectures (see Figure~\ref{fig: decoder_focused_model}). 

\subsubsection{PAD-Net}
\begin{figure}
    \centering
    \includegraphics[width=.8\linewidth]{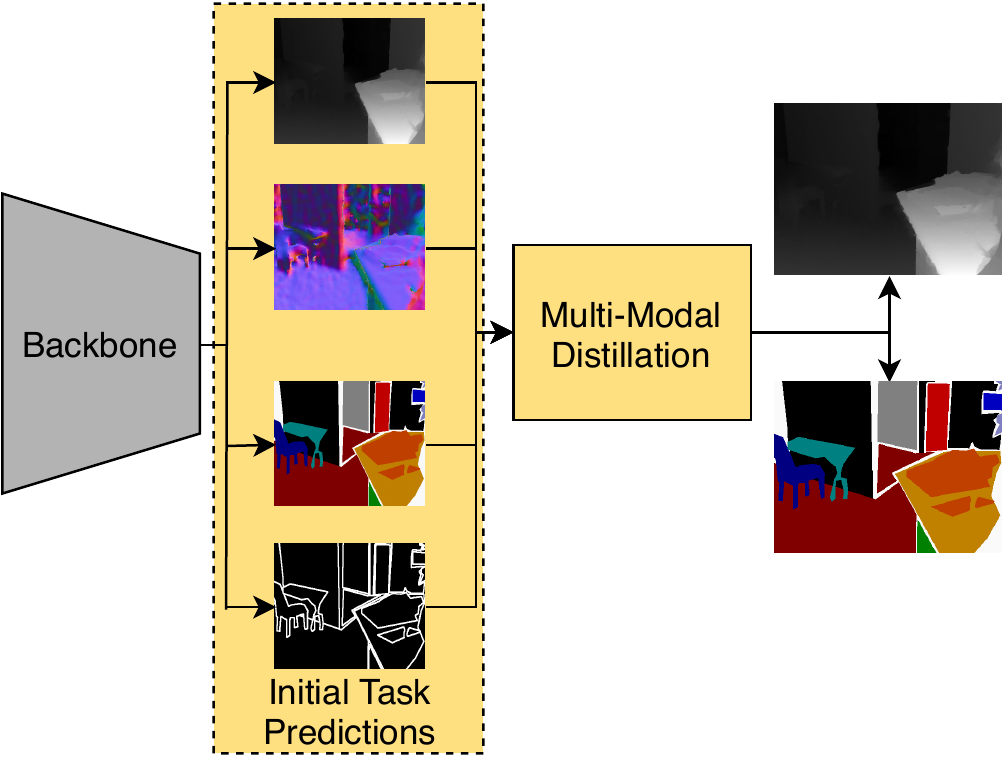}
    \caption[PAD-Net]{The architecture in PAD-Net~\cite{xu2018pad}. Features extracted by a backbone network are passed to task-specific heads to make initial task predictions. The task features from the different heads are then combined through a distillation unit to make the final predictions. Note that, auxiliary tasks can be used in this framework, i.e. tasks for which only the initial predictions are generated, but not the final ones.}
    \label{fig: padnet}
\end{figure}

\textbf{PAD-Net}~\cite{xu2018pad} was one of the first decoder-focused architectures. The model itself is visualized in Figure~\ref{fig: padnet}. As can be seen, the input image is first processed by an off-the-shelf backbone network. The backbone features are further processed by a set of task-specific heads that produce an initial prediction for every task. These initial task predictions add deep supervision to the network, but they can also be used to exchange information between tasks, as will be explained next. The \textit{task features} in the last layer of the task-specific heads contain a per-task feature representation of the scene. PAD-Net proposed to re-combine them via a multi-modal distillation unit, whose role is to extract cross-task information, before producing the final task predictions. 

PAD-Net performs the multi-modal distillation by means of a spatial attention mechanism. Particularly, the output features $F_{k}^o$ for task $k$ are calculated as
\begin{equation}
\label{eq: spatial_attention}
F_{k}^{o} = F_{k}^i + \sum_{l \neq k} \sigma \left(W_{k,l} F_{l}^{i} \right) \odot F_l^i,
\end{equation}
where $\sigma \left(W_{k,l} F_{l}^{i} \right)$ returns a spatial attention mask that is applied to the initial task features $F_{l}^i$ from task $l$. The attention mask itself is found by applying a convolutional layer $W_{k,l}$ to extract local information from the initial task features. Equation~\ref{eq: spatial_attention} assumes that the task interactions are location dependent, i.e. tasks are not in a constant relationship across the entire image. This can be understood from a simple example. Consider two dense prediction tasks, e.g. monocular depth prediction and semantic segmentation. Depth discontinuities and semantic boundaries often coincide. However, when we segment a flat object, e.g. a magazine, from a flat surface, e.g. a table, we will still find a semantic boundary where the depth map is rather continuous. In this particular case, the depth features provide no additional information for the localization of the semantic boundaries. The use of spatial attention explicitly allows the network to select information from other tasks at locations where its useful. 

The encoder-focused approaches in Section~\ref{subsec: related_encoder} shared features amongst tasks using the intermediate representations in the encoder. Differently, PAD-Net models the task interactions by applying a spatial attention layer to the features in the task-specific heads. In contrast to the intermediate feature representations in the encoder, the task features used by PAD-Net are already disentangled according to the output task. We hypothesize that this makes it easier for other tasks to distill the relevant information. This multi-step decoding strategy from PAD-Net is applied and refined in other decoder-focused approaches.  

\subsubsection{Pattern-Affinitive Propagation Networks}

\begin{figure}
    \centering
    \includegraphics[width=.8\linewidth]{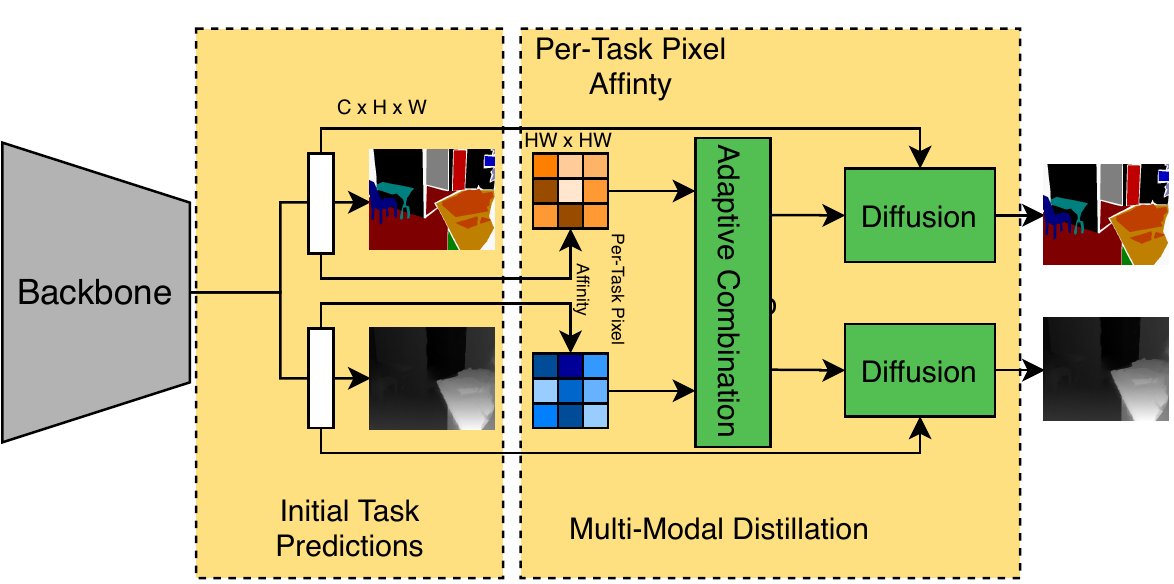}
    \caption[Pattern-Affinitive Propagation Networks]{The architecture in PAP-Net~\cite{zhang2019pattern}. Features extracted by a backbone network are passed to task-specific heads to make initial task predictions. The task features from the different heads are used to calculate a per-task pixel affinity matrix. The affinity matrices are adaptively combined and diffused back into the task features space to spread the cross-task correlation information across the image. The refined features are used to make the final task predictions.}
    \label{fig: papnet}
\end{figure}

\textbf{Pattern-Affinitive Propagation Networks} (PAP-Net)~\cite{zhang2019pattern} used an architecture similar to PAD-Net (see Figure~\ref{fig: papnet}), but the multi-modal distillation in this work is performed in a different manner. The authors argue that directly working on the task features space via the spatial attention mechanism, as done in PAD-Net, might be a suboptimal choice. As the optimization still happens at a different space, i.e. the task label space, there is no guarantee that the model will learn the desired task relationships. Instead, they statistically observed that pixel affinities tend to align well with common local structures on the task label space. Motivated by this observation, they proposed to leverage pixel affinities in order to perform multi-modal distillation.

To achieve this, the backbone features are first processed by a set of task-specific heads to get an initial prediction for every task. Second, a per-task pixel affinity matrix $M_{T_j}$ is calculated by estimating pixel-wise correlations upon the task features coming from each head. Third, a cross-task information matrix $\hat{M}_{T_j}$ for every task $T_j$ is learned by adaptively combining the affinity matrices $M_{T_{i}}$ for tasks $T_i$ with learnable weights $\alpha_i^{T_j}$ 
\begin{equation}
\hat{M}_{T_j} = \sum_{T_i} \alpha_i^{T_j} \cdot M_{T_i}.
\end{equation}
Finally, the task features coming from each head $j$ are refined using the cross-task information matrix $\hat{M}_{T_j}$. In particular, the cross-task information matrix is diffused into the task features space to spread the correlation information across the image. This effectively weakens or strengthens the pixel correlations for task $T_j$, based on the pixel affinities from other tasks $T_i$. The refined features are used to make the final predictions for every task. 

\subsubsection{Joint Task-Recursive Learning}

\begin{figure}
    \centering
    \includegraphics[width=.8\linewidth]{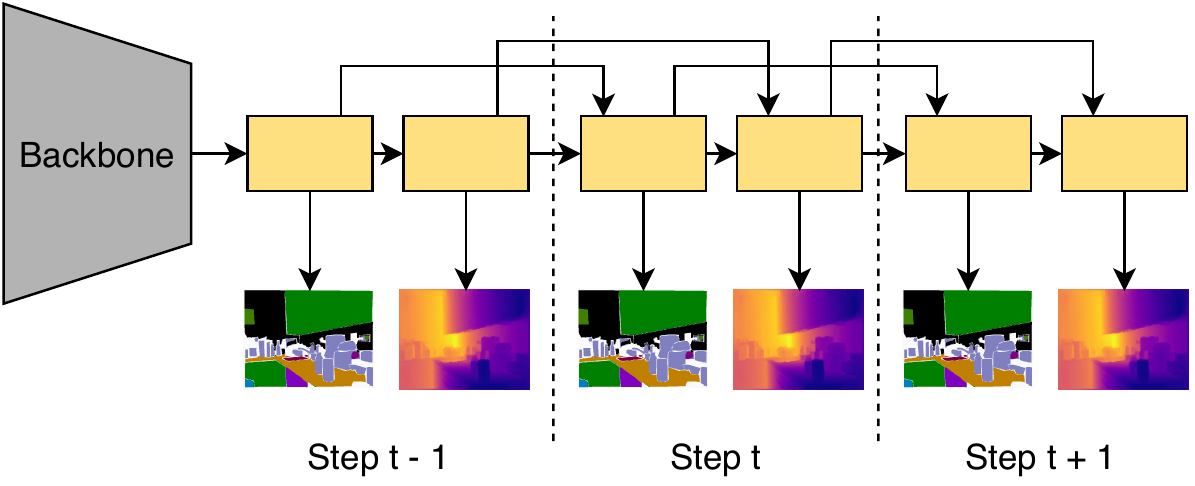}
    \caption[Joint-Task Recursive Learning]{The architecture in Joint Task-Recursive Learning~\cite{zhang2018joint}. The features of two tasks are progressively refined in an intertwined manner based on past states.}
    \label{fig: jtrl}
\end{figure}

\textbf{Joint Task-Recursive Learning} (JTRL)~\cite{zhang2018joint} recursively predicts two tasks at increasingly higher scales in order to gradually refine the results based on past states. The architecture is illustrated in Figure~\ref{fig: jtrl}. Similarly to PAD-Net and PAP-Net, a multi-modal distillation mechanism is used to combine information from earlier task predictions, through which later predictions are refined. Differently, the JTRL model predicts two tasks sequentially, rather than in parallel, and in an intertwined manner. The main disadvantage of this approach is that it is not straightforward, or even possible, to extent this model to more than two tasks given the intertwined manner at which task predictions are refined. 

\subsubsection{Towards Multi-Scale Task Interaction Networks (Chapter~\ref{ch: mti_net})}

The decoder-focused architectures perform multi-modal distillation at a fixed scale, e.g. the features of the backbone's last layer. This rests on the assumption that all relevant task interactions can solely be modeled through a single filter operation with a specific receptive field. In this manuscript, we show that this is a rather strict assumption. In fact, tasks can influence each other differently at different receptive fields. To account for this, we propose a \textbf{Multi-Scale Task Interaction Network (MTI-Net)} that explicitly takes into account task interactions at multiple scales. This research is presented in Chapter~\ref{ch: mti_net}.

\subsection{Other Approaches}
\label{subsec: related_architectures_other}
A number of approaches that fall outside the aforementioned categories have been proposed in the literature. For example, multilinear relationship networks~\cite{long2017learning} used tensor normal priors for the parameter set of the task-specific heads to allow interactions in the decoding stage. Different from the standard parallel ordering scheme, where layers are aligned and shared (e.g.~\cite{misra2016cross,gao2019nddr}), soft layer ordering~\cite{meyerson2018beyond} proposed a flexible sharing scheme across tasks and network depths. Yang et al.~\cite{yang2016deep} generalized matrix factorisation approaches to MTL in order to learn cross-task sharing structures in every layer of the network. Routing networks~\cite{rosenbaum2018routing} proposed a principled approach to determine the connectivity of a network's function blocks through routing. Piggyback~\cite{mallya2018piggyback} showed how to adapt a single, fixed neural network to a multi-task network by learning binary masks. Huang et al.~\cite{huang2018gnas} introduced a method rooted in Neural Architecture Search (NAS) for the automated construction of a tree-based multi-attribute learning network. Stochastic filter grouping~\cite{bragman2019stochastic} re-purposed the convolution kernels in each layer of the network to support shared or task-specific behaviour. In a similar vein, feature partitioning~\cite{newell2019feature} presented partitioning strategies to assign the convolution kernels in each layer of the network into different tasks. Attentive Single-Tasking of Multiple Tasks (ASTMT)~\cite{maninis2019attentive} proposed to take a 'single-tasking' route for the MTL problem. That is, within a multi-tasking framework they perform separate forward passes, one for each task, that activate shared responses among all tasks, plus some residual responses that are task-specific. 

\section{Optimization in MTL}
\label{sec: related_optimization}
In the previous section, we discussed the construction of network architectures that are able to learn multiple tasks concurrently. Still, a significant challenge in MTL stems from the optimization procedure itself. In particular, we need to carefully balance the joint learning of all tasks to avoid a scenario where one or more tasks have a dominant influence in the network weights. In this section, we discuss several methods that have considered this \textit{task balancing} problem. 

\subsection{Task Balancing Approaches}
\label{subsec: related_task_balancing}
Without loss of generality, the optimization objective in a MTL problem, assuming task-specific weights $w_i$ and task-specific loss functions $\mathcal{L}_i$, can be formulated as 
\begin{equation}
\label{eq: related_mtl_loss}
\mathcal{L}_{MTL} = \sum_{i} w_i \cdot \mathcal{L}_i.
\end{equation}
When using stochastic gradient descent to minimize the objective from Equation~\ref{eq: related_mtl_loss}, which is the standard approach in the deep learning era, the network weights in the shared layers $W_{sh}$ are updated by the following rule
\begin{equation}
\label{eq: related_update_rule}
W_{sh} = W_{sh} - \gamma \sum_{i} w_i \pdv{\mathcal{L}_i}{W_{sh}}.
\end{equation}

From Equation~\ref{eq: related_update_rule} we can draw the following conclusions. First, the network weight update can be suboptimal when the task gradients conflict, or dominated by one task when its gradient magnitude is much higher w.r.t. the other tasks. This motivated researchers~\cite{kendall2018multi,guo2018dynamic,liu2019end,chen2018gradnorm} to balance the gradient magnitudes by setting the task-specific weights $w_i$ in the loss. To this end, other works~\cite{sener2018multi,zhao2018modulation,suteu2019regularizing} have also considered the influence of the direction of the task gradients. Second, each task's influence on the network weight update can be controlled, either \textit{indirectly} by adapting the task-specific weights $w_i$ in the loss, or \textit{directly} by operating on the task-specific gradients $\pdv{\mathcal{L}_i}{W_{sh}}$. A number of methods that tried to address these problems are discussed next. 

\subsubsection{Uncertainty Weighting}
Kendall et al.~\cite{kendall2018multi} used the \textbf{homoscedastic uncertainty} to balance the single-task losses. The homoscedastic uncertainty or \textit{task-dependent uncertainty} is not an output of the model, but a quantity that remains constant for different input examples of the same task. The optimization procedure is carried out to maximize a Gaussian likelihood objective that accounts for the homoscedastic uncertainty. In particular, they optimize the model weights $W$ and the noise parameters $\sigma_1$, $\sigma_2$ to minimize the following objective 
\begin{equation}
\label{eq: uncertainty_weighting}
\mathcal{L}\left(W,\sigma_1,\sigma_2\right) = \frac{1}{2\sigma^2_1} \mathcal{L}_1 \left(W\right) + \frac{1}{2\sigma^2_2} \mathcal{L}_2 \left(W\right) + \log \sigma_1 \sigma_2.
\end{equation}
The loss functions $\mathcal{L}_1$, $\mathcal{L}_2$ belong to the first and second task respectively. By minimizing the loss $\mathcal{L}$ w.r.t. the noise parameters $\sigma_1$, $\sigma_2$, one can essentially balance the task-specific losses during training. The optimization objective in Equation~\ref{eq: uncertainty_weighting} can easily be extended to account for more than two tasks too. The noise parameters are updated through standard backpropagation during training. 

Note that, increasing the noise parameter $\sigma_i$ reduces the weight for task $i$. Consequently, the effect of task $i$ on the network weight update is smaller when the task's homoscedastic uncertainty is high. This is advantageous when dealing with noisy annotations since the task-specific weights will be lowered automatically for such tasks.  

\subsubsection{Gradient Normalization}
\textbf{Gradient normalization} (GradNorm)~\cite{chen2018gradnorm} proposed to control the training of multi-task networks by stimulating the task-specific gradients to be of similar magnitude. By doing so, the network is encouraged to learn all tasks at an equal pace. Before presenting this approach, we introduce the necessary notations in the following paragraph.

We define the L2 norm of the gradient for the weighted single-task loss $w_i\left(t\right) \cdot L_i\left(t\right)$ at step $t$ w.r.t. the weights $W$, as $G_i^W\left(t\right)$. We additionally define the following quantities, 
\begin{itemize}
\item the mean task gradient $\bar{G}^{W}$ averaged across all task gradients $G_i^{W}$ w.r.t the weights $W$ at step $t$: $\bar{G}^{W} \left(t\right) = E_{task}\left[G_i^W\left(t\right)\right]$;
\item the inverse training rate $\tilde{L}_{i}$ of task $i$ at step $t$: $ \tilde{L}_{i}\left(t\right) = L_i \left(t\right) / L_i\left(0\right)$;
\item the relative inverse training rate of task $i$ at step $t$: $r_i\left(t\right) = \tilde{L}_i\left(t\right) / E_{task}\left[\tilde{L}_i\left(t\right)\right]$.
\end{itemize}

GradNorm aims to balance two properties during the training of a multi-task network. First, balancing the gradient magnitudes $G_i^W$. To achieve this, the mean gradient $\bar{G}^W$ is considered as a common basis from which the relative gradient sizes across tasks can be measured. Second, balancing the pace at which different tasks are learned. The relative inverse training rate $r_i\left(t\right)$ is used to this end. When the relative inverse training rate $r_i\left(t\right)$ increases, the gradient magnitude $G_i^W\left(t\right)$ for task $i$ should increase as well to stimulate the task to train more quickly. GradNorm tackles both objectives by minimizing the following loss

\begin{equation}
\label{eq: gradnorm}    
\left\vert G_i^W \left(t\right) - \bar{G}^W \left(t\right) \cdot r_i\left(t\right) \right\vert.
\end{equation}

Remember that, the gradient magnitude $G_i^W(t)$ for task $i$ depends on the weighted single-task loss $w_i(t)\cdot L_i(t)$. As a result, the objective in Equation~\ref{eq: gradnorm} can be minimized by adjusting the task-specific weights $w_i$. In practice, during training these task-specific weights are updated in every iteration using backpropagation. After every update, the task-specific weights $w_i\left(t\right)$ are re-normalized in order to decouple the learning rate from the task-specific weights.

Note that, calculating the gradient magnitude $G_i^W\left(t\right)$ requires a backward pass through the task-specific layers of every task $i$. However, savings on computation time can be achieved by considering the task gradient magnitudes only w.r.t. the weights in the last shared layer. 

Different from uncertainty weighting, GradNorm does not take into account the task-dependent uncertainty to re-weight the task-specific losses. Rather, GradNorm tries to balance the pace at which tasks are learned, while avoiding gradients of different magnitude. 

\subsubsection{Dynamic Weight Averaging}
Similar to GradNorm, Liu et al.~\cite{liu2019end} proposed a technique, termed \textbf{Dynamic Weight Averaging} (DWA), to balance the pace at which tasks are learned. Differently, DWA only requires access to the task-specific loss values. This avoids having to perform separate backward passes during training in order to obtain the task-specific gradients. In DWA, the task-specific weight $w_i$ for task $i$ at step $t$ is set as 
\begin{equation}
\label{eq: dynamic_weight_averaging}
w_i\left(t\right) = \frac{N \exp \left(r_i \left( t-1 \right)/T \right)}{\sum_n \exp \left(r_n \left(t-1\right)/T\right)}, r_n\left(t-1\right) = \frac{L_n\left(t-1\right)}{L_n\left(t-2\right)},
\end{equation}
with $N$ being the number of tasks. The scalars $r_n\left(\cdot\right)$ estimate the relative descending rate of the task-specific loss values $L_n$. The temperature $T$ controls the softness of the task weighting in the softmax operator. When the loss of a task decreases at a slower rate compared to other tasks, the task-specific weight in the loss is increased.

Note that, the task-specific weights $w_i$ are solely based on the rate at which the task-specific losses change. Such a strategy requires to balance the overall loss magnitudes beforehand, else some tasks could still overwhelm the others during training. GradNorm avoids this problem by balancing both the training rates and the gradient magnitudes through a single objective (see Equation~\ref{eq: gradnorm}).

\subsubsection{Dynamic Task Prioritization}
The presented task balancing techniques optimized the task-specific weights $w_i$ as part of a Gaussian likelihood objective~\cite{kendall2018multi}, or in order to balance the pace at which the different tasks are learned~\cite{liu2019end,chen2018gradnorm}. In contrast, \textbf{Dynamic Task Prioritization} (DTP)~\cite{guo2018dynamic} opted to prioritize the learning of 'difficult' tasks by assigning them a higher task-specific weight. The motivation is that the network should spend more effort to learn the 'difficult' tasks. Note that, this is opposed to uncertainty weighting, where a higher weight is assigned to the 'easy' tasks. We hypothesize that the two techniques do not necessarily conflict, but uncertainty weighting seems better suited when tasks have noisy labeled data, while DTP makes more sense when we have access to clean ground-truth annotations.  

To measure the task difficulty, one could consider the progress on every task using the loss ratio $\tilde{L}_i(t)$ defined by GradNorm. However, since the loss ratio depends on the initial loss $L_i(0)$, its value can be rather noisy and initialization dependent. Furthermore, measuring the task progress using the loss ratio might not accurately reflect the progress on a task in terms of qualitative results. Therefore, DTP proposes the use of key performance indicators (KPIs) to quantify the difficulty of every task. In particular, a KPI $\kappa_i$ is selected for every task $i$, with $0 < \kappa_i < 1$. The KPIs are picked to have an intuitive meaning, e.g. accuracy for classification tasks. For regression tasks, the prediction error can be thresholded to obtain a KPI that lies between 0 and 1. Further, we define a task-level focusing parameter $\gamma_i \geq 0 $ that allows to adjust the weight at which easy or hard tasks are down-weighted. DTP sets the task-specific weight $w_i$ for task $i$ at step $t$ as
\begin{equation}
\label{eq: dynamic_task_priorization}
w_i\left(t\right) =  -\left( 1 - \kappa_i \left(t\right)\right)^{\gamma_i} \log \kappa_i \left(t\right).
\end{equation}
Note that, Equation~\ref{eq: dynamic_task_priorization} employs a focal loss expression~\cite{lin2017focal} to down-weight the task-specific weights for the 'easy' tasks. In particular, as the value for the KPI $\kappa_i$ increases, the weight $w_i$ for task $i$ is being reduced.

DTP requires to carefully select the KPIs. For example, consider choosing a threshold to measure the performance on a regression task. Depending on the threshold's value, the task-specific weight will be higher or lower during training. We conclude that the choice of the KPIs in DTP is not determined in a straightforward manner. Furthermore, similar to DWA, DTP requires to balance the overall magnitude of the loss values beforehand. After all, Equation~\ref{eq: dynamic_task_priorization} does not take into account the loss magnitudes to calculate the task-specific weights. As a result, DTP still involves manual tuning to set the task-specific weights. 

\subsubsection{MTL as Multi-Objective Optimization}
A global optimum for the multi-task optimization objective in Equation~\ref{eq: related_mtl_loss} is hard to find. Due to the complex nature of this problem, a certain choice that improves the performance for one task could lead to performance degradation for another task. The task balancing methods discussed beforehand try to tackle this problem by setting the task-specific weights in the loss according to some heuristic. Differently, Sener and Koltun~\cite{sener2018multi} view MTL as a multi-objective optimization problem, with the overall goal of finding a Pareto optimal solution among all tasks. 

In MTL, a Pareto optimal solution is found when the following condition is satisfied: the loss for any task can not be decreased without increasing the loss on any of the other tasks. A \textbf{multiple gradient descent algorithm} (MGDA)~\cite{desideri2012multiple} was proposed in~\cite{sener2018multi} to find a Pareto stationary point. In particular, the shared network weights are updated by finding a common direction among the task-specific gradients. As long as there is a common direction along which the task-specific losses can be decreased, we have not reached a Pareto optimal point yet. An advantage of this approach is that since the shared network weights are only updated along common directions of the task-specific gradients, conflicting gradients are avoided in the weight update step.

Lin et al.~\cite{lin2019pareto} observed that MGDA only finds one out of many Pareto optimal solutions. Moreover, it is not guaranteed that the obtained solution will satisfy the users' needs. To address this problem, they generalized MGDA to generate a set of well-representative Pareto solutions from which a preferred solution can be selected. So far, however, the method was only applied to small-scale datasets (e.g. Multi-MNIST).

\subsubsection{Heuristics for Balancing Tasks in MTL (Chapter~\ref{ch: optimization})}
In Chapter~\ref{ch: optimization}, we uncover several discrepancies between existing task balancing strategies. For example, uncertainty weighting sets a lower weight for difficult tasks, while DTP advocates the opposite. We hypothesize that the performance of different task balancing methods is coupled with specific properties of the training dataset. This hypothesis is verified through an extensive experimental comparison of several task balancing approaches. Further, based upon the results, we propose a set of heuristics to set the task-specific weights. The heuristics are of practical use and show more robust results across different benchmarks. 

\subsection{Other Approaches}
\label{subsec: related_optimization_other}
The task balancing works in Section~\ref{subsec: related_task_balancing} can be plugged into most existing multi-task architectures to regulate the task learning. Another group of works also tried to regulate the training of multi-task networks, albeit for more specific setups. We touch upon several of these approaches here. Note that, some of these concepts can be combined with task balancing strategies too. 

Zhao et al.~\cite{zhao2018modulation} empirically found that tasks with gradients pointing in the opposite direction can cause the destructive interference of the gradient. This observation is related to the update rule in Equation~\ref{eq: related_update_rule}. They proposed to add a modulation unit to the network in order to alleviate the competing gradients issue during training. 

Liu et al.~\cite{liu2017adversarial} considered a specific multi-task architecture where the feature space is split into a shared and a task-specific part. They argue that the shared features should contain more common information, and no information that is specific to a particular task only. The network weights are regularized by enforcing this prior. More specifically, an adversarial approach is used to avoid task-specific features from creeping into the shared representation. Similarly,~\cite{sinha2018gradient,maninis2019attentive} added an adversarial loss to the single-task gradients in order to make them statistically indistinguishable from each other in the shared parts of the network.

Chen et al.~\cite{chen2020just} proposed gradient sign dropout, a modular layer that can be plugged into any network with multiple gradient signals. Following MGDA, the authors argue that conflicts in the weight update arise when the gradient values of the different learning signals have opposite signs. Gradient sign dropout operates by choosing the sign of the gradient based on the distribution of the gradient values, and masking out the gradient values with opposite sign. It is shown that the method has several desirable properties, and increases performance and robustness compared to competing works.

Finally, some works relied on heuristics to balance the tasks. Sanh et al.~\cite{sanh2019hierarchical} trained the network by randomly sampling a single task for the weight update during every iteration. The sampling probabilities were set proportionally to the available amount of training data for every task. Raffel et al.~\cite{raffel2019exploring} used temperature scaling to balance the tasks. So far, however, both procedures were used in the context of natural language processing. 

\section{Conclusion}
\label{sec: related_conclusion}
We reviewed recent methods for multi-task learning within the scope of deep neural networks, including both architectural and optimization based strategies. For each method, we described its key aspects, discussed the commonalities and differences with related works, and presented the possible advantages or disadvantages. Further, starting from the literature, we motivated the various innovations that will be presented in this manuscript:
\begin{itemize}
\item Chapter~\ref{ch: branched_mtl} studies Branched Multi-Task Networks - a hard-parameter sharing architecture that achieves a better trade-off between the performance versus computational resources. The proposed method automatically derives a scheme for sharing the layers amongst tasks. 
\item Chapter~\ref{ch: mti_net} shows how decoder-focused architectures can be improved by distilling the task interactions at a multitude of scales. The novel architecture improves over a set of single-tasking baselines in terms of the memory footprint, number of operations and performance.
\item Chapter~\ref{ch: optimization} performs an experimental comparison of several task balancing strategies. We unravel what properties of the data render some methods more successful than others. Additionally, we propose a set of heuristics to set the task-specific weights under the multi-task learning setup. 
\end{itemize}
\cleardoublepage%

  \graphicspath{{\chaptersdir/branchedmtl/image/}}%
\chapter{Branched Multi-Task Networks}
\label{ch: branched_mtl} 
In this chapter, we study branched multi-task networks as an alternative for existing encoder-focused models. Branched multi-task networks typically start with a number of shared layers, after which different tasks branch out into their own sequence of layers. Understandably, as the number of possible network configurations is combinatorially large, deciding what layers to share and where to branch out becomes cumbersome. Therefore, prior works have relied on ad hoc methods to determine the level of layer sharing, which is found suboptimal. We go beyond these limitations and propose an approach to automatically construct branched multi-task networks, by leveraging the employed tasks' affinities. Given a specific budget, i.e. number of learnable parameters, the proposed approach generates architectures, in which shallow layers are task-agnostic, whereas deeper ones gradually grow more task-specific. An extensive experimental analysis across numerous, diverse multi-tasking datasets shows that, for a given budget, our method consistently yields networks with the highest performance, while for a certain performance threshold it requires the least amount of learnable parameters. 

The content of this chapter is based upon the following publication:
\begin{itemize}
     \item \emph{Vandenhende, S., Georgoulis, S., De Brabandere, B., \& Van Gool, L. (2020). Branched Multi-Task Networks: Deciding What Layers To Share. The British Machine Vision Conference.}
\end{itemize}

\section{Introduction}
\label{sec: branched_motivation}
Consider the following observation: deep neural networks tend to learn hierarchical image representations~\cite{yosinski2014transferable}. The early layers tend to focus on more general low-level image features, such as edges, corners, etc., while the deeper layers tend to extract high-level information that is more task-specific. Motivated by this observation, branched MTL networks opt to learn a similar hierarchical encoding structure (e.g.~\cite{lu2017fully,guo2020learning}). These ramified networks typically start with a number of shared layers, after which different (groups of) tasks branch out into their own sequence of layers. In doing so, the different branches gradually become more task-specific as we move to the deeper layers. This behaviour aligns well with the hierarchical representations learned by deep neural nets.

A significant challenge is to decide on the layers that need to be shared among tasks. In particular, the number of possible network configurations grows quickly with the number of tasks. As a result, a trial-and-error procedure to define the optimal architecture becomes unwieldy. Therefore, previous works opted for the simple strategy of sharing the initial layers in the network, after which all tasks branch out simultaneously. The point at which the branching occurs is usually determined ad hoc~\cite{kendall2018multi,guo2018dynamic,sener2018multi}. This situation hurts performance, as a suboptimal grouping of tasks can lead to the sharing of information between unrelated tasks, known as \emph{negative transfer}~\cite{zhao2018modulation}.

In this chapter, we go beyond the aforementioned limitations and propose a novel approach to decide on the degree of layer sharing between multiple visual recognition tasks in order to eliminate the need for manual exploration. To this end, we base the layer sharing on measurable levels of \emph{task affinity} or \emph{task relatedness}: two tasks are strongly related, if their single task models rely on a similar set of features. Zamir~\emph{et~al.}~\cite{zamir2018taskonomy} quantified this property by measuring the performance when solving a task using a variable sets of layers from a model pretrained on a different task. However, their approach is considerably expensive, as it scales quadratically with the number of tasks. Recently,~\cite{dwivedi2019representation} proposed a more efficient alternative that uses representation similarity analysis (RSA) to obtain a measure of task affinity, by computing correlations between models pretrained on different tasks. Given a dataset and a number of tasks, our approach uses RSA to assess the task affinity at arbitrary locations in a neural network. The task affinity scores are then used to construct a branched multi-task network in a fully automated manner. In particular, our task clustering algorithm groups similar tasks together in common branches, and separates dissimilar tasks by assigning them to different branches, thereby reducing the negative transfer between tasks. Additionally, our method allows to trade network complexity against task similarity. We provide extensive empirical evaluation of our method, showing its superiority in terms of multi-task performance versus computational resources.  

\section{Related Work}
\label{sec: branched_related}
\paragraph{Fully-Adaptive Feature Sharing} Our work bears some similarity to fully-adaptive feature sharing~\cite{lu2017fully} (FAFS), which starts from a thin network where tasks initially share all layers, but the final one, and dynamically grows the model in a greedy layer-by-layer fashion. Task groupings, in this case, are decided on the probability of concurrently simple or difficult examples across tasks. Differently, (1) our method clusters tasks based on feature affinity scores, rather than example difficulty, which is shown to achieve better results for a variety of datasets; (2) the tree structure is determined offline using the precalculated affinities for the whole network, and not online in a greedy layer-by-layer fashion, which promotes task groupings that are optimal in a global, rather than local, sense. 

\paragraph{Neural Architecture Search} Neural architecture search (NAS)~\cite{elsken2019neural} aims to automate the construction of the network architecture. Different algorithms can be characterized based on their search space, search strategy or performance estimation strategy. Most existing works on NAS, however, are limited to task-specific models~\cite{zoph2017neural,liu2018hierarchical,pham2018efficient,liu2018progressive,real2019regularized}. This is to be expected as when using NAS for MTL, layer sharing has to be jointly optimized with the layers types, their connectivity, etc., rendering the problem considerably expensive. To alleviate the heavy computation burden, a recent work~\cite{liang2018evolutionary} implemented an evolutionary architecture search for multi-task networks, while other researchers explored more viable alternatives, like routing~\cite{rosenbaum2018routing}, stochastic filter grouping~\cite{bragman2019stochastic}, and feature partitioning~\cite{newell2019feature}. In contrast to traditional NAS, the proposed methods do not build the architecture from scratch, but rather start from a predefined backbone network for which a layer sharing scheme is automatically determined.

\paragraph{Transfer Learning} Transfer learning~\cite{pan2010survey} uses the knowledge obtained when solving one task, and applies it to a different but related task. Our work is loosely related to transfer learning, as we use it to measure levels of task affinity. Zamir~\emph{et~al.}~\cite{zamir2018taskonomy} provided a taxonomy for task transfer learning to quantify such relationships. However, their approach scales unfavorably w.r.t. the number of tasks, and we opted for a more efficient alternative proposed by~\cite{dwivedi2019representation}. The latter uses RSA to obtain a measure of task affinity, by computing correlations between models pretrained on different tasks. In our method, we use the performance metric from their work to compare the usefulness of different feature sets for solving a particular task. 

\begin{figure}
     \centering
     \begin{subfigure}[t]{.66\textwidth}
        \centering
        \includegraphics[width=\textwidth]{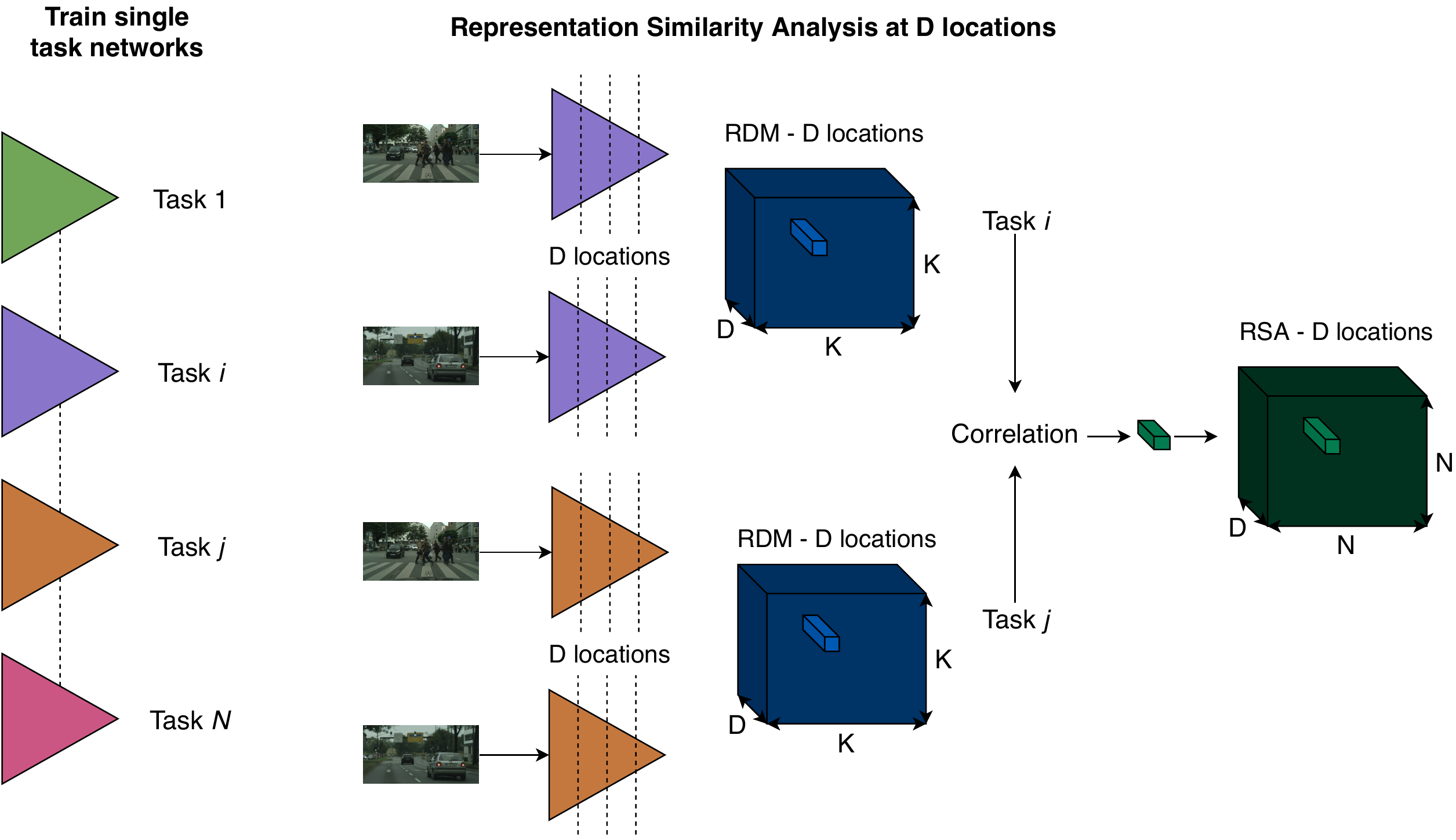}
        \caption[Measuring task relatedness for branched multi-task networks]{Pipeline overview. (left) We train a single-task model for every task $t \in \mathcal{T}$. (middle) We use RSA to measure the task affinity at $D$ predefined locations in the sharable encoder. In particular, we calculate the representation dissimilarity matrices (RDM) for the features at $D$ locations using $K$ images, which gives a $D \times K \times K$ tensor per task. (right) The affinity tensor $\mathbf{A}$ is found by calculating the correlation between the RDM matrices, which results in a three-dimensional tensor of size $D \times N \times N$, with $N$ the number of tasks.}
        \label{fig: branched_rsa}
     \end{subfigure}
     \hfill
     \begin{subfigure}[t]{.32\textwidth}
        \centering
        \includegraphics[width=\textwidth]{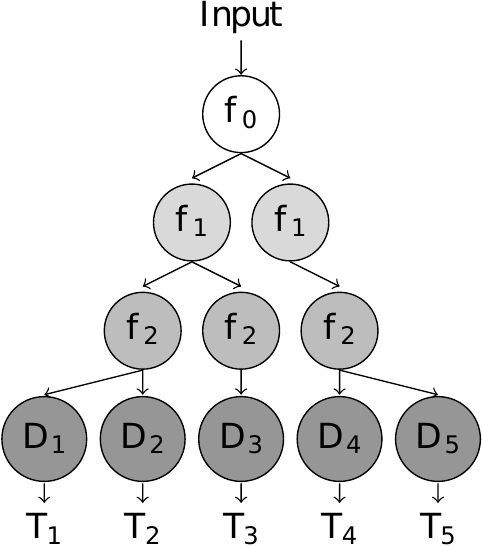}
        \caption[Branched multi-task networks]{Our pipeline's output is a branched multi-task network, similar to how NAS techniques output sample architectures. An example branched multi-task network is visualized here.}
        \label{fig: branched_tree}
     \end{subfigure}
     \caption[Method overview of branched multi-task networks]{The proposed method: (a) calculate task affinities at various locations in the sharable encoder; (b) build a branched multi-task network based on the computed affinities.}
    \label{fig: branched_method}
\end{figure}


\section{Method}
\label{sec: branched_method}
We aim to jointly solve $N$ different visual recognition tasks $\mathcal{T}=\left\{t_1, \ldots, t_N\right\}$ given a computational budget $\mathcal{C}$, i.e. number of parameters or FLOPS. Consider a backbone architecture: an encoder, consisting of a sequence of shared layers or blocks $f_l$, followed by a decoder with a few task-specific layers. We assume an appropriate structure for layer sharing to take the shape of a tree. In particular, the first layers are shared by all tasks, while later layers gradually split off as they show more task-specific behavior. The proposed method aims to find an effective task grouping for the sharable layers $f_l$ of the encoder, i.e. grouping related tasks together in the same branches of the tree. When two tasks are strongly related, we expect their single-task models to rely on a similar feature set~\cite{zamir2018taskonomy}. Based on this viewpoint, the proposed method derives a task affinity score at various locations in the sharable encoder. The number of locations $D$ can be freely determined as the number of candidate branching locations. As such, the resulting task affinity scores are used for the automated construction of a branched multi-task network that fits the computational budget $\mathcal{C}$. Figure~\ref{fig: branched_method} illustrates our pipeline, while Algorithm~\ref{alg: branched_algorithm} summarizes the whole procedure. 

\subsection{Calculate Task Affinity Scores}
As mentioned, we rely on RSA to measure task affinity scores. This technique has been widely adopted in the field of neuroscience to draw comparisons between behavioral models and brain activity (see~\cite{kriegeskorte2008representational}). Inspired by how~\cite{dwivedi2019representation} applied RSA to select tasks for transfer learning, we use the technique to assess the task affinity at predefined locations in the sharable encoder. Consequently, using the measured levels of task affinity, tasks are assigned in the same or different branches of a branched multi-task network, subject to the computational budget $\mathcal{C}$. 

The procedure to calculate the task affinity scores is the following. As a first step, we train a single-task model for each task $t_{i} \in \mathcal{T}$. The single-task models use an identical encoder $E$ - made of all sharable layers $f_l$ - followed by a task-specific decoder $D_{t_{i}}$. The decoder contains only task-specific operations and is assumed to be significantly smaller in size compared to the encoder. As an example, consider jointly solving a classification and a dense prediction task. Some fully connected layers followed by a softmax operation are typically needed for the classification task, while an additional decoding step with some upscaling operations is required for the dense prediction task. Of course, the appropriate loss functions are applied in each case. Such operations are part of the task-specific decoder $D_{t_{i}}$. The different single-task networks are trained under the same conditions.

At the second step, we choose $D$ locations in the sharable encoder $E$ where we calculate a two-dimensional task affinity matrix of size $N \times N$. When concatenated, this results in a three-dimensional tensor $\mathbf{A}$ of size $D \times N \times N$ that holds the task affinities at the selected locations. To calculate these task affinities, we have to compare the representation dissimilarity matrices (RDM) of the single-task networks - trained in the previous step - at the specified $D$ locations. To do this, a held-out subset of $K$ images is required. The latter images serve to compare the dissimilarity of their feature representations in the single-task networks for every pair of images. Specifically, for every task $t_i$, we characterize these learned feature representations at the selected locations by filling a tensor of size $D \times K \times K$. This tensor contains the dissimilarity scores $1-\rho$ between feature representations, with $\rho$ the Pearson correlation coefficient. Specifically, $\mathbf{RDM}_{d,i,j}$ is found by calculating the dissimilarity score between the features at location $d$ for image $i$ and $j$. The 3-D tensors are linearized to 1-D tensors to calculate the pearson correlation coefficient.

For a specific location $d$ in the network, the computed RDMs are symmetrical, with a diagonal of zeros. For every such location, we measure the similarity between the upper or lower triangular part of the RDMs belonging to the different single-task networks. We use the Spearman's correlation coefficient $r_s$ to measure similarity. When repeated for every pair of tasks, at a specific location $d$, the result is a symmetrical matrix of size $N \times N$, with a diagonal of ones. Concatenating over the $D$ locations in the sharable encoder, we end up with the desired task affinity tensor of size $D \times N \times N$. Note that, in contrast to prior work~\cite{lu2017fully}, the described method focuses on the features used to solve the single tasks, rather than the examples and how easy or hard they are across tasks, which is shown to result in better task groupings in Section~\ref{sec: branched_experiments}. Furthermore, the computational overhead to determine the task affinity scores based on feature correlations is negligible. We conclude that the computational cost of the method boils down to pre-training $N$ single task networks. A computational cost analysis can be found in the experiments section.

Other measures of task similarity~\cite{achille2019task2vec,cui2018large} probed the features from a network pre-trained on ImageNet. This avoids the need to pre-train a set of single-task networks first. However, in this case, the task dictionaries only consisted of various, related classification problems. Differently, we consider more diverse, and loosely related task (see Section~\ref{sec: branched_experiments}). In our case, it is arguably more important to learn the task-specific information needed to solve a task. This motivates the use of pre-trained single-task networks. 

\begin{algorithm}[t]
\caption{Branched Multi-Task Networks - Task clustering}
\label{alg: branched_algorithm}
\small{
\begin{algorithmic}[1]
\State \textbf{Input:} Tasks $\mathcal{T}$, $K$ images $\mathcal{I}$, a sharable encoder $E$ with $D$ locations where we can branch, a set of task specific decoders $D_t$ and a computational budget $\mathcal{C}$.
\For{$t$ in $\mathcal{T}$} 
\State Train the encoder $E$ and task-specific decoder $D_t$ for task $t$.
\State $RDM^{t} \leftarrow RDM(E, D, \mathcal{I})$ \Comment{RDM for task $t$.}
\EndFor
\State $A_{d,i,j} \leftarrow r_s\left( \text{triu} \left( RDM^{t_i}_{d,:,:} \right), \text{triu} \left( RDM^{t_j}_{d,:,:}  \right) \right) $ \textbf{for} $t_i, t_j$ in $\mathcal{T}$ and $d$ in locations 
\State $D = 1 - A$ \Comment{Task dissimilarity}
\State \textbf{Return:} Task-grouping with minimal task dissimilarity that fits within $\mathcal{C}$
\end{algorithmic}
}
\end{algorithm}

\subsection{Construct a Branched Multi-Task Network}
Given a computational budget $\mathcal{C}$, we need to derive how the layers (or blocks) $f_l$ in the sharable encoder $E$ should be shared among the tasks in $\mathcal{T}$. Each layer $f_l \in E$ is represented as a node in the tree, i.e. the root node contains the first layer $f_0$, and nodes at depth $l$ contain layer(s) $f_l$. The granularity of the layers $f_l$ corresponds to the intervals at which we measure the task affinity in the sharable encoder, i.e. the $D$ locations. When the encoder is split into $b_{l}$ branches at depth $l$, this is equivalent to a node at depth $l$ having $b_{l}$ children. The leaves of the tree contain the task-specific decoders $D_t$. Figure~\ref{fig: branched_tree} shows an example of such a tree using the aforementioned notation. Each node is responsible for solving a unique subset of tasks.

The branched multi-task network is built with the intention to separate dissimilar tasks by assigning them to separate branches. To this end, we define the dissimilarity score between two tasks $t_i$ and $t_j$ at location $d$ as $1-\mathbf{A}_{d,i,j}$, with $\mathbf{A}$ the task affinity tensor\footnote{This is not to be confused with the dissimilarity score used to calculate the RDM elements $\mathbf{RDM}_{d,i,j}$.}. The branched multi-task network is found by minimizing the sum of the task dissimilarity scores at every location in the sharable encoder. In contrast to prior work~\cite{lu2017fully}, the task affinity (and dissimilarity) scores are calculated a priori. This allows us to determine the task clustering offline. Since the number of tasks is finite, we can enumerate all possible trees that fall within the given computational budget $\mathcal{C}$. Finally, we select the tree that minimizes the task dissimilarity score. The task dissimilarity score of a tree is defined as $C_{cluster} = \sum_{l} C_{cluster}^{l}$, where $C_{cluster}^{l}$ is found by averaging the maximum distance between the dissimilarity scores of the elements in every cluster. The use of the maximum distance encourages the separation of dissimilar tasks. By taking into account the clustering cost at all depths, the procedure can find a task grouping that is considered optimal in a global sense. This is in contrast to the greedy approach in~\cite{lu2017fully}, which only minimizes the task dissimilarity locally, i.e. at isolated locations in the network. 

\section{Experiments}
\label{sec: branched_experiments}

\subsection{Cityscapes}
\label{subsec: branched_cityscapes}

\subsubsection{Setup}
We tackle the semantic segmentation, instance segmentation and monocular depth estimation tasks on Cityscapes. As in prior works~\cite{kendall2018multi, sener2018multi}, we use a ResNet-50 encoder with dilated convolutions, followed by a Pyramid Spatial Pooling (PSP)~\cite{he2015spatial} decoder. Every input image is rescaled to 512 x 256 pixels. We reuse the approach from~\cite{kendall2018multi} for the instance segmentation task, i.e. we consider the proxy task of regressing each pixel to the center of the instance it belongs to. The depth estimation is learned by placing an L1 regression loss on the disparity map. 

\begin{figure}[t]
\begin{minipage}[t]{.48\linewidth}
  \begin{center}
  \includegraphics[width=\textwidth]{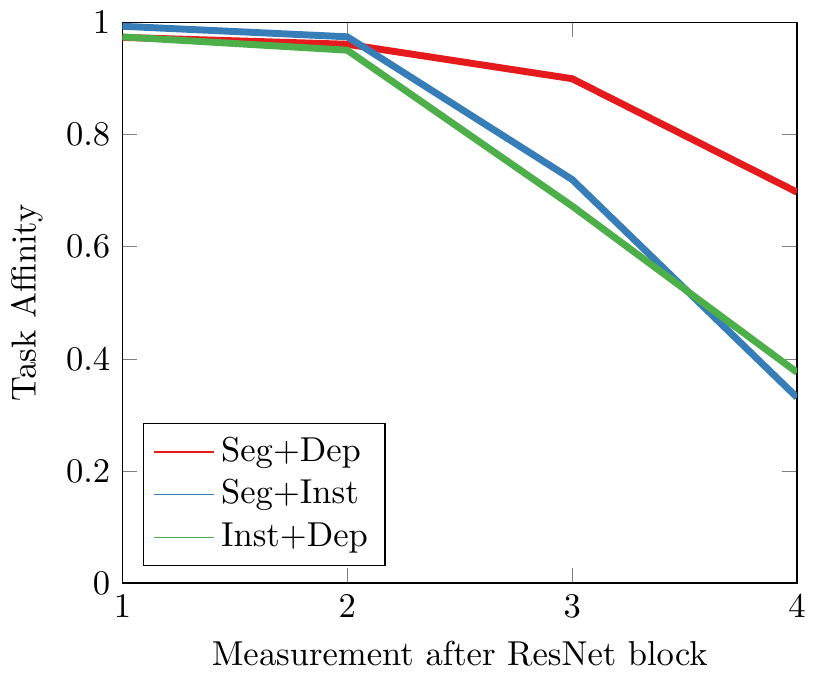}
  \end{center}
  \caption[Task relatedness measured on Cityscapes]{Task affinity scores measured after each residual block in a ResNet-50 model on Cityscapes.}
  \label{fig: branched_cityscapes_affinity}
 \end{minipage}%
 \hfill
 \begin{minipage}[t]{.48\linewidth}
  \begin{center}
  \includegraphics[width=\textwidth]{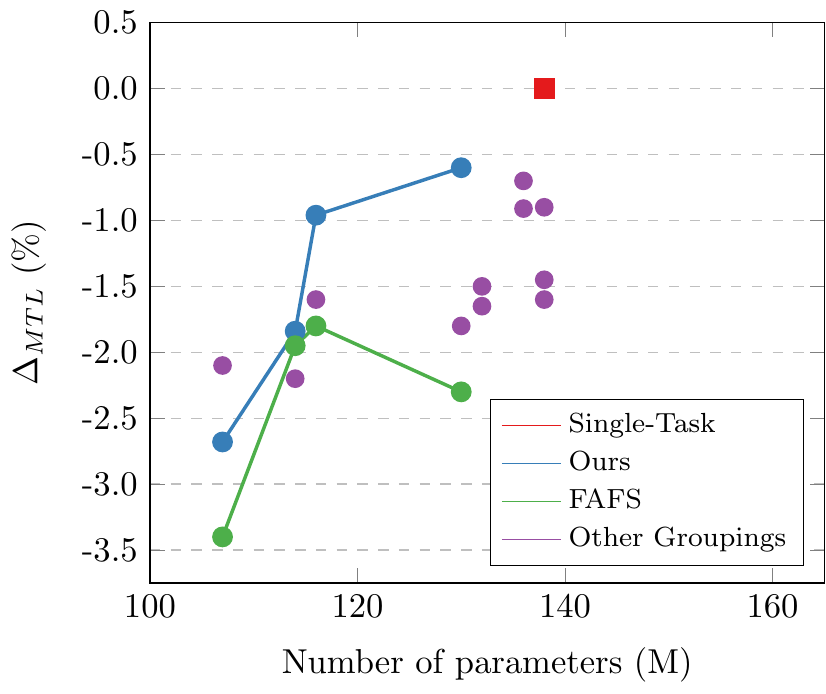}
  \end{center}
  \caption[Results for branched multi-task networks on Cityscapes]{Number of parameters versus multi-task performance on Cityscapes for different task groupings. The 'Other Groupings' contain any remaining tree structures that can be found by randomly branching the model in the last three ResNet blocks.}
  \label{fig: branched_cityscapes}
 \end{minipage}
 \end{figure}

\subsubsection{Results}
We measure the task affinity after every block (1 to 4) in the ResNet-50 model (see Figure~\ref{fig: branched_cityscapes_affinity}). The task affinity decreases in the deeper layers, due to the features becoming more task-specific. We compare the performance of the task groupings generated by our method with those by other approaches. 

We trained all possible task groupings that can be derived from branching the model in the last three ResNet blocks. Figure~\ref{fig: branched_cityscapes} visualizes performance versus number of parameters for the trained architectures. Depending on the available computational budget $\mathcal{C}$, our method generates a specific task grouping. We visualize these generated groupings as a blue path in Figure~\ref{fig: branched_cityscapes}, when gradually increasing the computational budget $\mathcal{C}$. Similarly, we consider the task groupings when branching the model based on the task affinity measure proposed by FAFS~\cite{lu2017fully} (green path). We find that, in comparison, the task groupings devised by our method achieve higher performance within a given computational budget $\mathcal{C}$. Furthermore, in the majority of cases, for a fixed budget $\mathcal{C}$ the proposed method is capable of selecting the best performing task grouping w.r.t. performance vs parameters metric (blue vs other).

\begin{figure}
\centering
\begin{subfigure}[t]{.33\linewidth}
  \centering
  \includegraphics[width=\textwidth]{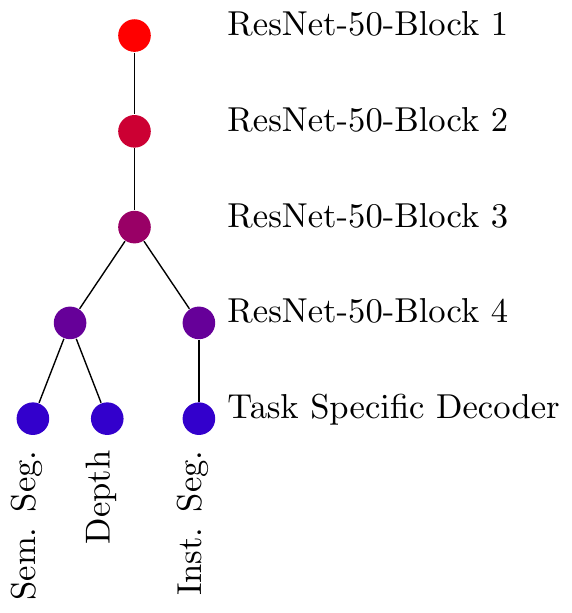}
  \caption{Ours - 1}
\end{subfigure}%
\begin{subfigure}[t]{.33\linewidth}
  \centering
  \includegraphics[width=\textwidth]{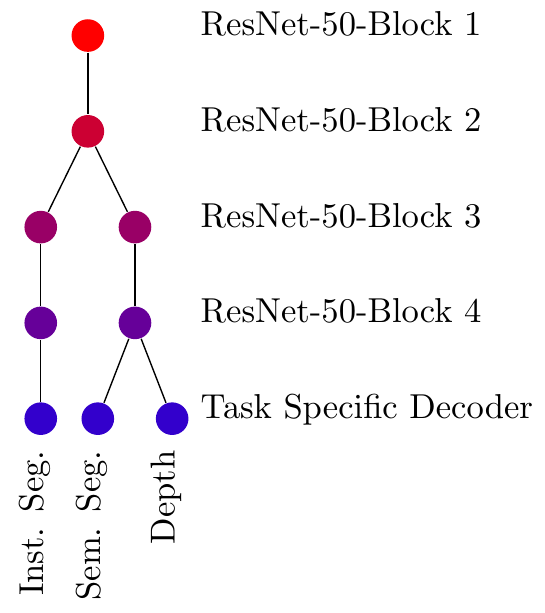}
  \caption{Ours - 2}
\end{subfigure}
\begin{subfigure}[t]{.33\linewidth}
  \centering
  \includegraphics[width=\textwidth]{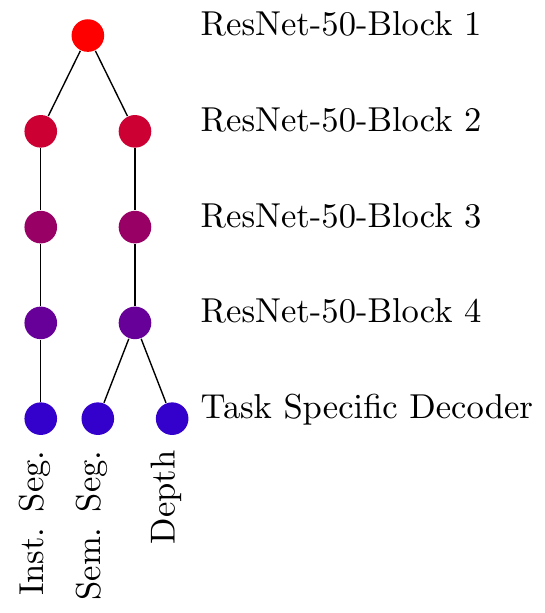}
  \caption{Ours - 3}
\end{subfigure}
\caption[Task groupings on Cityscapes]{Task groupings generated by our method on the Cityscapes dataset.}
\label{fig: our_tree_cityscapes}
\end{figure}

We also compare our branched multi-task networks with cross-stitch networks~\cite{misra2016cross}, NDDR-CNNs~\cite{gao2019nddr} and MTAN~\cite{liu2019end} in Table~\ref{tab: branched_cityscapes}. While cross-stitch nets and NDDR-CNNs give higher multi-task performance, attributed to their computationally expensive soft parameter sharing setting, our branched networks can strike a better trade-off between the performance and number of parameters. In particular, we can effectively sample architectures which lie between the extremes of a baseline multi-task model and a cross-stitch or NDDR-CNN architecture. Finally, our models provide a more computationally efficient alternative to the MTAN model, which reports similar performance while using more parameters. We visualize our task groupings from Table~\ref{tab: branched_cityscapes} in Figure~\ref{fig: our_tree_cityscapes}.

\begin{table}
    \caption[Results for branched multi-task networks on Cityscapes]{Results on the Cityscapes validation set.}
    \label{tab: branched_cityscapes}
    \centering
    \small{
    \begin{tabular}{l|ccccc}
    \toprule
    \textbf{Method} & \textbf{Sem. Seg.} & \textbf{Inst. Seg.} & \textbf{Disparity} & \textbf{\# Params} & $\mathbf{\Delta_{MTL}}$ \\
    & (IoU) $\uparrow$ & (px) $\downarrow$ & (px) $\downarrow$ & (M) $\downarrow$ & (\%)$\uparrow$ \\
    \midrule
    Single task & 65.2 & 11.7 & 2.57 & 138 & +0.00 \\
    MTL baseline & 61.5 & 11.8 & 2.66 & 92 & -3.33 \\
    MTAN & 62.8 & 11.8 & 2.66 & 113 & -2.53 \\
    Cross-stitch & 65.1 & 11.6 & 2.55 & 140 & +0.42 \\
    NDDR-CNN & 65.6 & 11.6 & 2.54 & 190 & +0.89 \\
    \midrule
    Ours - 1 & 62.1 & 11.7 & 2.66 & 107 & -2.68 \\
    Ours - 2 & 62.7 & 11.7 & 2.62 & 114 & -1.84 \\
    Ours - 3 & 64.1 & 11.6 & 2.62 & 116 & -0.96 \\
    \bottomrule
    \end{tabular}}
\end{table}

\subsection{Taskonomy}

\subsubsection{Setup}
We tackle the scene categorization (C), semantic segmentation (S), edge detection (E), monocular depth estimation (D) and keypoint detection (K) tasks. All input images were rescaled to 256 x 256 pixels. We use a ResNet-50 encoder and replace the last stride 2 convolution by a stride 1 convolution. A 15-layer fully-convolutional decoder is used for the pixel-to-pixel prediction tasks. The decoder is composed of five convolutional layers followed by alternating convolutional and transposed convolutional layers. We use ReLU as non-linearity. Batch normalization is included in every layer except for the output layer. We use Kaiming He's initialization for both encoder and decoder. We use an L1 loss for the depth (D), edge detection (E) and keypoint detection (K) tasks. The scene categorization task is learned with a KL-divergence loss. We report performance on the scene categorization task by measuring the overlap in top-5 classes between the predictions and ground truth.

The multi-task models were optimized with task weights $w_s = 1, w_d = 1, w_k = 10, w_e = 10$ and $w_c = 1$. The ground-truth heatmaps were linearly rescaled to lie between 0 and 1. During training we normalize the depth map by the standard deviation. 

\begin{table}
    \caption[Results for branched multi-task networks on Taskonomy]{Results on the tiny Taskonomy test set. The results for edge (E) and keypoints (K) detection were multiplied by a factor of 100 for better readability. The FAFS models refer to generating the task groupings with the task affinity technique proposed by~\cite{lu2017fully}.}
    \label{tab: branched_taskonomy}
    \small{
    \begin{tabular}{l|ccccccc}
    \toprule
    \textbf{Method} & \textbf{Depth} & \textbf{Sem. Seg} & \textbf{Categ.} & \textbf{Edge} & \textbf{Keyp.} & \textbf{\# Params} & $\mathbf{\Delta_{MTL}}$ \\
    & (L1) $\downarrow$ & (IoU) $\uparrow$ & (Top-5) $\uparrow$ & (L1) $\downarrow$ & (L1) $\downarrow$ & (M) $\downarrow$ & (\%) $\uparrow$ \\
    \midrule
    Single task & 0.60 & 43.5 & 66.0 & 0.99 & 0.23 & 224 & + 0.00 \\
    MTL baseline & 0.75 & 47.8 & 56.0 & 1.37 & 0.34 & 130 & - 22.50 \\
    MTAN & 0.71 & 43.8 & 59.6 & 1.86 & 0.40 & 158 & -37.36 \\
    Cross-stitch & 0.61 & 44.0 & 58.2 & 1.35 & 0.50 & 224 & - 32.29 \\
    NDDR-CNN & 0.66 & 45.9 & 64.5 & 1.05 & 0.45 & 258 & - 21.02 \\
    \midrule
    FAFS - 1 & 0.74 & 46.1 & 62.7 & 1.30 & 0.39 & 174 & - 24.5 \\
    FAFS - 2 & 0.80 & 39.9 & 62.4 & 1.68 & 0.52 & 188 & - 48.32 \\
    FAFS - 3 & 0.74 & 46.1 & 64.9 & 1.05 & 0.27 & 196 & - 8.48 \\
    \midrule
    Ours - 1 & 0.76 & 47.6 & 63.3 & 1.12 & 0.29 & 174 & - 11.88 \\
    Ours - 2 & 0.74 & 48.0 & 63.6 & 0.96 & 0.35 & 188 & - 12.66 \\
    Ours - 3 & 0.74 & 47.9 & 64.5 & 0.94 & 0.26 & 196 & - 4.93 \\
    \bottomrule
    \end{tabular}}
\end{table}

\subsubsection{Results}
\begin{figure}
\centering
\begin{subfigure}[t]{.33\linewidth}
  \centering
  \includegraphics[width=\textwidth]{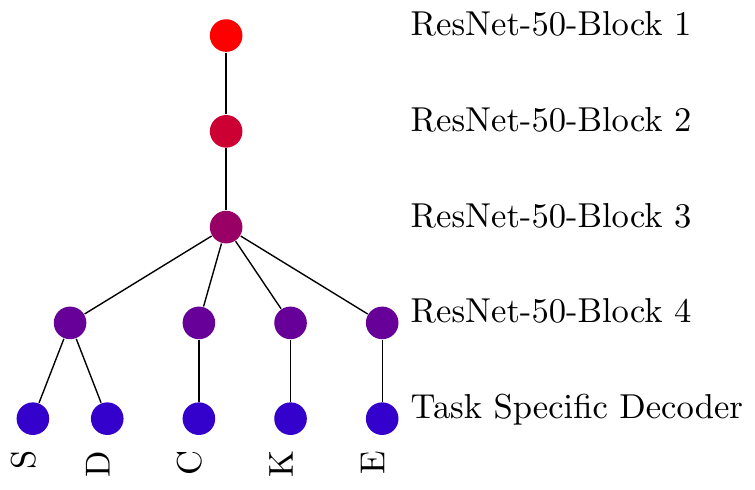}
  \caption{Ours - 1}
\end{subfigure}%
\begin{subfigure}[t]{.33\linewidth}
  \centering
  \includegraphics[width=\textwidth]{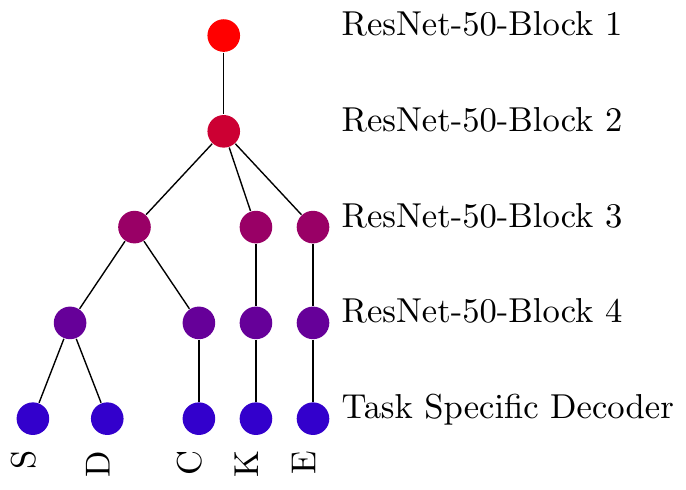}
  \caption{Ours - 2}
\end{subfigure}
\begin{subfigure}[t]{.33\linewidth}
  \centering
  \includegraphics[width=\textwidth]{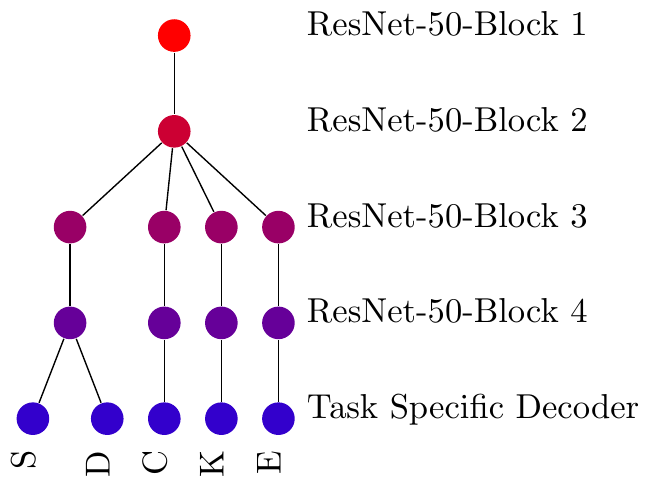}
  \caption{Ours - 3}
\end{subfigure}
\caption[Task groupings on Taskonomy]{Task groupings generated by our method on the Taskonomy dataset.}
\label{fig: our_tree_taskonomy}
\end{figure}

The task affinity is again measured after every ResNet block. Since the number of tasks increased to five, it is very expensive to train all task groupings exhaustively, as done above. Instead, we limit ourselves to three architectures that are generated when gradually increasing the parameter budget. As before, we compare our task groupings against the method from~\cite{lu2017fully}. The numerical results can be found in Table~\ref{tab: branched_taskonomy}. The task groupings themselves are shown in Figure~\ref{fig: our_tree_taskonomy}.

The effect of the employed task grouping technique can be seen from comparing the performance of our models against the corresponding FAFS models, generated by~\cite{lu2017fully}. The latter are consistently outperformed by our models. Compared to the results on Cityscapes (Figure~\ref{fig: branched_cityscapes}), we find that the multi-task performance is much more susceptible to the employed task groupings, possibly due to negative transfer. Furthermore, we observe that none of the soft parameter sharing models can handle the larger, more diverse task dictionary: the performance decreases when using these models, while the number of parameters increases. This is in contrast to our branched multi-task networks, which seem to handle the diverse set of tasks rather positively. As opposed to~\cite{zamir2018taskonomy}, but in accordance with~\cite{maninis2019attentive}, we show that it is possible to solve many heterogeneous tasks simultaneously when the negative transfer is limited, by separating dissimilar tasks from each other in our case. In fact, our approach is the first to show such consistent performance across different multi-tasking scenarios and datasets. Existing approaches seem to be tailored for particular cases, e.g. few/correlated tasks, synthetic-like data, binary classification only tasks, etc., whereas we show stable performance across the board of different experimental setups.

\subsection{CelebA}

\subsubsection{Setup}
The CelebA dataset~\cite{liu2015faceattributes} contains over 200k images of celebrities, labeled with 40 facial attribute categories. We treat the prediction of each facial attribute as a single binary classification task, as in~\cite{lu2017fully,sener2018multi,huang2018gnas}. To ensure a fair comparison: we reuse the thin-$\omega$ model from~\cite{lu2017fully} in our experiments; the parameter budget $\mathcal{C}$ is set for the model to have the same amount of parameters as prior work. The CNN architecture is based on the VGG-16 model~\cite{simonyan2015very}. The number of convolutional features is set to the minimum between $\omega$ and the width of the corresponding layer in the VGG-16 model. The fully connected layers contain $2 \cdot \omega$ features. The loss function is a sigmoid cross-entropy loss with uniform weighing scheme. 
\subsubsection{Results}
\begin{figure}[H]
\includegraphics[width=\linewidth]{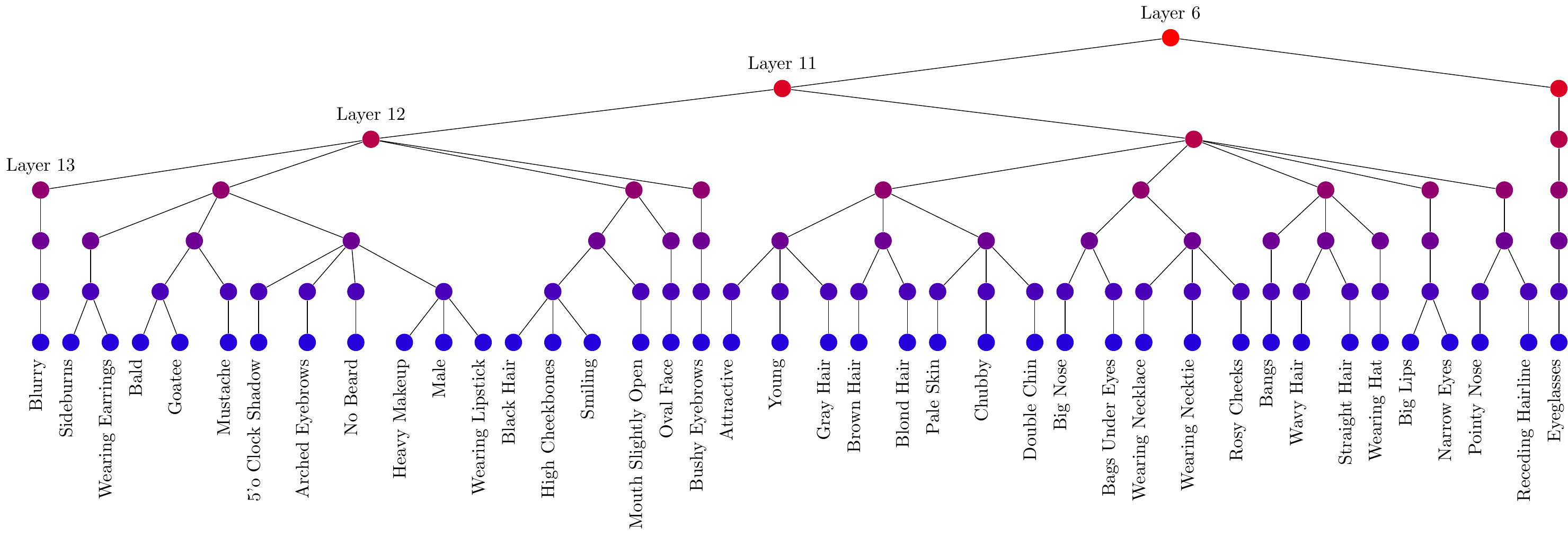}
\caption[Task groupings on CelebA]{Grouping of 40 person attribute classification tasks on CelebA in a thin VGG-16 architecture (Ours-32).}
\label{fig: tree_celeba}
\end{figure}

Table~\ref{tab: branched_celeba} shows the results on the CelebA test set. Our branched multi-task networks outperform earlier works~\cite{lu2017fully, huang2018gnas} when using a similar amount of parameters. Since the Ours-32 model (i.e. $\omega$ is 32) only differs from the FAFS model on the employed task grouping technique, we can conclude that the proposed method devises more effective task groupings for the attribute classification tasks on CelebA. Furthermore, the Ours-32 model performs on par with the VGG-16 model, while using 64 times less parameters. Finally, we show a visualization of the tree structure employed by the Ours-32 model in Figure~\ref{fig: tree_celeba}.

\begin{table}
    \caption[Results for branched multi-task networks on CelebA]{Results on the CelebA test set. The \textbf{Ours-32}, \textit{Ours-64} architectures are found by optimizing the task clustering for the parameter budget that is used in the \textbf{FAFS}, \textit{GNAS} model respectively.}
    \label{tab: branched_celeba}
    \centering
    \small{
    \begin{tabular}{l|cc}
    \toprule
    \textbf{Method} & \textbf{Acc.} (\%) & \textbf{\#Params} (M) $\downarrow$\\
    \midrule
    LNet+ANet \cite{wang2016walk} & 87 & - \\
    Walk and Learn \cite{wang2016walk} & 88 & - \\
    MOON \cite{rudd2016moon} & 90.94 & 119.73 \\
    Independent Group \cite{hand2017attributes} & 91.06 & - \\ 
    MCNN \cite{hand2017attributes} & 91.26 & - \\
    MCNN-AUX \cite{hand2017attributes} & 91.29 & - \\
    VGG-16 \cite{lu2017fully} & 91.44 & 134.41 \\
    \textbf{FAFS} \cite{lu2017fully} &  \textbf{90.79} & \textbf{2.09} \\
    \textit{GNAS} \cite{hand2017attributes} & \textit{91.63} & \textit{7.73} \\
    Res-18 (Uniform) \cite{sener2018multi} & 90.38 & 11.2 \\
    Res-18 (MGDA) \cite{sener2018multi} & 91.75 & 11.2 \\
    \midrule
    \textbf{Ours-32}& \textbf{91.46} & \textbf{2.20} \\
    \textit{Ours-64} & \textit{91.73} & \textit{7.73} \\
    \bottomrule
    \end{tabular}}
\end{table}

\subsection{Computational Analysis}
We provide an analysis to identify the computational costs related to the different steps when calculating the task affinity scores. The following three steps can be identified:

\begin{itemize}
    \item Train $N$ single task networks. It is possible to use a subset of the available training data to reduce the training time. We verified that using a random subset of 500 training images (1/5) on Cityscapes results in the same task groupings. 
    \item Compute the RDM matrix for all $N$ networks at $D$ pre-determined layers. This requires to compute the features for a set of $K$ images at the $D$ pre-determined layers in all $N$ networks. The $K$ images are usually taken as held-out images from the train set. We used $K =500$ in our experiments. In practice this means that computing the image features comes down to evaluating every model on $K$ images. The computed features are stored on disk afterwards. The RDM matrices are calculated from the stored features. This requires to calculate $N \times D \times K \times K$ correlations between two feature vectors (can be performed in parallel). In practice, the computation time is negligible compared to training the single-task networks. 
    \item Compute the RSA matrix at $D$ locations for all $N$ tasks. This requires to calculate $D \times N \times N$ correlations between the lower triangle part of the $K \times K$ RDM matrices. Again, the computation time can be neglected. 
\end{itemize}

In conclusion, the computational cost of our method boils down to training $N$ single task networks plus some small overhead. Notice that cross-stitch networks~\cite{misra2016cross} and NDDR-CNNs~\cite{gao2019nddr} also pre-train a set of single-task networks first, before combining them together using a soft parameter sharing mechanism. Differently, our method only requires to use a subset of the training images. We conclude that our method does not suffer from overhead compared to these methods.  

\section{Conclusion}
\label{sec: branched_conclusion}
We introduced a principled approach to automatically construct branched multi-task networks for a given computational budget. To this end, we leverage the employed tasks' affinities as a quantifiable measure for layer sharing. The proposed approach can be seen as an abstraction of NAS for MTL, where only layer sharing is optimized, without having to jointly optimize the layers types, their connectivity, etc., as done in traditional NAS, which would render the problem considerably expensive. Extensive experimental analysis shows that our method outperforms existing ones w.r.t. the important metric of multi-tasking performance vs number of parameters, while at the same time showing consistent results across a diverse set of multi-tasking scenarios and datasets.  

\paragraph{Limitations} Branched Multi-Task Networks allow to strike a better trade-off between the performance and the used amount of computational resources. This allows to significantly reduce the size of the model w.r.t. single-task learning. Unfortunately, we do not observe similar gains for the performance. Therefore, in the next chapter, we introduce a novel architecture specifically designed for tackling multiple dense prediction tasks. The resulting model does deliver on the full promise of multi-task learning, i.e. improved performance, smaller memory footprint and reduced number of FLOPS. 

\paragraph{Extensions/Follow-Up Work} Different from our work, \textbf{Branched Multi-Task Architecture Search}~\cite{bruggemann2020automated} and \textbf{Learning To Branch}~\cite{guo2020learning} have directly optimized the network topology without relying on pre-computed task relatedness scores. More specifically, they rely on a tree-structured network design space where the branching points are casted as a Gumbel softmax operation. This strategy has the advantage over our work that the task groupings can be directly optimized end-to-end for the tasks under consideration. Moreover, both methods can easily be applied to any set of tasks, including both image classification and per-pixel prediction tasks. Similar to~\cite{lu2017fully,vandenhende2019branched}, a compact network topology can be obtained by including a resource-aware loss term. In this case, the computational budget is jointly optimized with the multi-task learning objective in an end-to-end fashion.

\cleardoublepage%

  \graphicspath{{\chaptersdir/mtinet/image/}}%
\chapter{Multi-Scale Multi-Task Interaction Networks}
\label{ch: mti_net} 
In this chapter, we revisit the decoder-focused approaches. First, we argue about the importance of considering task interactions at multiple scales when distilling task information in a multi-task learning setup. In contrast to common belief, our experimental analysis shows that tasks with high affinity at a certain scale are not guaranteed to retain this behavior at other scales, and vice versa. 

Next, we propose a novel architecture, namely MTI-Net, that builds upon our findings in three ways. First, MTI-Net explicitly models task interactions at every scale via a multi-scale multi-modal distillation unit. Second, MTI-Net propagates distilled task information from lower to higher scales via a feature propagation module. Third, MTI-Net aggregates the refined task features from all scales via a feature aggregation unit to produce the final per-task predictions. 

Finally, extensive experiments on two multi-task dense labeling datasets show that, unlike prior work, our multi-task model delivers on the full potential of multi-task learning, that is, smaller memory footprint, reduced number of calculations, and better performance w.r.t. single-task learning. 

The content of this chapter is based upon the following publication:
\begin{itemize}
    \item \emph{Vandenhende, S., Georgoulis, S., \& Van Gool, L. (2020). MTI-Net: Multi-scale task interaction networks for multi-task learning. European Conference on Computer Vision.}
\end{itemize}

\section{Motivation}
\label{sec: mti_motivation}
First, we introduce the necessary background materials. Next, we motivate the use of a decoder-focused model that distills task information at multiple scales.

\subsection{Decoder-Focused Approaches}
Decoder-focused approaches~\cite{xu2018pad,zhang2018joint,zhang2019pattern,zhou2020pattern} try to explicitly distill information from other tasks, as a complementary signal to improve task performance. Typically, this is achieved by combining an existing backbone network with a multi-step decoding process. 

\begin{figure*}[t]
    \centering
    \includegraphics[width=.8\textwidth]{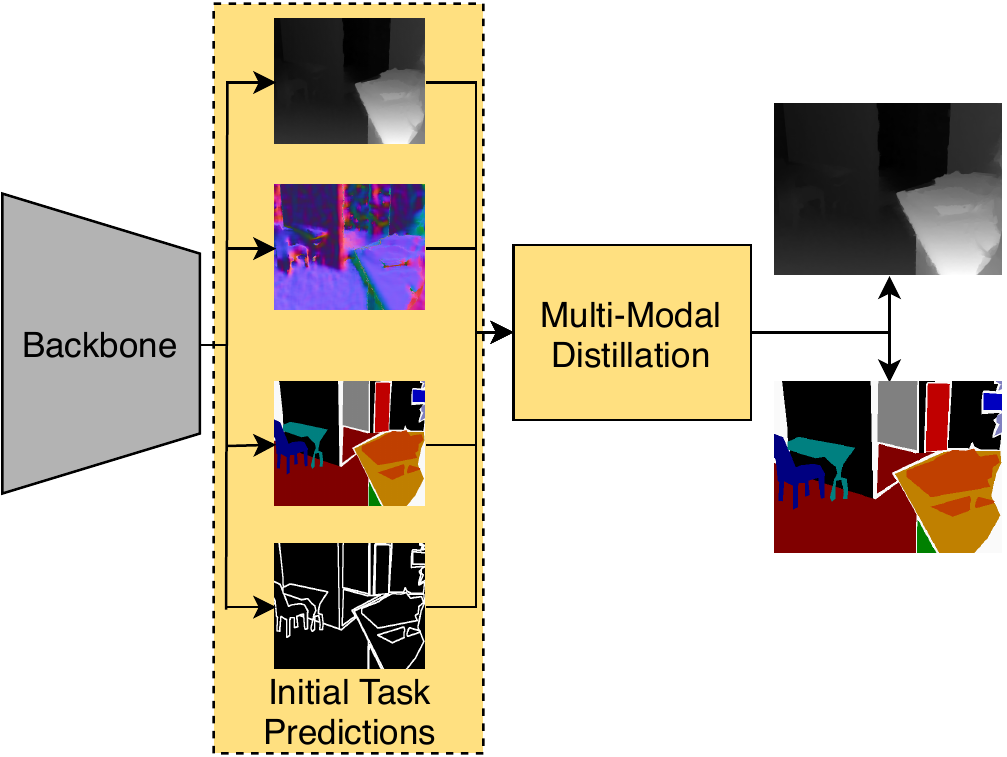}
    \caption[PAD-Net]{An overview of the architecture used in PAD-Net~\cite{xu2018pad} and PAP-Net~\cite{zhang2019pattern}. Features extracted from a backbone network are used to make initial task predictions. The task features are combined through a distillation unit before making the final task predictions.}
    \label{fig: mti_padnet}
\end{figure*}

In more detail, the shared features of the backbone network are processed by a set of task-specific heads, that produce an initial prediction for every task. We further refer to the backbone and the task-specific heads as the \textit{front-end} of the network. The task-specific heads produce a per-task feature representation of the scene that is more task-aware than the shared features of the backbone network. The information from these task-specific feature representations is then combined via a multi-modal distillation unit, before making the final task predictions. As shown in Figure~\ref{fig: mti_padnet}, it is possible that some tasks are only predicted in the front-end of the network. The latter are known as auxiliary tasks, since they serve as proxies in order to improve the performance on the final tasks. We refer the interested reader to Section~\ref{subsec: related_decoder} for an in-depth discussion on existing decoder-focused approaches. 

For brevity, we adopt the following notations. 
\textit{Backbone features}: the shared features (at the last layer) of the backbone network; 
\textit{Task features}: the task-specific feature representations (at the last layer) of each task-specific head; 
\textit{Distilled task features}: the task features after multi-modal distillation; 
\textit{Initial task predictions}: the per-task predictions at the front-end of the network; 
\textit{Final task predictions}: the network outputs. 
Note that, backbone features, task features and distilled task features can be defined at a single scale or multiple scales.

\subsection{Task interactions at different scales}
\label{subsec: multi_scale_task_interactions}
\begin{figure}
\begin{minipage}[t]{.39\linewidth}
    \centering
    \includegraphics[width=\textwidth]{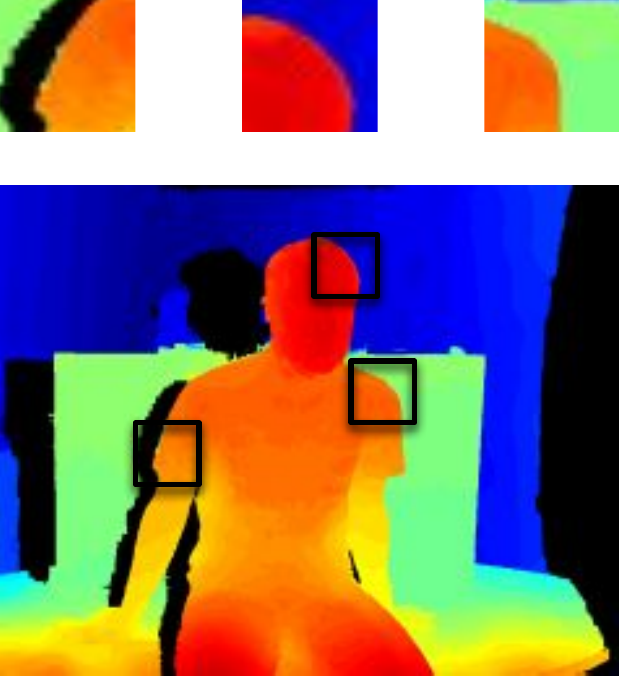}
    \caption[Task interactions change with scale - qualitative analysis]{Three local patches from a depth map. Depending on the patch size, i.e. receptive field, depth information can be utilized differently by other tasks, e.g. semantic segmentation and edges.}
    \label{fig: intuitive}
\end{minipage}
\hfill
\begin{minipage}[t]{.53\linewidth}
    \centering
    \includegraphics[width=\textwidth]{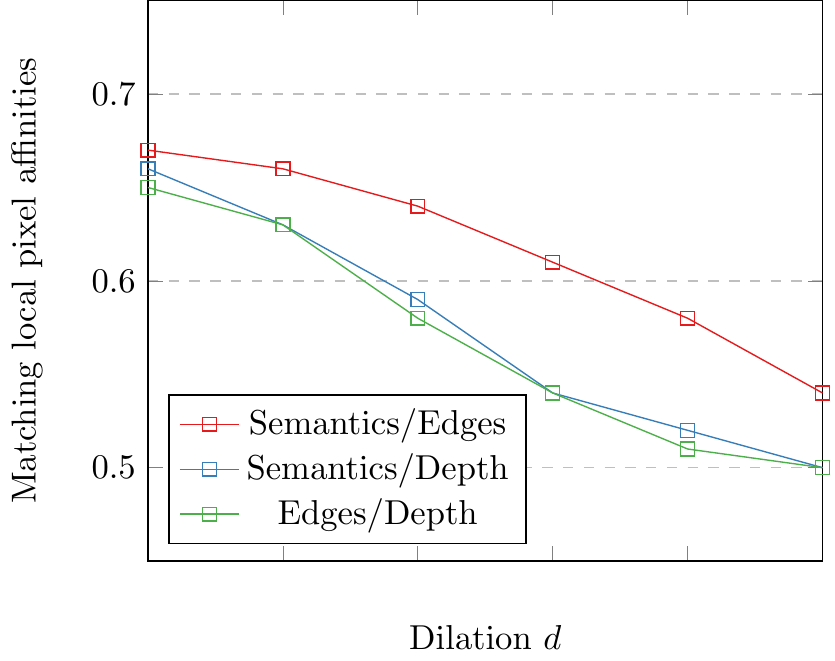}
    \caption[Task interactions change with scale - quantitative analysis]{To quantify task interactions w.r.t. scale, pixel affinities on the label space of each task, as defined in~\cite{zhang2019pattern}, are calculated in local patches. The correspondences in the affinity patterns between tasks are plotted as a function of the patch size, i.e. kernel dilation.}
    \label{fig: affinity}
\end{minipage}
\end{figure}

Existing decoder-focused approaches follow a common pattern: they perform multi-modal distillation at a fixed scale, i.e. the backbone features. This rests on the assumption that all relevant task interactions can solely be modeled through a single filter operation with specific receptive field. For all we know, this is not always the case. In fact, tasks can influence each other differently for different receptive field sizes. Consider, for example, Figure~\ref{fig: intuitive}. The local patches in the depth map provide little information about the semantics of the scene. However, when we enlarge the receptive field, the depth map reveals a person's shape, hinting at the scene's semantics. Note that the local patches can still provide valuable information, e.g. to improve the local alignment of edges between tasks. 

To quantify the degree to which tasks share common local structures w.r.t. the size of the receptive field, we conduct the following experiment. We measure the pixel affinity in local patches on the label space of each task, using kernels of fixed size. The size of the receptive field can be selected by choosing the dilation for the kernel. We consider the tasks of semantic segmentation, depth estimation and edge detection on the NYUD dataset. A pair of semantic pixels is considered similar when both pixels belong to the same category. For the depth estimation task, we threshold the relative difference between pairs of pixels; pixels below the threshold are similar. Once the pixel affinities are calculated for every task, we measure how well similar and dissimilar pairs are matched across tasks. We repeat this experiment using different dilations for the kernel, effectively changing the receptive field. Figure~\ref{fig: affinity} illustrates the result.

A first observation is that affinity patterns are matched well across tasks, with up to $65\%$ of pair correspondence in some cases. This indicates that different tasks can share common structures in parts of the image. A second observation is that the degree to which the affinity patterns are matched across tasks is dependent on the receptive field, which in turn, corresponds to the used dilation. This validates our initial assumption that the statistics of task interactions do not always remain constant, but rather depend on the scale, i.e. receptive field. 
Based on these findings, in the next section we introduce a model that distills information from different tasks at multiple scales. By doing so, we are able to capture the unique task interactions at each individual scale, overcoming the limitations of existing decoder-focused models.

\section{Method}
\label{sec: mti_method}

\subsection{Multi-scale multi-modal distillation}
\label{subsec: multi_scale_distillation}
\begin{figure}[t]
\centering
\includegraphics[width=.9\linewidth]{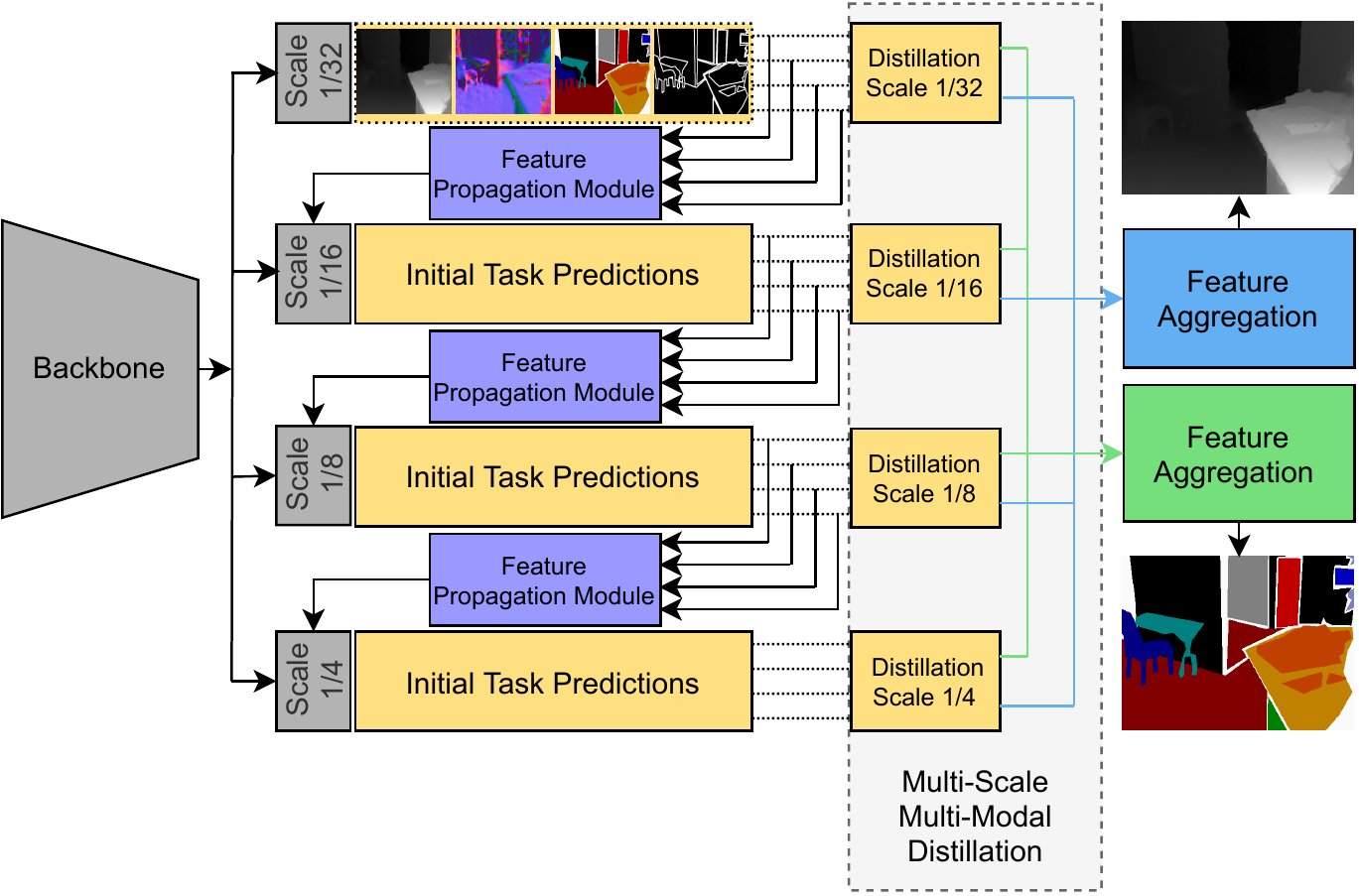}
\caption[MTI-Net]{\textbf{Overview of the proposed MTI-Net}. Starting from a backbone that extracts multi-scale features, initial task predictions are made at each scale. The task features are distilled separately at every scale, allowing our model to capture task interactions at multiple scales, i.e. receptive fields. After distillation, the distilled task features from all scales are aggregated to make the final task predictions. To boost performance, we extend our model with a feature propagation mechanism that passes distilled information from lower resolution task features to higher ones.}
\label{fig: mti_method}
\end{figure}

We propose a multi-task architecture that explicitly takes into account task interactions at multiple scales. Figure~\ref{fig: mti_method} shows a visualization of our model. First, an off-the-shelf backbone network extracts a multi-scale feature representation from the input image. Such multi-scale feature extractors have been used in semantic segmentation~\cite{ronneberger2015u,wang2019deep,kirillov2019panoptic}, object detection~\cite{lin2017feature,wang2019deep}, pose estimation~\cite{newell2016stacked,sun2019deep}, etc. In Section~\ref{sec: mti_experiments} we verify our approach using two such backbones, i.e. HRNet~\cite{wang2019deep} and FPN~\cite{lin2017feature}, but any multi-scale feature extractor can be used instead.

From the multi-scale feature representation (i.e. backbone features) we make initial task predictions at each scale. These initial task predictions at a particular scale are found by applying a set of task-specific heads to the backbone features extracted at that scale. The result is a per-task representation of the scene (i.e. task features) at a multitude of scales. Not only does this add deep supervision to our network, but the task features can now be distilled at each scale separately. This allows us to have multiple task interactions, each modeled for a specific receptive field size, as proposed in Section~\ref{subsec: multi_scale_task_interactions}.

Next, we refine the task features by distilling information from the other tasks using a spatial attention mechanism~\cite{xu2018pad}. Yet, our multi-modal distillation process is repeated at each scale, i.e. we apply multi-scale, multi-modal distillation. The distilled task features $F_{k, s}^{o}$ for task $k$ at scale $s$ are found as:
\begin{equation}
    F_{k,s}^{o} = F_{k,s}^{i} + \sum_{l \neq k} \sigma \left( W_{k,l,s} F_{l,s}^{i} \right) \odot \left( W_{k,l,s}^{'} F_{l,s}^{i} \right),
\end{equation}
where $\sigma \left( W_{k,l,s} F^{i}_{l,s} \right)$ returns a per-scale spatial attention mask, that is applied to the task features $F_{l,s}^{i}$ from task $l$ at scale $s$. Note that our approach is not necessarily limited to the use of spatial attention, but any type of feature distillation (e.g. squeeze-and-excitation~\cite{hu2018squeeze}) can easily be plugged in. Through repetition, we calculate the distilled task features at every scale. As the bulk of filter operations is performed on low resolution feature maps, the computational overhead of our model is limited. We make a detailed resource analysis in Section~\ref{sec: mti_experiments}.

\subsection{Feature propagation across scales}
\label{subsec: feature_propagation}
In Section~\ref{subsec: multi_scale_distillation} actions at each scale were performed in isolation. To sum up, we made initial task predictions at each scale, from which we refined the task features through multi-modal distillation at each individual scale separately. However, as the higher resolution scales have a limited receptive field, the front-end of the network could have a hard time to make good initial task predictions at these scales, which in turn, would lead to low quality task features there. To remedy this situation we introduce a feature propagation mechanism, where the backbone features of a higher resolution scale are concatenated with the task features from the preceding lower resolution scale, before feeding them to the task-specific heads of the higher resolution scale to get the task features there.

\begin{figure}[t]
\centering
\includegraphics[width=.9\linewidth]{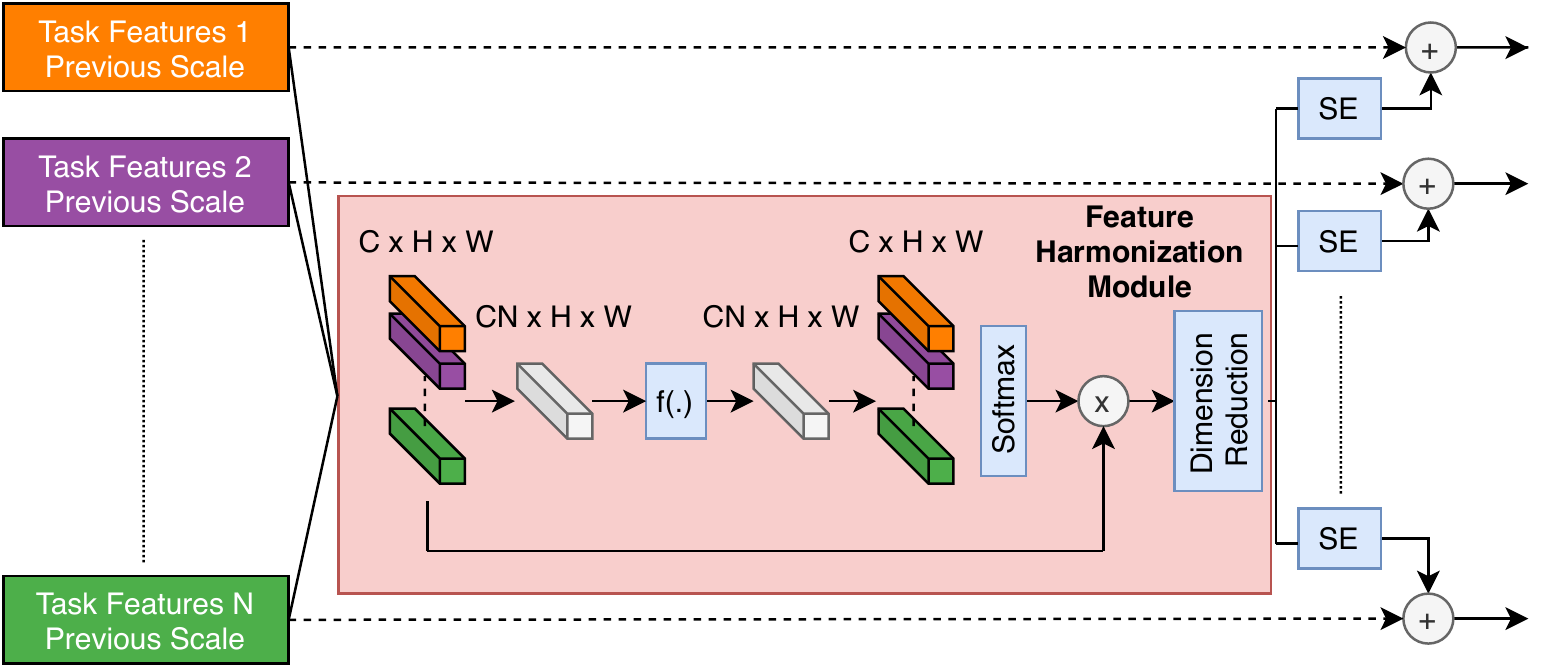}
\caption[Feature Propagation Module in MTI-Net]{\textbf{Overview of the proposed Feature Propagation Module (FPM).} First, task features from a lower scale are concatenated and mapped to a shared representation by the feature harmonization module. The task features are then refined by extracting information from the shared representation through squeeze-and-excitation (SE)~\cite{hu2018squeeze}, and are added as a residual to the original ones. Finally, these refined task features will be concatenated with the backbone features of the preceding higher scale.}
\label{fig: feature_propagation}
\end{figure}

A trivial implementation for our Feature Propagation Module (FPM) would be to just upsample the task features from the previous scale and pass them to the next scale. We opt for a different approach however, and design the FPM to behave similarly to the multi-modal distillation unit of Section~\ref{subsec: multi_scale_distillation}, in order to model task interactions at this stage too. Figure~\ref{fig: feature_propagation} gives an overview of our FPM. We first use a \textit{feature harmonization} block to combine the task features from the previous scale to a shared representation. We then use this shared representation to refine the task features from the previous scale, before passing them to the next scale. The refinement happens by selecting relevant information from the shared representation through a \textit{squeeze-and-excitation} block~\cite{hu2018squeeze}. Note that, since we refine the features from a single shared representation, instead of processing each task independently as done in the multi-modal distillation unit of Section~\ref{subsec: multi_scale_distillation}, the computational cost is significantly smaller.

\noindent\textbf{Feature harmonization}.
The FPM receives as input the task features from $N$ tasks of shape $C \times H \times W$. Our feature harmonization module combines the received task features into a shared representation. In particular, the set of $N$ task features is first concatenated and processed by a learnable non-linear function $f$. The output is split into $N$ chunks along the channel dimension, that match the original number of channels $C$. We then apply a softmax function along the task dimension to generate a task attention mask. The attended features are concatenated and further processed to reduce the number of channels from $N \cdot C$ to $C$. The output is a shared representation based on information from all tasks. 

\noindent\textbf{Refinement through Squeeze-And-Excitation}.
The use of a shared representation can degrade performance when tasks are unrelated. We resolve this situation by applying a per-task channel gating function to the shared representation. This effectively allows each task to select the relevant features from the shared representation. The channel gating mechanism is implemented here as a squeeze-and-excitation (SE) block~\cite{hu2018squeeze}. This is due to the fact that SE has shown great potential in MTL (e.g.~\cite{maninis2019attentive}), yet other gating mechanisms could be used instead. After applying the SE module, the refined task features are added as a residual to the original task features. 

\subsection{Feature aggregation}
\label{subsec: feature_aggregation}
The multi-scale, multi-modal distillation described in Section~\ref{subsec: multi_scale_distillation} results in distilled task features at every scale. The latter are upsampled to the highest scale and concatenated, resulting in a final feature representation for every task. The final task predictions are found by decoding these final feature representations by task-specific heads again. All implementation details are discussed in Section~\ref{sec: mti_experiments}. It is worth mentioning that our model has the possibility to add auxiliary tasks in the front-end of the network, similar to PAD-Net~\cite{xu2018pad}. In our case however, the auxiliary tasks are predicted at multiple scales. 

\subsection{Optimization}
The model is trained in an end-to-end fashion. Assume we make initial task predictions at scales $\mathcal{S}$. Further, let $\mathcal{T}$ denote the set of main tasks and $\mathcal{A}$ denote the set of auxiliary tasks. We use the notation $\mathcal{L}_{t}^{s}$ for the loss of the initial predictions of task $t$ at scale $s$, and $\mathcal{L}_{t}^{o}$ for the loss of the final predictions of task $t$. Further, we define the task-specific weight for task $t$ as $w_t$. Note, we adopt the same task-specific weights for the initial and final task predictions. The optimization objective can be written as
\begin{equation}
    \mathcal{L}_{total} = \sum_{t \in \mathcal{T}}  \sum_{s \in \mathcal{S}}  w_t \cdot \mathcal{L}_t^s + \sum_{a \in \mathcal{A}} \sum_{s \in \mathcal{S}}  w_a \cdot \mathcal{L}_{a}^{s} + \sum_{t \in \mathcal{T}} w_t \cdot \mathcal{L}_{t}^{o}.
\end{equation}

\section{Experiments}
\label{sec: mti_experiments}

\subsection{Setup}
\label{subsec: mti_setup}

\subsubsection{Datasets \& Implementation Details}
We perform our experimental evaluation on the PASCAL~\cite{everingham2010pascal} and NYUD~\cite{silberman2012indoor} datasets (see Section~\ref{sec: datasets}). We build our approach on top of two different backbone networks, i.e. FPN~\cite{lin2017feature} and HRNet~\cite{sun2019deep}. We use the different output scales of the selected backbone networks to perform multi-scale operations. This translates to four scales (1/4, 1/8, 1/16, 1/32). The task-specific heads are implemented as two basic residual blocks~\cite{he2016deep}. All our experiments are conducted with pre-trained ImageNet weights. 

We use the L1 loss for depth estimation and the cross-entropy loss for semantic segmentation on NYUD. As in prior work~\cite{kokkinos2015pushing,maninis2017convolutional,maninis2019attentive}, the edge detection task is trained with a positively weighted $w_{pos} = 0.95$ binary cross-entropy loss. We do not adopt a particular loss weighing strategy on NYUD, but simply sum the losses together. On PASCAL, we reuse the training setup from~\cite{maninis2019attentive} to facilitate a fair comparison. We reuse the loss weights from there. The initial task predictions in the front-end of the network use the same loss weighing as the final task predictions. In contrast to~\cite{xu2018pad,zhang2019pattern,zhang2018joint}, we do not use a two-step training procedure where the front-end is pre-trained separately. Instead, we simply train the complete architecture end-to-end.

\subsubsection{Evaluation Metrics}
We evaluate the performance of the backbone networks on the single tasks first. The optimal dataset F-measure (\textit{odsF})~\cite{martin2004learning} is used to evaluate the edge detection task. The semantic segmentation, saliency estimation and human part segmentation tasks are evaluated using mean intersection over union (\textit{mIoU}). We use the mean error (\textit{mErr}) in the predicted angles to evaluate the surface normals. The depth estimation task is evaluated using the root mean square error (\textit{rmse}). The multi-task learning performance $\Delta_{MTL}$ is defined in Equation~\ref{eq: evaluation_metric}. 

\subsubsection{Baselines}
On NYUD, we compare MTI-Net against the state-of-the-art PAD-Net~\cite{xu2018pad}. PAD-Net was originally designed for a single scale, but it is easy to plug-in a multi-scale backbone network and directly compare the two approaches. In contrast, a comparison with~\cite{zhang2018joint} is not possible, as this work was strictly designed for a pair of tasks, without any straightforward extension to the MTL setting. Finally, PAP-Net~\cite{zhang2019pattern} adopts an architecture that is similar to PAD-Net, but the multi-modal distillation is performed recursively on the feature affinities. We chose to draw the comparison with the more generic PAD-Net, since it performs on par with PAP-Net (see Section~\ref{subsec: mti_sota}). 

On PASCAL, we compare our method against the state-of-the-art ASTMT~\cite{maninis2019attentive}. Note that a direct comparison with ASTMT is also not straightforward, as this model is by design single-scale and heavily based on a DeepLab-v3+ (DLv3+) backbone network that contains dilated convolutions. Due to the latter, simply plugging the same DLv3+ backbone into MTI-Net would break the multi-scale features required to uniquely model the task interactions at a multitude of scales. Yet, we provide a fair comparison with ASTMT by combining it with a ResNet-50 FPN backbone, to show that it is not just using a multi-scale backbone that leads to improved results.

\subsection{Ablation Studies}
\label{subsec: mti_ablation}

\subsubsection{Network components}
In Table~\ref{tab: mti_ablation} we visualize the results of our ablation studies on NYUD and PASCAL with an HRNet18 backbone to verify how different components of our model contribute to the multi-task improvements. 

\begin{table}[t]
    \caption[Ablation study of MTI-Net]{Ablation studies on (a) NYUD and (b) PASCAL using an HRNet-18 backbone network. Auxiliary tasks are indicated between brackets.}
    \label{tab: mti_ablation}
    \begin{subtable}[t]{\linewidth}
    \caption{Results on NYUD.}
    \label{tab: mti_nyu_ablation}
    \centering
    \small{
    \begin{tabular}{l|ccc}
    \toprule
    \textbf{Method} & \textbf{Seg} $\uparrow$ & \textbf{Depth} $\downarrow$ & $\mathbf{\Delta_{m}\%} \uparrow$ \\
    \midrule
    Single task & 33.18 & 0.667 & + 0.00 \\
    MTL & 32.09 & 0.668 & - 1.71 \\
    PAD-Net & 32.80 & 0.660 & - 0.02 \\
    PAD-Net (N) & 33.85 & 0.658 & + 1.65 \\
    PAD-Net (N+E) & 32.92 & 0.655 & + 0.52 \\
    \midrule
    Ours (w/o FPM) & 34.38 & 0.640 & + 3.85 \\
    Ours (w/o FPM) (N) & 34.49 & 0.642 & + 3.84 \\
    Ours (w/o FPM) (N+E) & 34.68 & 0.637 & + 4.48 \\
    \midrule
    Ours (w/ FPM) & 35.12 & 0.620 & + 6.40 \\
    Ours (w/ FPM) (N) & 36.22 & \textbf{0.600} & + 9.57 \\
    Ours (w/ FPM) (N+E) & \textbf{37.49} & 0.607 & \textbf{+ 10.91} \\ 
    \bottomrule
    \multicolumn{4}{c}{}\\ 
    \end{tabular}} 
    \end{subtable} %
    \begin{subtable}[t]{\linewidth}
    \centering
    \caption{Results on PASCAL.}
    \label{tab: mti_pascal_ablation}
    \small{\begin{tabular}{l|cccccc}
    \toprule
    \textbf{Method} & \textbf{Seg} $\uparrow$ & \textbf{Parts} $\uparrow$ & \textbf{Sal} $\uparrow$ & \textbf{Edge} $\uparrow$ & \textbf{Norm} $\downarrow$ & $\mathbf{\Delta_{m} \%} \uparrow$ \\
    \midrule
    Single task & 60.07 & 60.74 & 67.18 & 69.70 & 14.59 & + 0.00 \\
    MTL (s) & 54.53 & 59.54 & 65.60 & - & - & - 4.26 \\
    MTL (a) & 53.60 & 58.45 & 65.13 & 70.60 & 15.08 & - 3.70 \\
    \midrule
    Ours (s) & 64.06 & 62.39 & 68.09 & - & - & + 3.35 \\
    Ours (s)(E) & 64.98 & 62.90 & 67.84 & - & - & + 3.98 \\
    Ours (s)(N) & 63.74 & 61.75 & 67.90 & - & - & + 2.69 \\
    Ours (s)(E+N) & 64.33 & 62.33 & 68.00 & - & - & + 3.36 \\
    Ours (a) & 64.27 & 62.06 & 68.00 & 73.40 & 14.75 & + 2.74 \\
    \bottomrule
    \end{tabular}}
    \end{subtable}%
\end{table}

We focus on the smaller NYUD dataset first (see Table~\ref{tab: mti_nyu_ablation}), that contains arguably related tasks. These are semantic segmentation (Seg) and depth prediction (Depth) as main tasks, edge detection (E) and surface normals (N) as auxiliary tasks. The MTL baseline (i.e. a shared encoder with task-specific heads) has lower performance ($-1.71\%$) than the single-task models. This is inline with prior work~\cite{vandenhende2019branched,maninis2019attentive}. PAD-Net retains performance over the set of single-task models ($-0.02\%$), and improves when adding the auxiliary tasks ($+0.52\%$). Using our model without the FPM between scales further improves the results (w/o auxiliary tasks: $+3.85\%$, w/ auxiliary tasks: $+4.48\%$). When including the FPM another significant boost in performance is achieved ($+6.40\%$). Further adding the auxiliary tasks can help to improve the quality of our predictions ($+10.91\%$).

\begin{figure}
\centering
\includegraphics[width=\textwidth]{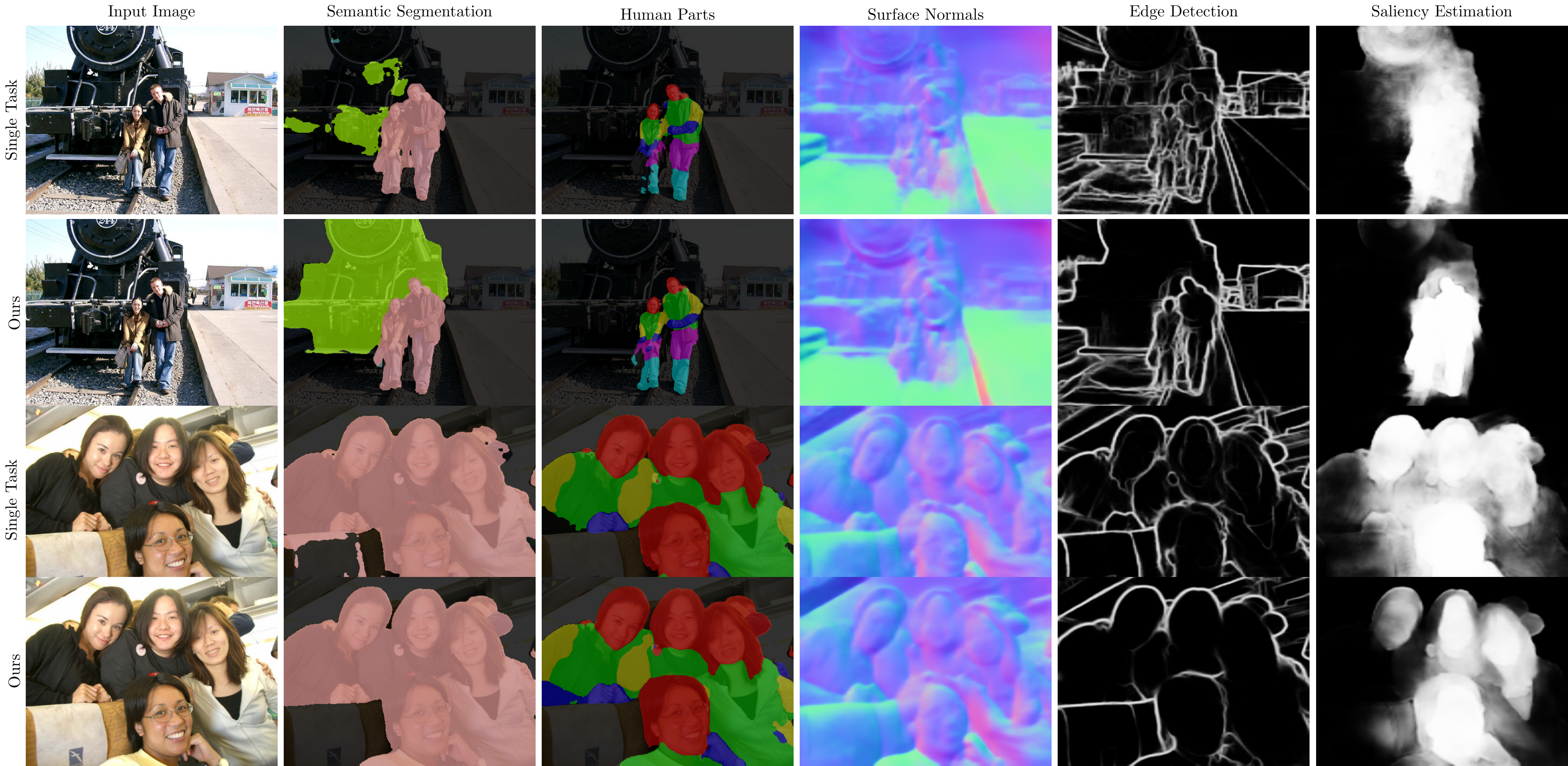}
\caption[Qualitative results on PASCAL]{\textbf{Qualitative results on PASCAL.} We compare the predictions made by a set of single-task models (first row for every image) against the predictions made by our MTI-Net (second row for every image). Differences can be seen for semantic segmentation, edge detection and saliency estimation.}
\label{fig: mti_results_pascal}
\end{figure}

Table~\ref{tab: mti_pascal_ablation} shows the ablation on PASCAL. We discriminate between a \textit{small set (s)} and a \textit{complete set (a)} of tasks. The small set contains the high-level (semantic and human parts segmentation) and mid-level (saliency) vision tasks. The complete set also adds the low-level (edges and normals) vision tasks. The MTL baseline leads to decreased performance, $-4.26\%$ and $-3.70\%$ on the small and complete set respectively. Instead, our model improves over the set of single-task models ($+ 3.35\%$) on the small task set (s), where we obtain solid improvements on all tasks. We also report the influence of adding additional auxiliary tasks to the front-end of the network. Adding edges improves the multi-task performance to $3.98\%$, adding normals slightly decreases it to $+2.69\%$, while adding both keeps it stable $+3.36\%$. Finally, when learning all five tasks together, our model outperforms ($+2.74\%$) the set of single-task models. In general, all tasks gain significantly, except for normals, where we observe a small decrease in performance. We argue that this is due to the inevitable negative transfer that characterizes all models with shared operations (also~\cite{xu2018pad,maninis2019attentive,zhang2019pattern}). Yet, to the best of our knowledge, this is the first work to not only report overall improved multi-task performance, but also to maximize the gains over the single-task models, when jointly predicting an increasing and diverse set of tasks. We refer to Figure~\ref{fig: mti_results_pascal} for qualitative results obtained with an HRNet-18 backbone.

\subsubsection{Influence of scales}
So far, our experiments included all four scales of the backbone network (1/4, 1/8, 1/16, 1/32). Here, we study the influence of using a different number of scales for the backbone. Table~\ref{tab: mti_nyu_scales} summarizes this ablation on NYUD. Note that the use of a single scale (1/4) reduces our model to a PAD-Net like architecture. Using an increasing number of scales (1/4 vs + 1/8 vs + 1/16, ...) gradually improves performance. The results confirm our hypothesis from Section~\ref{subsec: multi_scale_task_interactions}, i.e. task interactions should be modeled at multiple scales.

\begin{table}
    \caption[Ablation study of the number of scales]{Influence of using a different number of scales for the backbone network on NYUD.}
    \label{tab: mti_nyu_scales}
    \centering
    \small{
    \begin{tabular}{l|ccc}
    \toprule
    \textbf{Method} & \textbf{Seg} $\uparrow$ & \textbf{Depth} $\downarrow$ & $\mathbf{\Delta_{m} \%} \uparrow$ \\
    \midrule
    ST & 33.18 & 0.667 & + 0.00 \\
    1/4 (Pad-Net) & 32.80 & 0.660 & - 0.02 \\
    1/4, 1/8 & 34.88 & 0.650 & + 3.80 \\
    1/4, 1/8, 1/16 & 35.01 & 0.630 & + 5.53 \\
    1/4, 1/8, 1/16, 1/32 (Ours) & 35.12 & 0.620 & + 6.40 \\
    \bottomrule
    \end{tabular}}
\end{table}

\subsubsection{Information flow}
To quantify the flow of information, we measure the performance of the initial task predictions at different locations in the front-end of the network. Table~\ref{tab: mti_nyu_frontend} illustrates the results on NYUD. We observe that the performance gradually increases at the higher scales, due to the information that is being propagated from the lower scales via the FPM. The final prediction after aggregating the information from all scales is further improved substantially.

\begin{table}
    \caption[Ablation study of the information flow in MTI-Net]{Ablating the information flow within the proposed MTI-Net model on NYUD.}
    \label{tab: mti_nyu_frontend}
    \centering
    \small{
    \begin{tabular}{l|ccc}
    \toprule
    \textbf{Method} & \textbf{Seg} $\uparrow$ & \textbf{Depth} $\downarrow$ & $\mathbf{\Delta_{m} \%} \uparrow$ \\
    \midrule
    ST & 33.18 & 0.667 & + 0.00 \\
    Front-end @ 1/32 scale & 32.02 & 0.670 & - 1.87 \\
    Front-end @ 1/16 scale & 33.02 & 0.660 & + 0.02 \\
    Front-end @ 1/8 scale  & 33.67 & 0.640 & + 2.72 \\
    Front-end @ 1/4 scale & 34.05 & 0.633 & + 3.78 \\
    Final output & 35.12 & 0.620 & + 6.40 \\
    \bottomrule
    \end{tabular}}
\end{table}

\subsection{State-Of-The-Art Comparison}
\label{subsec: mti_sota}

\subsubsection{Comparison on PASCAL}
Table~\ref{tab: mti_sota_pascal} visualizes the comparison of our model against ASTMT and PAD-Net on PASCAL. We report the multi-tasking performance both w.r.t. the single-task models using the same backbone (ST) and the single-task models based on the R50-FPN backbone. As explained, in the only possible fair comparison, i.e. when using the same R50-FPN backbone, our model achieves higher multi-tasking performance compared to ASTMT ($+1.32\%$ vs $-0.74\%$). Yet, as ASTMT is by design single-scale and heavily based on DLv3+, we also report results using different backbones. Overall, MTI-Net achieves significantly higher gains over its single-task variants compared to ASTMT (see $\Delta_m \uparrow $ (ST)). Surprisingly, we find that our model with R18-FPN backbone even outperforms the deeper ASTMT R50-DLv3+ model in terms of multi-tasking performance, despite the fact that the ASTMT single-task models perform better than ours, due to the use of the stronger DLv3+ backbone. Note that we are the first to report consistent multi-task improvements when solving such a diverse task dictionary. Finally, our model also outperforms PAD-Net in terms of multi-tasking performance  ($+3.15 \%$ vs $-5.58\%$).

\begin{table}
    \caption[State-of-the-art comparison on PASCAL]{Comparison with the state-of-the-art on PASCAL. $\Delta_{MTL}$ is reported for a set of single-tasking models using the same backbone.}
    \label{tab: mti_sota_pascal}
    \centering
    \small{
    \begin{tabular}{ll|cccccc}
    \toprule
    \textbf{Method} & \textbf{Backbone} & \textbf{Seg} $\uparrow$ & \textbf{Parts} $\uparrow$ & \textbf{Sal} $\uparrow$ & \textbf{Edge} $\uparrow$ & \textbf{Norm} $\downarrow$ & $\mathbf{\Delta_{m} \%} \uparrow$ \\
    \midrule
    Single-Task & R18-FPN & 64.5 & 57.4 & 66.4 & 68.2 & 14.8 & + 0.00 \\
    & R50-FPN & 67.7 & 61.8 & 67.2 & 71.1 & 14.8 & + 0.00\\
    & R26-DLv3+ & 64.9 & 57.1 & 64.2 & 71.3 & 14.9 & + 0.00\\
    & R50-DLv3+ & 68.3 & 60.7 & 65.4 & 72.7 & 14.6 & + 0.00\\
    & HRNet-18 & 60.1 & 60.7 & 67.2 & 69.7 & 14.6 & + 0.00\\
    \midrule
    ASTMT & R50-FPN & 66.8 & 61.1 & 66.1 & 70.9 & 14.7 & - 0.74\\
    & R26-DLv3+ & 64.6 & 57.3 & 64.7 & 71.0 & 15.0 & - 0.11\\
    & R50-DLv3+ & 68.9 & 61.1 & 65.7 & 72.4 & 14.7 & - 0.04\\
    PAD-Net & HRNet-18 & 53.6 & 59.6 & 65.8 & 72.5 & 15.3 & - 5.58\\
    \midrule
    Ours & R18-FPN & 65.7 & 61.1 & 66.8 & 73.9 & 14.6 & + 3.41\\
    Ours & R50-FPN & 66.6 & 63.3 & 66.6 & 74.9 & 14.6 & + 1.32\\
    Ours & HRNet-18 & 64.3 & 62.1 & 68.0 & 73.4 & 14.6 & + 3.15\\
    \bottomrule
    \end{tabular}}
\end{table}

\subsubsection{Comparison on NYUD}
\begin{figure}
    \centering
    \includegraphics[width=\textwidth]{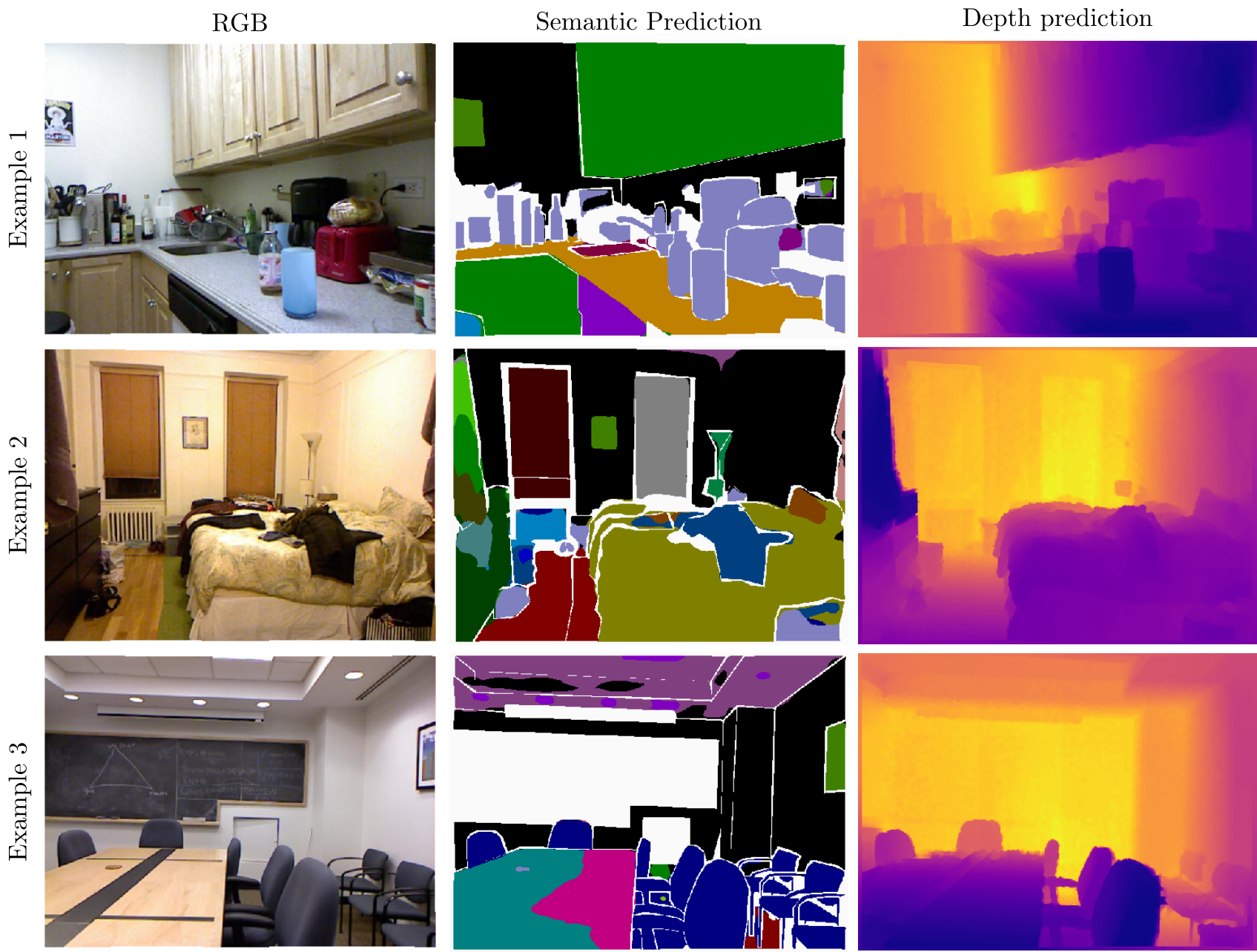}
    \caption[Qualitative results on NYUD]{Qualitative results on NYUD. We visualize semantic segmentation and depth predictions made by our HRNet-48 model.}
    \label{fig: mti_nyu}
\end{figure}

Table~\ref{tab: mti_nyu_sota} shows a comparison with the state-of-the-art approaches on NYUD. We leave out methods that rely on extra input modalities, or additional training data. As these methods are built on top of stronger single-scale backbones, we also use the multi-scale HRNet48-v2 backbone here. Again, our model improves w.r.t the single-task models. Furthermore, we perform on par with the state-of-the-art on the depth estimation task, while performing slightly worse on the semantic segmentation task. Figure~\ref{fig: mti_nyu} shows some qualitative results. 

\begin{table}
    \caption[State-of-the-art comparison on NYUD]{Comparison with the state-of-the-art on NYUD.}
    \label{tab: mti_nyu_sota}
    \begin{subtable}[t]{\linewidth}
    \centering
    \caption{Results on depth estimation.}
    \label{tab: mti_nyu_depth}
    \small{\begin{tabular}{l|cccccc}
    \toprule
    \textbf{Method} & \textbf{rmse} & \textbf{rel} & $\mathbf{\delta_1}$ & $\mathbf{\delta_2}$ & $\mathbf{\delta_3}$ \\
    \midrule
    HCRF \cite{li2015depth} & 0.821 & 0.232 & 0.621 & 0.886 & 0.968 \\
    DCNF \cite{liu2015learning} & 0.824 & 0.230 & 0.614 & 0.883 & 0.971 \\
    Wang \cite{wang2015towards} & 0.745 & 0.220 & 0.605 & 0.890 & 0.970 \\
    NR forest \cite{roy2016monocular} & 0.774 & 0.187 & - & - & - \\
    Xu \cite{xu2018structured} & 0.593 & 0.125 & 0.806 & 0.952 & 0.986 \\
    PAD-Net \cite{xu2018pad} & 0.582 & \textbf{0.120} & 0.817 & 0.954 & 0.987 \\
    PAP-Net \cite{zhang2019pattern} & 0.530 & 0.144 & 0.815 & 0.962 & 0.992 \\
    \midrule
    ST - HRNet-48 & 0.547 & 0.138 & 0.828 & 0.966 & 0.993 \\
    Ours - HRNet-48 & \textbf{0.529} & 0.138 & \textbf{0.830} & \textbf{0.969} & \textbf{0.993} \\
    \bottomrule
    \multicolumn{6}{c}{}\\
    \end{tabular}}
    \end{subtable} %
    \begin{subtable}[t]{\linewidth} 
    \centering
    \caption{Results on semantic segmentation.}
    \label{tab: mti_nyu_sem}
    \small{
    \begin{tabular}{l|cccc}
    \toprule
    \textbf{Method} & \textbf{pixel-acc} & \textbf{mean-acc} & \textbf{IoU} \\
    \midrule
    FCN \cite{long2015fully} & 60.0 & 49.2 & 29.2 \\
    Context \cite{lin2016efficient} & 70.0 & 53.6 & 40.6 \\
    Eigen \cite{eigen2015predicting} & 65.6 & 45.1 & 34.1 \\
    B-SegNet \cite{kendall2015bayesian} & 68.0 & 45.8 & 32.4 \\
    RefineNet-101 \cite{lin2017refinenet} & 72.8 & 57.8 & 44.9 \\
    PAD-Net \cite{xu2018pad} & 75.2 & 62.3 & 50.2 \\
    TRL-ResNet50 \cite{zhang2018joint} & 76.2 & 56.3 & 46.4 \\
    PAP-Net \cite{zhang2019pattern} & \textbf{76.2} & 62.5 & \textbf{50.4} \\
    \midrule
    ST - HRNet-48 & 73.4 & 58.1 & 45.7 \\
    Ours - HRNet-48 & 75.3 & \textbf{62.9} & 49.0\\
    \bottomrule
    \end{tabular}}
    \end{subtable}
\end{table}

\subsubsection{Resource analysis}
We compare our model in terms of computational requirements against PAD-Net and ASTMT. The comparison with PAD-Net is performed on NYUD using the HRNet-18 backbone, while for the comparison with ASTMT on PASCAL we use a ResNet-50 FPN backbone. Table~\ref{tab: mti_resource_analysis} reports the results relative to the single-tasking models. On NYUD, MTI-Net reduces the number of FLOPS while improving the performance compared to the single-task models. The reason for the increased amount of parameters is the use of a shallow backbone, and the small number of tasks (i.e. 2). Furthermore, we significantly outperform PAD-Net in terms of FLOPS and multi-task performance. This is due to the fact that PAD-Net performs the multi-modal distillation at a single higher scale (1/4) with $4 \cdot C$ channels, $C$ being the number of backbone channels at a single scale. Instead, we perform most of the computations at smaller scales (1/32, 1/16, 1/8), while operating on only $C$ channels at the higher scale (1/4). On PASCAL, we significantly improve on all three metrics compared to the single-task models. We also outperform ASTMT in terms of FLOPS and multi-task performance, as the latter has to perform a separate forward pass per task. 

\begin{table}
    \caption[Computational resource analysis]{Computational resource analysis (number of parameters and FLOPS).}
    \label{tab: mti_resource_analysis}
    \begin{subtable}[t]{\linewidth}
    \caption{Results on NYUD (HRNet-18).}
    \label{tab: mti_nyu_resources}
    \centering
    \small{
    \begin{tabular}{l|ccc}
    \toprule
    \textbf{Method} & \textbf{Params} (M) & \textbf{FLOPS} (G) & $\mathbf{\Delta_{m} \%}$ \\
    \midrule
    Single Task & 8.0 & 22.0 & $+0.00\%$ \\
    Multi-Task & $-50\%$ & $-45\%$ & $-1.71\%$ \\
    PAD-Net & $-15\%$ & $+204\%$ & $-0.02\%$ \\
    MTI-Net (Ours) & $+57\%$ & $-13\%$ & $+6.40\%$ \\
    \bottomrule
    \multicolumn{4}{c}{}\\
    \end{tabular}}
    \end{subtable} %
    \begin{subtable}[t]{\linewidth}
    \centering
    \caption{Results on PASCAL (Res-50 FPN).}
    \label{tab: mti_pascal_resources}
    \small{\begin{tabular}{l|ccc}
    \toprule
    \textbf{Method} & \textbf{Params} (M) & \textbf{FLOPS} (G) & $\mathbf{\Delta_{m} \%}$ \\
    \midrule
    Single Task	& 140 & 219	& $+0.00\%$	\\
    Multi-Task & $-75.0\%$ & $-66\%$ & $-4.55\%$ \\
    ASTMT &	$-51.0\%$ & $-1.0\%$ & $-0.87\%$ \\
    MTI-Net (Ours) & $-35.0\%$ & $-19.9\%$ & $+1.36\%$ \\
    \bottomrule
    \end{tabular}}
    \end{subtable}%
\end{table}

\section{Conclusion}
\label{sec: mti_conclusion}
We have shown the importance of modeling task interactions at multiple scales, enabling tasks to maximally benefit each other. We achieved this by introducing dedicated modules on top of an off-the-shelf multi-scale feature extractor, i.e. multi-scale multi-modal distillation, feature propagation across scales, and feature aggregation. Our multi-task model delivers on the full potential of multi-task learning, i.e. smaller memory footprint, reduced number of calculations and better performance. Our experiments show that our multi-task models consistently outperform their single-tasking counterparts by medium to large margins. 

\paragraph{Limitations} MTI-Net was designed with dense prediction tasks in mind. However, its easy to plug-in image-level tasks as well. Since MTI-Net only adds modules in the decoder (after the backbone) to model the task interactions, one could simply include a separate linear decoder for predicting the image-level tasks. After all, image-level tasks are defined at the image-level and do not require a multi-step multi-scale decoding process. 

\paragraph{Extensions/ Follow-Up Work} We summarize two recent works that proposed extensions to MTI-Net:
\begin{itemize}
    \item Borse \emph{et~al.}~\cite{borse2021inverseform} proposed a boundary-aware loss term for semantic segmentation using an inverse-transformation network, which learns the degree of parametric transformations between estimated and target boundaries. This plug-in loss can complement the cross-entropy loss in capturing boundary transformations and allows to further improve the results obtained with MTI-Net. 
    \item Bruggemann \emph{et~al.}~\cite{bruggemann2020automated} proposed a neural architecture search method to select the refinement units between tasks. The newly generated modules can be used as a drop-in refinement module for any supervised multi-task architecture. 
    \end{itemize}
Finally, an interesting direction for future work is combining the design concepts from MTI-Net with recent architectures like transformers.
\cleardoublepage%

  \graphicspath{{\chaptersdir/optimization/image/}}%
\chapter{Task Balancing Strategies}
\label{ch: optimization}
So far, we focused on designing better architectures for multi-task learning. Another challenge associated with jointly learning multiple tasks is properly weighting the task-specific loss functions. Therefore, in this chapter, we revisit the task balancing techniques from Section~\ref{subsec: related_task_balancing}. We carefully analyze several existing methods. First, we draw a qualitative comparison listing the most important commonalities and differences between works. Additionally, we perform an experimental evaluation on a variety of benchmarks. We draw several important conclusions: (i) there are various discrepancies between works, (ii) the use of carefully selected fixed weights is competitive with the best performing task balancing methods and (iii) the performance of existing methods is coupled with biases in the training dataset. Based upon these results, we propose a set of heuristics to weigh the task-specific losses. The heuristics are of more practical use and obtain better results across different benchmarks. 

The content of this chapter is based upon the following publication:
\begin{itemize}
    \item \emph{Vandenhende, S., Georgoulis, S., Van Gansbeke, W., Proesmans, M., Dai, D., \& Van Gool, L. (2021). Multi-task learning for dense prediction tasks: A survey. IEEE Transactions on Pattern Analysis and Machine Intelligence.}
\end{itemize}

\section{Introduction}
\label{sec: optimization_introduction}
The multi-task learning optimization objective can be formulated as
\begin{equation}
\label{eq: optim_mtl_loss}
\mathcal{L}_{MTL} = \sum_{i} w_i \cdot \mathcal{L}_i
\end{equation}
where $w_i$ and $\mathcal{L}_i$ denote the task-specific weights and loss functions respectively. We need to carefully balance the joint learning of all tasks to avoid a scenario where one or more tasks have a dominant influence in the network weights. This chapter discusses how to achieve this by selecting the task-specific weights $w_i$ in the optimization objective. 

There have been several attempts already to automate the selection of the task-specific weights. Note that we provided a detailed description of these works in Chapter~\ref{ch: optimization}. We provide a brief recap here. Early work~\cite{kendall2018multi} used the homoscedastic uncertainty of each task to weigh the losses. Gradient normalization~\cite{chen2018gradnorm} balances the learning of tasks by dynamically adapting the gradient magnitudes in the network. Liu~\emph{et~al.}~\cite{liu2019end} weigh the losses to match the pace at which different tasks are learned. Dynamic task prioritization~\cite{guo2018dynamic} prioritizes the learning of difficult tasks. Sener~\emph{et~al.}~\cite{sener2018multi} cast MTL as a multi-objective optimization problem, with the overall objective of finding a Pareto optimal solution. 

Section~\ref{sec: optimization_qualitative} compares different task balancing methods from a qualitative point-of-view. Section~\ref{sec: optimization_experiments} provides an experimental comparison. Based upon the results, we propose a set of heuristics to select the task-specific weights in Equation~\ref{eq: optim_mtl_loss}. 

\section{Qualitative Comparison}
\label{sec: optimization_qualitative}
We draw a qualitative comparison between the following methods: uncertainty weighting~\cite{kendall2018multi}, GradNorm~\cite{chen2018gradnorm}, DWA~\cite{liu2019end}, DTP~\cite{guo2018dynamic} and MGDA~\cite{sener2018multi}. First, we consider whether a method balances the loss magnitudes and/or the pace at which tasks are learned. Second, we show what tasks are prioritized during the training stage, e.g. DTP prioritizes the learning of difficult tasks. Third, methods that require access to the task-specific gradients are indicated. Fourth, we consider whether the task gradients are enforced to be non-conflicting, e.g. MGDA only backpropagates along common directions of the task-specific gradients. Finally, we consider whether the proposed method requires to tune additional hyperparameters, e.g. manually choosing the weights, KPIs, etc. 

Table~\ref{tab: task_balancing} shows the results. Our conclusions are summarized below:
\begin{itemize}
    \item We observe a number of discrepancies between the included methods. For example, uncertainty weighting assigns a higher weight to the 'easy' tasks, while DTP advocates the opposite. The latter can be attributed to the experimental evaluation of the different task balancing strategies that was often done in the literature using different datasets or task dictionaries. We suspect that an appropriate task balancing strategy should be decided for each case individually.
    \item We also find a few commonalities between the described methods, e.g. both uncertainty weighting, GradNorm and MGDA opted to balance the loss magnitudes as part of their learning strategy. 
    \item Finally, some works (e.g. DWA, DTP) still require careful manual tuning of task-specific hyperparameters, which limits their applicability for dealing with a large number of tasks. 
\end{itemize}

Section~\ref{sec: optimization_experiments} provides an extensive ablation study under more common datasets or task dictionaries to verify what task balancing strategies are most useful to improve the MTL performance, and under under which circumstances.

\section{Experiments}
\label{sec: optimization_experiments}

\subsection{Setup}
\label{subsec: optimization_setup}
The experiments are performed on three different datasets: CelebA, NYUD and PASCAL. Section~\ref{sec: datasets} provides a description of the included datasets. The multi-task models consist of a shared backbone (i.e. ResNet) with task-specific heads. The setup from Chapter~\ref{ch: branched_mtl} is used for CelebA. The backbone is a ResNet-18. We attach a separate linear head for every attribute. The fixed weights scheme simply sums the task-specific losses together, i.e. the task-specific weights are set equal to one. We adopt the setup from Chapter~\ref{ch: mti_net} for the experiments on NYUD and PASCAL. A ResNet-18 and ResNet-50 backbone are used for PASCAL and NYUD respectively. The task-specific heads use an ASPP module. On NYUD, we sum the losses together, meaning the fixed weight scheme overlaps with the uniform weights. Differently, on PASCAL, we obtain the fixed weights from a grid search: $w_{seg}=1.0$, $w_{parts}=2.0$, $w_{sal}=5.0$, $w_{edge}=50.0$ and $w_{normals}=10.0$.

\begin{landscape}
\begin{table*}[t]
\caption[Qualitative comparison of task balancing techniques]{A qualitative comparison between task balancing techniques. First, we consider whether a method balances the loss magnitudes (\textbf{Balance Magnitudes}) and/or the pace at which tasks are learned (\textbf{Balance Learning}). Second, we show what tasks are prioritized during the training stage (\textbf{Prioritize}). Third, we show whether the method requires access to the task-specific gradients (\textbf{Grads Required}). Fourth, we consider whether the tasks gradients are enforced to be non-conflicting (\textbf{Non-Competing Grads}). Finally, we consider whether the proposed method requires additional tuning, e.g. manually choosing the weights, KPIs, etc. (\textbf{No Extra Tuning}). For a detailed description of each technique see Section~\ref{sec: related_optimization}.}
\label{tab: task_balancing}
\centering
\small{
\begin{tabular}{lccccccl}
\toprule
\textbf{Method} & \textbf{\multirowcell{2}[0pt][l]{Balance \\ Magnitudes}} & \textbf{\multirowcell{2}[0pt][l]{Balance \\ Learning}} &
\textbf{Prioritize} & 
\textbf{\multirowcell{2}[0pt][l]{Grads \\ Required}} &
\textbf{\multirowcell{2}[0pt][l]{Non-Competing \\ Grads}} &
\textbf{\multirowcell{2}[0pt][l]{No Extra \\ Tuning }} & 
\textbf{Motivation} \\
& & & & & & \\
\midrule
Uncertainty~\cite{kendall2018multi} & $\checkmark$ & & Low Noise & & &  $\checkmark$ & Homoscedastic uncertainty \\
GradNorm~\cite{chen2018gradnorm} & $\checkmark$ & $\checkmark$ & & $\checkmark$ & & $\checkmark$ &  Balance learning and magnitudes \\
DWA~\cite{liu2019end} & & $\checkmark$ & & & & &  Balance learning \\
DTP~\cite{guo2018dynamic} & & & Difficult & & & & Prioritize difficult tasks \\
MGDA~\cite{sener2018multi} & $\checkmark$ & $\checkmark$ & & $\checkmark$ & $\checkmark$ & $\checkmark$ & Pareto Optimum \\
\bottomrule
\end{tabular}}
\end{table*}
\end{landscape}

\subsection{Results}
\label{subsec: optimization_results}

\begin{table}
\caption[Experimental comparison of task balancing techniques]{Comparison of different tasks balancing techniques. The setup is described in Section~\ref{subsec: optimization_setup}.}
\label{tab: optimization_results}
\begin{subtable}[t]{\linewidth}
\caption[Task balancing results on CelebA]{Results on CelebA for 40 person attribute classification tasks.}
\label{tab: optimization_celeba}
\centering
\small{
\begin{tabular}{l|cccc}
\toprule
\textbf{Method} & \textbf{Accuracy} (\%) $\uparrow$\\
\midrule
Single-task & 91.52 \\
\midrule
Fixed & \textbf{91.91} \\
Uncertainty & 91.62 \\
GradNorm & 91.76 \\
DWA & 91.84 \\
MGDA & 91.30 \\
\bottomrule
\multicolumn{2}{c}{}\\
\end{tabular}}
\end{subtable} %

\begin{subtable}[t]{\linewidth}
    \centering
    \caption[Task balancing results on NYUD]{Results on NYUD.}
    \label{tab: optimization_nyud}
    \small{
    \begin{tabular}{l|ccc}
    \toprule
    \textbf{Method} & \textbf{Seg. } & \textbf{Depth} &  $\mathbf{\Delta_{MTL}}$ \\
    & (IoU) $\uparrow$ & (rmse) $\downarrow$ & (\%)$\uparrow$\\
    \midrule
    Single-Task & 43.9 & 0.585 & + 0.00 \\
    \midrule
    Fixed & \textbf{44.4} & 0.587 & + 0.41 \\
    Uncertainty & 44.0 & 0.590 & - 0.23 \\
    DWA & 44.1 & 0.591 & - 0.28 \\
    GradNorm & 44.2 & \textbf{0.581} & \textbf{+ 1.45} \\
    MGDA & 43.2 & 0.576 & + 0.02 \\
    \bottomrule
    \multicolumn{4}{c}{} \\
    \end{tabular}}
\end{subtable} %

\begin{subtable}[t]{\linewidth}
    \centering
    \caption[Task balancing results on PASCAL]{Results on PASCAL.}
    \label{tab: optimization_pascal}
    \small{
    \begin{tabular}{l|ccccccc}
    \toprule
    \textbf{Method} & \textbf{Seg.}& \textbf{H. Parts} & \textbf{Norm.} & \textbf{Sal.} & \textbf{Edge} & $\mathbf{\Delta_{MTL}}$\\
    &  (IoU) $\uparrow$ & (IoU) $\uparrow$ & (mErr) $\downarrow$ & (IoU) $\uparrow$ & (odsF) $\uparrow$ & $(\%)\uparrow$ \\
    \midrule
    Single-Task & \textbf{66.2} & \textbf{59.9} & \textbf{13.9} & \textbf{66.3} & 68.8 & \textbf{+ 0.00} \\
    \midrule
    Fixed (Grid S.) & 63.8 & 58.6 & 14.9 & 65.1 & \textbf{69.2} & - 2.86 \\
    Uncertainty & 65.4 & 59.2 & 16.5 & 65.6 & 68.6 & - 4.60 \\
    DWA & 63.4 & 58.9 & 14.9 & 65.1 & 69.1 & - 2.94 \\
    GradNorm & 64.7 & 59.0 & 15.4 & 64.5 & 67.0 & - 3.97 \\
    MGDA & 64.9 & 57.9 & 15.6 & 62.5 & 61.4 & -6.81 \\
    \bottomrule
    \end{tabular}}
    \end{subtable}
\end{table}

\subsubsection{CelebA} 
Table~\ref{tab: optimization_celeba} reports the results for CelebA. The multi-task model trained with fixed uniform weights outperforms the single-task networks ($91.91\%$ vs $91.52\%$). We argue that the additional supervisory signals from other attributes can act as a regularizer for the backbone features. Surprisingly, none of the proposed task balancing strategies improves over the model trained with fixed uniform weights on CelebA, although all methods are very close. We conclude that the uniform weighting scheme is already close to optimal in case of multi-attribute learning, where all individual losses are binary cross-entropy losses. Note that, a similar experiment was performed in~\cite{sener2018multi}, but different conclusions were reported. First, they found that the multi-task model trained with fixed weights performs worse when compared against the single-task networks. Second, the authors observed that the loss balancing strategies improve over the uniform weights. In this manuscript, we find that with sufficient and comparable amounts of hyperparameter tuning both conclusions no longer stand. 

\subsubsection{NYUD} 
Table~\ref{tab: optimization_nyud} shows the results when training the MTL baseline with a ResNet-50 backbone using different task balancing strategies. On NYUD, optimizing the loss weights with a grid search procedure resulted in a uniform loss weighting scheme. Therefore, in this case, the use of fixed uniform weights and the use of fixed weights from a grid search overlap. 

The MTL baseline with fixed weights improves over the single-task networks ($+0.41\%$ for $\Delta_{MTL}$). GradNorm can further boost the performance by adjusting the task-specific weights in the loss during training ($+1.45\%$). We conclude that uniform weights are suboptimal for training a multi-task model when the tasks employ different loss functions, although the differences are limited.

Similar to GradNorm, DWA tries to balance the pace at which tasks are learned, but does not equalize the gradient magnitudes. From Table~\ref{tab: optimization_nyud}, we conclude that the latter is important as well ($-0.28\%$ with DWA vs $+1.45\%$ with GradNorm). Uncertainty weighting results in reduced performance compared to GradNorm ($-0.23\%$ vs $1.45\%$). Uncertainty weighting assigns a smaller weight to noisy or difficult tasks. Since the annotations on NYUD were not distilled, the noise levels are rather small. When we have access to clean ground-truth annotations, it seems better to balance the learning of tasks rather than lower the weights for the difficult tasks. Further, MGDA reports lower performance compared to GradNorm ($+0.02\%$ vs $+1.45\%$). MGDA only updates the weights along directions that are common among all task-specific gradients. The results suggest that it is better to allow some competition between the task-specific gradients in the shared layers, as this could help to avoid local minima. 

Finally, we conclude that the use of fixed loss weights optimized with a grid search still outperforms several existing task balancing methods. In particular, the solution obtained with fixed uniform weights outperforms the models trained with uncertainty weighting, MGDA and DWA on both the semantic segmentation and depth estimation tasks.  Perhaps not surprisingly, we found that the uniform weights resulted in task-specific losses of similar magnitudes. In other words, it seems sufficient to make sure that the losses are of similar magnitude to obtain good results. 
\subsubsection{PASCAL} 
The task balancing techniques are compared on PASCAL in Table~\ref{tab: optimization_pascal}. We find that a grid-search on the weight space works better than the use of automated task balancing procedures. A similar observation was made by~\cite{maninis2019attentive}. We hypothesize that this is due to the imbalance between the optimal parameters, e.g. the weight for the edge detection loss is 50 times higher than for the semantic segmentation loss. Also, we observe that the grid-search weights result in task-specific losses of similar magnitudes. This again hints that its useful to ensure that the losses are of similar magnitude. 

Uncertainty weighting reports the highest performance losses on the distilled tasks, i.e. normals and saliency. This is because uncertainty weighting assigns a smaller weight to tasks with higher homoscedastic uncertainty. Differently, MGDA fails to properly learn the edge detection task. This is another indication (cf. NYUD results) that avoiding competing gradients by only back propagating along a common direction in the shared layers does not necessarily improve performance. This is particularly the case for the edge detection task on PASCAL because of two reasons. First, the edge detection loss has a much smaller magnitude compared to the other tasks. Second, when the edge annotations are converted into segments they provide an over-segmentation of the image. As a consequence, the loss gradient of the edge detection task often conflicts with other tasks since they have a smoother gradient. Because of this, MGDA prefers to mask out the gradient from the edge detection task by assigning it a smaller weight. Finally, GradNorm does not report higher performance when compared against uniform weighting. We hypothesize that this is due to the re-normalization of the loss weights during the optimization. The latter does not work well when the optimal loss magnitudes are very imbalanced. 

\subsubsection{Performance Profiles} 
\begin{figure}
    \caption[Performance profiles of different task balancing strategies]{Performance profile of task balancing strategies for the semantic segmentation and depth estimation tasks on NYUD. We show the results obtained with different hyperparameter settings in the same color. Bottom-right is better.}
    \label{fig: optimization_hyperparam}
    \centering
    \includegraphics[width=0.8\linewidth]{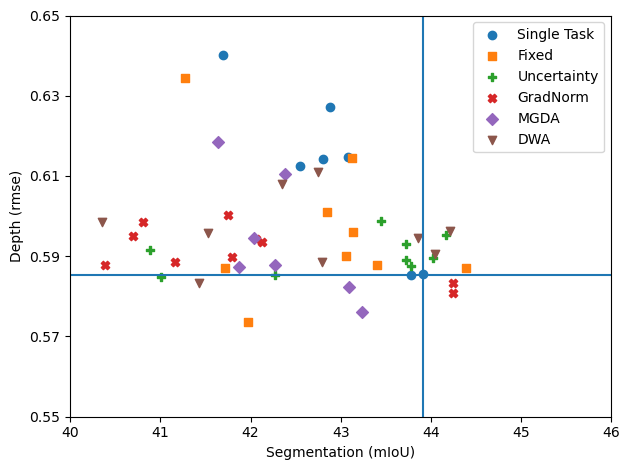}
\end{figure}

We use the performance profile from Figure~\ref{fig: optimization_hyperparam} to analyze the hyperparameter sensitivity of the task balancing methods on NYUD. GradNorm is the only technique that outperforms training a separate model for each task. However, when less effort is spent to tune the hyperparameters of the single-task models, all task balancing techniques result in improved performance. This observation explains the increased performance relative to the single-task case reported by prior works~\cite{kendall2018multi,chen2018gradnorm,sener2018multi}. We stress the importance of carefully training the single-task baselines.

Second, the MTL models seem more robust to hyperparameter changes compared to the single-task networks. In particular, the performance drops slower when using less optimal hyperparameter settings in the MTL case. 

Finally, fixed loss weighting and uncertainty weighting are most robust against hyperparameter changes. These techniques report high performance for a large number of hyperparameter settings. For example, out of the top-20\% best models, 40\% of those were trained with uncertainty weighting. 

\subsection{Discussion}
\label{subsec: optimization_discussion}
To our surprise, we found that grid-search is competitive or better compared to existing task balancing techniques. Unfortunately, performing a grid search becomes intractable, if not infeasible, when considering large task dictionaries. However, there are several tricks that practitioners can apply to circumvent this problem. In particular, we recommend the adoption of a set of heuristics to obtain an effective task weighting in an efficient manner. We observed that the task-specific weights are competitive with the ones from the grid-search procedure. 

\begin{itemize}
    \item \textbf{Heuristic 1: Related tasks using the same loss functions can use the same weights.} For example, on CelebA, the person attribute classification tasks are strongly related and use the same loss functions. In this case, using uniform, fixed weights gives good results (see Table~\ref{tab: optimization_celeba}).
    
    \item \textbf{Heuristic 2: If the tasks use different task-specific losses, then adjust the task-specific weights such that all losses are of similar magnitude.} This can be realized by tracking the task-specific losses over a few iterations for every task. The task-specific weights can be set based upon the averaged loss values. For example, assume two tasks with the average loss value of the first task being 100 times larger compared to the average loss value of task 2. In this case, the weights are set as $w_1$ = 1 and $w_2$ = 100. This trick applies to PASCAL and NYUD (see Tables~\ref{tab: optimization_nyud}-\ref{tab: optimization_pascal}). We verified that the obtained weights are similar to the ones from the grid search and the differences in performance are negligible.  
    
    \item \textbf{Heuristic 3: Increase the task-specific weights for tasks that are more important.} This ensures more effort is spend to optimize the tasks of interest.
    
    \item \textbf{Heuristic 4: Increase the task-specific weights for tasks that are more difficult.} For example, the task-specific weights could be updated at regular intervals during training following the procedure outlined for Heuristic 2. This heuristic is based upon Dynamic Task Prioritization~\cite{guo2018dynamic}. 

    \item \textbf{Heuristic 5: When multiple datasets with different task dictionaries are used, compose the batches such that they show more examples for the more challenging tasks.} Bell~\emph{et~al.}~\cite{bell2020groknet} show how this can be realized. Assume $n$ datasets. The design space is explored by performing two-pair dataset ablations to understand the relative balance of the datasets. For example, if for some task the accuracy is insensitive to the batch size, then we can reduce that dataset's contribution to each batch, leaving more slots for the challenging tasks. 
\end{itemize}

Further, a number of techniques performed worse than anticipated. Gong~\emph{et~al.}~\cite{gong2019comparison} obtained similar results with ours, albeit only for a few loss balancing strategies. Also, Maninis~\emph{et~al.}~\cite{maninis2019attentive} found that performing a grid search for the weights could outperform a state-of-the-art loss balancing scheme. These observations seem particularly true for larger datasets with more diverse task dictionaries like PASCAL (see Table~\ref{tab: optimization_pascal}). We argue that new methods should also evaluate their approach under more challenging conditions. Finally, existing task balancing methods can still be useful to train MTL networks. For example, when dealing with noisy annotations, uncertainty weighting can help to automatically readjust the weights on the noisy tasks.

\section{Conclusion}
\label{sec: conclusion}
We analyzed multiple task balancing strategies and isolated the elements that are most effective for balancing task learning, e.g. down-weighing noisy tasks, balancing loss magnitudes, etc. Yet, many optimization aspects still remain poorly understood. For example, opposed to recent works, our analysis indicates that avoiding gradient competition between tasks can hurt performance. Furthermore, our study revealed that some task balancing strategies still suffer from shortcomings and highlighted several discrepancies between existing methods. Finally, we proposed a set of heuristics based upon our experimental results. These aim to make it easier for practitioners to train multi-task learning models in an effective way. 

\paragraph{Limitations} The optimization strategies have been studied using the multi-task learning baseline model. It would be interesting to see if our conclusions translate to other, possibly more sophisticated models. Further, we studied the optimization aspect from an experimental point-of-view. For future work, it would be useful to develop a theoretical framework to examine these methods.  

\cleardoublepage%

  \graphicspath{{\chaptersdir/ssl/image/}}%
\chapter{Self-Supervised Learning}
\label{ch: ssl}
The experiments in this thesis were performed using supervised pretraining on ImageNet. This setup builds upon the fact that the weights of convolutional neural networks transfer well between tasks. As a result, pretraining on a large dataset like ImageNet allows to perform well on tasks with only a modest amount of labelled data (e.g. segmentation on NYUD or PASCAL). Pretraining is a powerful technique that is widely known and applies to both single- and multi-task learning. 

Unfortunately, supervised pretraining relies on an annotated pretraining dataset, which limits it's applicability. To resolve this problem, researchers have begun to explore self-supervised learning methods. The latter group of methods learn visual representations by solving \emph{pretext tasks} for which the learning signal comes for free. In this chapter, we aim to explore the use of self-supervised pretraining to improve models for scene understanding. 

Recently, self-supervised learning has outperformed supervised pretraining on many downstream tasks like segmentation and object detection. However, current methods are still primarily applied to curated datasets like ImageNet. In this chapter, we first study how biases in the dataset affect existing methods. Our results show that an approach like MoCo~\cite{he2019momentum} works surprisingly well across: (i) object- versus scene-centric, (ii) uniform versus long-tailed and (iii) general versus domain-specific datasets. Second, given the generality of the approach, we try to realize further gains with minor modifications. We show that learning additional invariances - through the use of multi-scale cropping, stronger augmentations and nearest neighbors - improves the representations. Next, we observe that MoCo learns spatially structured representations when trained with a multi-crop strategy. The representations can be used for semantic segment retrieval and video instance segmentation without finetuning. Moreover, the results are on par with specialized models. Finally, we apply contrastive self-supervised pretraining to improve a multi-task learning model for indoor scene understanding. 

The content of this chapter is based upon the following preprint\footnote{* denotes equal contribution.}: 
\begin{itemize}
    \item Van Gansbeke, W.*, \textbf{Vandenhende, S.}*, Georgoulis, S., \& Van Gool, L. (2021). Revisiting Contrastive Methods for Unsupervised Learning of Visual Representations. NeurIPS.
\end{itemize}

\section{Introduction}
\label{sec: ssl_introduction}
Self-supervised learning (SSL)~\cite{jing2020self} aims to learn powerful representations without relying on human annotations. The representations can be used for various purposes, including transfer learning~\cite{he2019momentum}, clustering~\cite{asano2020labelling,van2020scan,van2021unsupervised} or semi-supervised learning~\cite{chen2020big}. Recent self-supervised methods~\cite{chen2020simple,he2019momentum,caron2020unsupervised,grill2020bootstrap} learn visual representations by imposing invariances to various data transformations. A popular way of formulating this idea is through the instance discrimination task~\cite{wu2018unsupervised} - which treats each image as a separate class. Augmentations of the same image are considered as positive examples of the class, while other images serve as negatives. To handle the large number of instance classes, the task is expressed as a non-parametric classification problem using the contrastive loss~\cite{gutmann2010noise,oord2018representation}. 

Despite the recent progress, most methods still train on images from ImageNet~\cite{deng2009imagenet}. This dataset has specific properties: (1) the images depict a single object in the center of the image, (2) the classes follow a uniform distribution and (3) the images have discriminative visual features. To deploy self-supervised learning into the wild, we need to quantify the dependence on these properties. Therefore, in this chapter, we first study the influence of dataset \emph{biases} on the representations. We take a utilitarian view and transfer different representations to a variety of downstream tasks.

Our results indicate that an approach like MoCo~\cite{he2019momentum} works well for both object- and scene-centric datasets. We delve deeper to understand these results. A key component is the augmentation strategy which involves random cropping. For an object-centric dataset like ImageNet, two crops from the same image will show a portion of the same object and no other objects. However, when multiple objects are present, a positive pair of non-overlapping crops could lead us to wrongfully match the feature representations of different objects. This line of thought led recent studies~\cite{purushwalkam2020demystifying,selvaraju2020casting} to believe that contrastive SSL benefits from object-centric data. 

So how can contrastive methods learn useful representations when applied to more complex, scene-centric images? We propose a hypothesis that is two-fold. First, the default parameterization of the augmentation strategy avoids non-overlapping views. As a result, positive pairs will share information, which means that we can match their representations. Second, when applying more aggressive cropping, we only observe a small drop in the transfer learning performance. Since patches within the same image are strongly correlated, maximizing the agreement between non-overlapping views still provides a useful learning signal. We conclude that, in visual pretraining, combining the instance discrimination task with random cropping is universally applicable. 

The common theme of recent advances is to learn representations that are invariant to different transformations. Starting from this principle, we try to improve the results obtained with an existing framework~\cite{he2019momentum}. More specifically, we investigate three ways of generating a more diverse set of positive pairs. First, the multi-crop transform from~\cite{caron2020unsupervised} is revisited. Second, we examine the use of a stronger augmentation policy. Third, we leverage nearest neighbors mined online during training as positive views. The latter imposes invariances which are difficult to learn using handcrafted image transformations. The proposed implementation requires only a few lines of code and provides a simple, yet effective alternative to clustering based methods~\cite{caron2020unsupervised,li2020prototypical}. Each of the proposed additions is found to boost the performance of the representations under the transfer learning setup. 

The multi-crop transform realizes significant gains. We probe into what the network learns to explain the improvements. The multi-crop transform maximizes the agreement between smaller (local) crops and a larger (global) view of the image. This forces the model to learn a more spatially structured representation of the scene. As a result, the representations can be directly used to solve several dense prediction tasks without any finetuning. In particular, we observe that the representations already model class semantics and dense correspondences. Furthermore, the representations are competitive with specialized methods~\cite{jabri2020walk,zhang2020self}. In conclusion, the multi-crop setup provides a viable alternative to learn dense representations without relying on video data~\cite{jabri2020walk,lai2020mast} or handcrafted priors~\cite{hwang2019segsort,zhang2020self,van2021unsupervised}.

In summary, the overall goal of this chapter is to learn more effective representations through contrastive self-supervised learning without relying too much on specific dataset biases. We further illustrate the usefulness of self-supervised learning for an application in multi-task learning. The remainder of this chapter is structured as follows. Section~\ref{sec: ssl_framework} introduces the framework. Section~\ref{sec: ssl_contrastive_wild} studies whether the standard SimCLR augmentations transfer across different datasets. This question is answered positively. Based upon this result, Section~\ref{sec: ssl_invariances} then studies the use of additional invariances to further improve the learned representations. Finally, Section~\ref{sec: ssl_application} applies our pretrained weights to improve a multi-task learning model for indoor scene understanding. We hope this chapter will provide useful insights to other researchers. .

\section{Framework}
\label{sec: ssl_framework}
We briefly introduce the contrastive learning framework. The idea is to generate feature representations that maximize the agreement between similar (\emph{positive}) images and minimize the agreement between dissimilar (\emph{negative}) images. Let $x$ be an image. Assume that a set of positives for $x$ can be acquired, denoted by $\mathcal{X}^+$. Similarly, a set of negatives $\mathcal{X}^-$ is defined. We learn an embedding function $f$ that maps each sample on a normalized hypersphere. The contrastive loss~\cite{gutmann2010noise,oord2018representation} takes the following form
\begin{equation}
\mathcal{L}_{\text{contrastive}} = -\sum_{x^+ \in \mathcal{X^+}} \log \frac{\exp\left[f(x)^T \cdot f(x^+) / \tau\right]}{\exp\left[ f(x)^T \cdot f(x^+) / \tau\right] + \sum_{x^- \in\mathcal{X^-}} \exp\left[f(x)^T \cdot f(x^-) / \tau\right]}
\label{eq: contrastive_loss}
\end{equation}
where $\tau$ is a temperature hyperparameter. We will further refer to the image $x$ as the \emph{anchor}. 

SSL methods obtain positives and negatives by treating each image as a separate class~\cite{wu2018unsupervised}. More specifically, augmented views of the same image are considered as positives, while other images are used as negatives. The data augmentation strategy is an important design choice as it determines the invariances that will be learned. Today, most works rely on a similar set of augmentations that consists of (1) cropping, (2) color distortions, (3) horizontal flips and (4) Gaussian blur.

In this manuscript, we build upon MoCo~\cite{he2019momentum} - a widely known and competitive framework. However, our findings are expected to apply to other related methods (e.g. SimCLR~\cite{chen2020simple}) as well. The embedding function $f$ with parameters $\theta_f$ consists of a backbone $g$ (e.g. ResNet~\cite{he2016deep}) and a projection MLP head $h$. The contrastive loss is applied after the projection head $h$. MoCo uses a queue and a moving-averaged encoder $f'$ to keep a large and consistent set of negative samples. The parameters $\theta_{f'}$ of $f'$ are updated as: $\theta_{f'} = m \theta_{f'} + (1-m) \theta_f$ with $m$ a momentum hyperparameter. The momentum-averaged encoder $f'$ takes as input the anchor $x$, while the encoder $f$ is responsible for the positives $\mathcal{X}^+$. The queue maintains the encoded anchors as negatives. Algorithm~\ref{alg: moco} provides the pseudocode of MoCo. We refer to~\cite{he2019momentum} for more details. 

\begin{algorithm}
\newcommand{\hlp}{\makebox[0pt][l]{\color{Mulberry!15}\rule[-3pt]{0.72\linewidth}{9pt}}}
\newcommand{\hlb}{\makebox[0pt][l]{\color{NavyBlue!15}\rule[-3pt]{0.72\linewidth}{9pt}}}
\algcomment{\fontsize{7.2pt}{0em}\selectfont \texttt{bmm}: batch matrix mult.; \texttt{mm}: matrix mult.; \texttt{cat}: concatenate; \texttt{CE}: cross-entropy loss}
\definecolor{codeblue}{rgb}{0.25,0.5,0.5}
\lstset{
  backgroundcolor=\color{white},
  basicstyle=\fontsize{7.2pt}{7.2pt}\ttfamily\selectfont,
  columns=fullflexible,
  breaklines=true,
  captionpos=b,
  commentstyle=\fontsize{7.2pt}{7.2pt}\color{codeblue},
  keywordstyle=\fontsize{7.2pt}{7.2pt},
}
\begin{lstlisting}[language=python, escapechar=@]
@\vspace{-2.0em}@
# f, f_m: the model f and momentum updated model f_m
# f consists of the backbone g and projection head h
# queue: dictionary as a queue of K keys (C x K)
# m: momentum
# t: temperature

f_m.params = f.params # initialize momentum updated backbone

for batch in loader: # load a minibatch with B samples
   
    # randomly augment batch
    x_anchors = aug(batch)
    x_positives = aug(batch)
    
    # forward pass
    anchors, positives = f_m(x_anchors), f(x_positives) # B x C
    anchors = anchors.detach() # no gradient
    
    # compute logits
    l_pos = bmm(positives.view(B,1,C), anchors.view(B,C,1)) # B x 1
    l_neg = mm(positives, queue.view(C,K)) # B x K
    logits = cat([l_pos, l_neg], dim=1) # B x (K+1)
    
    # loss: (1) sharpen with temperature t, (2) apply cross-entropy loss
    loss = CE(logits/t, zeros(B))
    
    # SGD update
    loss.backward()
    update(f.params)
    
    # momentum update
    f_m.params = m * f_m.params + (1-m) * f.params
    
    # update dictionaries
    enqueue_dequeue(queue, anchors) 
\end{lstlisting}
\captionof{algorithm}{Pseudocode of MoCo~\cite{he2019momentum} in a PyTorch-like style.}
\label{alg: moco}
\end{algorithm}

\section{Contrastive Learning in the Wild}
\label{sec: ssl_contrastive_wild}
Most contrastive self-supervised methods train on unlabeled images from ImageNet~\cite{deng2009imagenet}. This is a curated dataset with unique characteristics. First, the images are \emph{object-centric}, i.e. they depict only a single object in the center of the image. This differs from other datasets~\cite{lin2014microsoft,kuznetsova2020open} which contain more complex scenes with several objects. Second, the underlying classes are \emph{uniformly} distributed. Third, the images have \emph{discriminative} visual features. For example, ImageNet covers various bird species which can be distinguished by a few key features. In contrast, domain-specific datasets (e.g. BDD100K~\cite{yu2020bdd100k}) contain less discriminative scenery showing the same objects like cars, pedestrians, etc. In this section, we study the influence of dataset biases for contrastive self-supervised methods. 

\subsection{Setup} 
\begin{figure}
    \centering
    \includegraphics[width=0.8\linewidth]{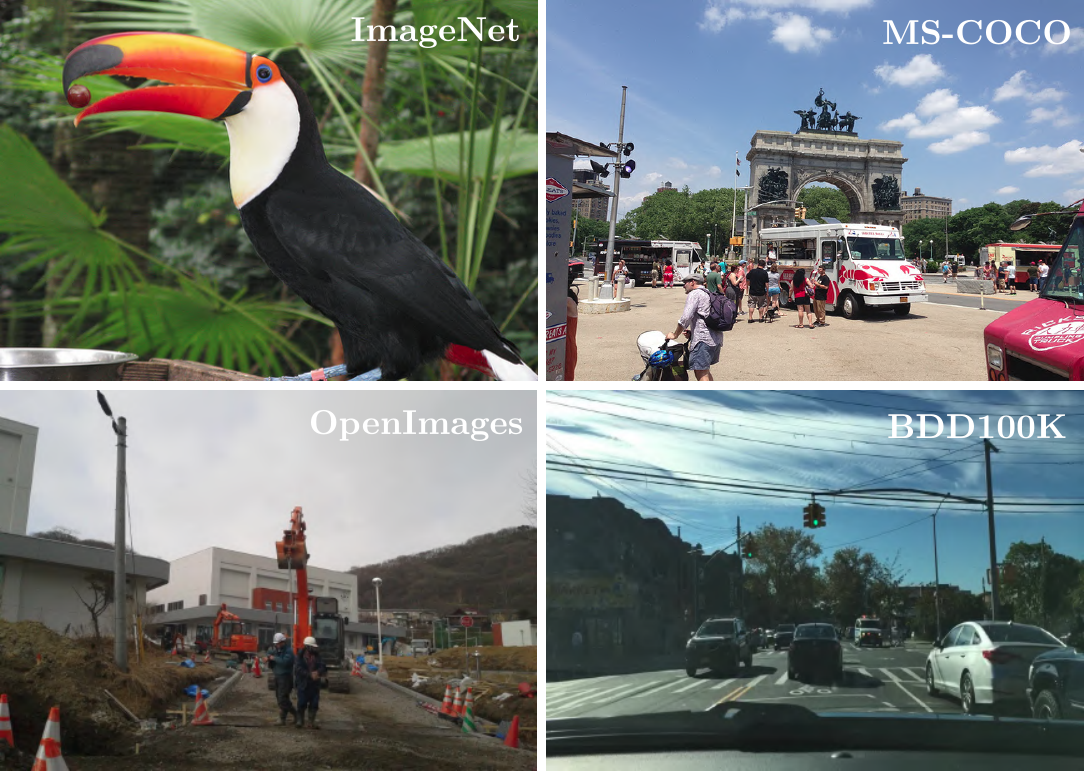}
    \caption{Example images from the pretraining datasets.}
    \label{fig: ssl_datasets}
\end{figure}

We train MoCo-v2~\cite{chen2020improved} on a variety of datasets. Table~\ref{tab: ssl_datasets} shows an overview. Figure~\ref{fig: ssl_datasets} shows a representative sample from each dataset. The representations are evaluated on six downstream tasks: linear classification, semantic segmentation, object detection, video instance segmentation and depth estimation. We adopt the following target datasets for linear classification: CIFAR10~\cite{krizhevsky2009learning}, Food-101~\cite{krause2013collecting}, Pets~\cite{parkhi2012cats}, Places365~\cite{zhou2017places}, Stanford Cars~\cite{krause2013collecting}, SUN397~\cite{xiao2010sun} and VOC 2007~\cite{everingham2010pascal}. The semantic segmentation task is evaluated on Cityscapes~\cite{cordts2016cityscapes}, PASCAL VOC~\cite{everingham2010pascal} and NYUD~\cite{silberman2012indoor}. We use PASCAL VOC~\cite{everingham2010pascal} for object detection. The DAVIS-2017 benchmark~\cite{davis2017} is used for video instance segmentation. Finally, depth estimation is performed on NYUD~\cite{silberman2012indoor}. The model, i.e. a ResNet-50 backbone, is pretrained for 400 epochs using batches of size 256. The initial learning rate is set to $0.3$ and decayed using a cosine schedule. We use the default values for the temperature ($\tau=0.2$) and momentum ($m=0.999$) hyperparameters. 

\newcommand{\tabtop}[1]{\fontsize{7.5pt}{1em}\selectfont \textbf{#1}}
\newcommand{\tabelem}[1]{{#1}}
\begin{table}
    \caption[Overview of the pretraining datasets]{Overview of the pretraining datasets. We sample a uniform and long-tailed (LT) subset of 118K images from ImageNet. On OpenImages, we sample a random subset of 118K images. The complete train splits are used for COCO and BDD100K.}
    \label{tab: ssl_datasets}
    \begin{tabular}{lccccc}
    \toprule
    \tabtop{Pretrain Data} & ~ & \tabtop{\#Imgs} & \tabtop{\#Obj/Img} & \tabtop{Uniform} & \tabtop{Discriminative} \\
    \hline 
    \tabelem{ImageNet-118K~\cite{deng2009imagenet}} & ~ & \tabelem{118 K} & \tabelem{1.7} & \tabelem{$\checkmark$} & $\checkmark$ \\
    \tabelem{ImageNet-118K-LT~\cite{deng2009imagenet}} & ~ & \tabelem{118 K} & \tabelem{1.7} & \tabtop{\xmark} & \tabelem{$\checkmark$} \\
    \tabelem{COCO~\cite{lin2014microsoft}} & ~ & \tabelem{118 K} & \tabelem{7.3} & \tabtop{\xmark} & \tabelem{$\checkmark$} \\
    \tabelem{OpenImages-118K~\cite{kuznetsova2020open}} & ~ & \tabelem{118 K} & \tabelem{8.4} & \tabtop{\xmark} & \tabelem{$\checkmark$} \\
    \tabelem{BDD100K~\cite{yu2020bdd100k}} & ~ & \tabelem{90 K} & - & \tabtop{\xmark} & \tabtop{\xmark} \\
    \bottomrule
    \end{tabular}
\end{table}

\newpage
\begin{landscape}

\newcommand{\hr}[1]{\textcolor{red}{#1}}
\newcommand{\hg}[1]{\textcolor{green}{#1}}

\newcommand{\ressup}[2]{\tablestyle{1pt}{1} 
\begin{tabular}{z{16}y{18}} {#1} & {} \end{tabular}}

\newcommand{\ressupnyud}[2]{\tablestyle{1pt}{1} 
\begin{tabular}{z{20}y{28}} {#1} & {} \end{tabular}}

\newcommand{\res}[3]{
\tablestyle{1pt}{1}
\begin{tabular}{z{16}y{18}}
{#1} & \fontsize{7.0pt}{1em}\selectfont{~(${#2}${#3})}
\end{tabular}}

\newcommand{\reshg}[3]{
\tablestyle{1pt}{1} 
\begin{tabular}{z{16}y{18}}
{#1} &
\fontsize{7.0pt}{1em}\selectfont{~\hg{(${#2}$\textbf{#3})}}
\end{tabular}}

\newcommand{\reshgnyud}[3]{
\tablestyle{1pt}{1} 
\begin{tabular}{z{20}y{28}}
{#1} &
\fontsize{7.0pt}{1em}\selectfont{~\hg{(${#2}$\textbf{#3})}}
\end{tabular}}

\newcommand{\reshn}[3]{
\tablestyle{1pt}{1} 
\begin{tabular}{z{16}y{18}}
{#1} &
\fontsize{7.0pt}{1em}\selectfont{~(${#2}$\textbf{#3})}
\end{tabular}}

\newcommand{\reshr}[3]{
\tablestyle{1pt}{1} 
\begin{tabular}{z{16}y{18}}
{#1} &
\fontsize{7.0pt}{1em}\selectfont{~\hr{(${#2}$\textbf{#3})}}
\end{tabular}}

\newcommand{\demph}[1]{\textcolor{Gray}{#1}}
\newcommand{\resrand}[2]{\tablestyle{1pt}{1} \begin{tabular}{z{16}y{18}} \demph{#1} & {} \end{tabular}}

\begin{table}[!t]
    \label{tab: contrastive_in_the_wild}
    \centering
    \caption[Comparison of linear classification models using different pretraining datasets]{Comparison of linear classification models trained on top of frozen features (400 epochs pretraining).}
    \label{tab: ssl_linear_classification}
    \begin{tabular}{y{54}|x{46}|x{46}|x{46}|x{46}|x{46}|x{46}|x{46} c}
    \toprule
    \tabtop{Pretrain Data} & \tabtop{CIFAR10} & \tabtop{Cars} & \tabtop{Food} & \tabtop{Pets} & \tabtop{Places} & \tabtop{SUN} & \tabtop{VOC} \\
    \hline
    \tabelem{IN-118K} & \ressup{83.1}{} & \ressup{35.9}{} & \ressup{62.2}{} & \ressup{68.9}{} & \ressup{45.0}{} & \ressup{50.0}{} & \ressup{75.8}{} & ~\\
    \tabelem{COCO} & \reshr{77.4}{-}{5.7} & \reshr{33.9}{-}{2.0} & \reshr{62.0}{-}{0.2} & \reshr{62.6}{-}{6.3} & \reshg{47.3}{+}{2.3} & \reshg{53.6}{+}{3.6} & \reshg{80.9}{+}{5.1} & ~\\
    \tabelem{OI-118K} & \reshr{74.0}{-}{9.1} & \reshr{32.2}{-}{3.7} & \reshr{58.4}{-}{3.8} & \reshr{59.3}{-}{9.6} & \reshg{46.6}{+}{1.6} & \reshg{52.3}{+}{2.3} & \reshg{75.9}{+}{0.1} & ~\\
    \tabelem{IN-118K-LT} & \reshg{84.0}{+}{0.9} & \reshg{37.2}{+}{1.3} & \reshg{62.3}{+}{0.1} & \reshg{72.4}{+}{3.5} & \reshg{45.7}{+}{0.7} & \reshg{50.7}{+}{0.7} & \reshg{77.6}{+}{1.8} &  ~\\
    \bottomrule
    \end{tabular}
\end{table}

\begin{table}[!t]
    \caption[Comparison of representations under the transfer learning setup when using different pretraining datasets]{Comparison of different representations under the transfer learning setup (400 epochs pretraining).}
    \label{tab: ssl_transfer_learning}
    \centering
    \begin{tabular}{ry{28}|x{44}|x{44}|x{44}|x{50}|x{62}|x{50} c}
    \toprule
    ~ & ~ & \multicolumn{3}{c|}{\fontsize{7.5pt}{1em}\selectfont \textbf{Semantic seg. } (mIoU)} &
    {\fontsize{7.5pt}{1em}\selectfont \textbf{Detection} (AP)} &
    {\fontsize{7.5pt}{1em}\selectfont \textbf{Vid. seg.~ ($\mathcal{J}\&\mathcal{F})$}} & {\fontsize{7.5pt}{1em}\selectfont \textbf{Depth} (rmse)} & \\
    \multicolumn{2}{l|}{{\fontsize{7.5pt}{1em}\selectfont \textbf{Pretrain Data}}} &
    {\fontsize{7.5pt}{1em}\selectfont \textbf{VOC}} &
    {\fontsize{7.5pt}{1em}\selectfont \textbf{Cityscapes}} & 
    {\fontsize{7.5pt}{1em}\selectfont \textbf{NYUD}} &
    {\fontsize{7.5pt}{1em}\selectfont \textbf{VOC}} &
    {\fontsize{7.5pt}{1em}\selectfont \textbf{DAVIS}} &  
    {\fontsize{7.5pt}{1em}\selectfont \textbf{NYUD}} & \\
    \hline
    \multicolumn{2}{l|}{IN-118K} & \ressup{68.9}{} & \ressup{70.1}{} & \ressup{37.7}{} & \ressup{53.0}{} & \ressup{63.5}{} & \ressupnyud{0.625}{} & \\
    \multicolumn{2}{l|}{COCO} & \reshg{69.1}{+}{0.2} & \reshg{70.3}{+}{0.2} & \reshg{39.3}{+}{1.6} & \res{53.0}{+}{0.0} & \reshg{65.1}{+}{1.6} & \reshgnyud{0.612}{-}{0.013} & \\
    \multicolumn{2}{l|}{OI-118K} & \reshr{67.9}{-}{1.0} & \reshg{70.9}{+}{0.8} & \reshg{38.4}{+}{0.7} & \reshg{53.1}{+}{0.1} & \reshg{64.8}{+}{1.3} & \reshgnyud{0.609}{-}{0.016} & \\
    \multicolumn{2}{l|}{IN-118K-LT} & \reshg{69.2}{+}{0.3} & \reshg{70.2}{+}{0.1} & \reshg{38.5}{+}{0.8} & \reshg{53.1}{+}{0.1} & \reshg{63.7}{+}{0.2} & \reshgnyud{0.610}{-}{0.015} & \\
    \multicolumn{2}{l|}{BDD100K} & - & \res{70.1}{+}{0.0} & - & - & - & - & ~\\
    \bottomrule
    \end{tabular}
\end{table}

\end{landscape}

\subsection{Object-centric Versus Scene-centric}
\label{subsec: ssl_contrastive_wild_object_scene}
We compare the representations trained on an object-centric dataset - i.e. ImageNet (IN-118K) - against the ones obtained from two scene-centric datasets - i.e. COCO and OpenImages (OI-118K). Tables~\ref{tab: ssl_linear_classification}-\ref{tab: ssl_transfer_learning} show the results under the linear classification and transfer learning setup. 

\paragraph{Results.} The linear classification model yields better results on CIFAR10, Cars, Food and Pets when pretraining the representations on ImageNet. Differently, when tackling the classification task on Places, SUN and VOC, the representations from COCO and OpenImages are better suited. The first group of target benchmarks contains images centered around a single object, while the second group contains scene-centric images with multiple objects. We conclude that, for linear classification, the pretraining dataset should match the target dataset in terms of being object- or scene-centric.

Next, we consider finetuning. Perhaps surprisingly, we do not observe any significant disadvantages when using more complex images from COCO or OpenImages for pretraining. In particular, for the reported tasks, the COCO pretrained model even outperforms its ImageNet counterpart. We made a similar observation when increasing the size of the pretraining dataset (see~\cite{van2021revisiting}).

\begin{figure}
\begin{minipage}[b]{.32\linewidth} 
\centering
\includegraphics[width=\textwidth]{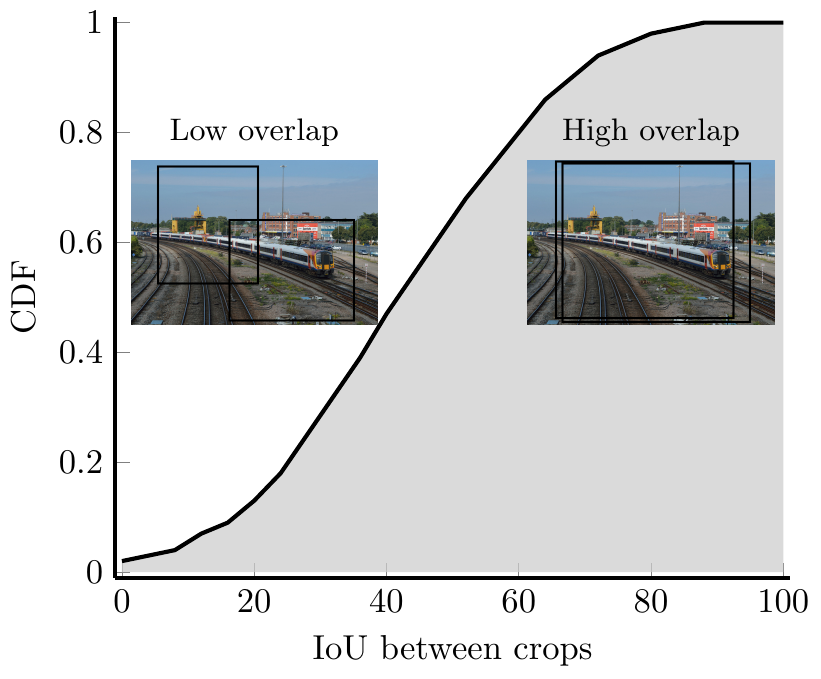}
\caption[IoU between random resized crops in self-supervised contrastive frameworks]{IoU between random resized crops for existing frameworks.}
\label{fig: ssl_area_histogram}
\end{minipage} %
\hfill
\begin{minipage}[b]{.32\linewidth} 
\centering
\includegraphics[width=\textwidth]{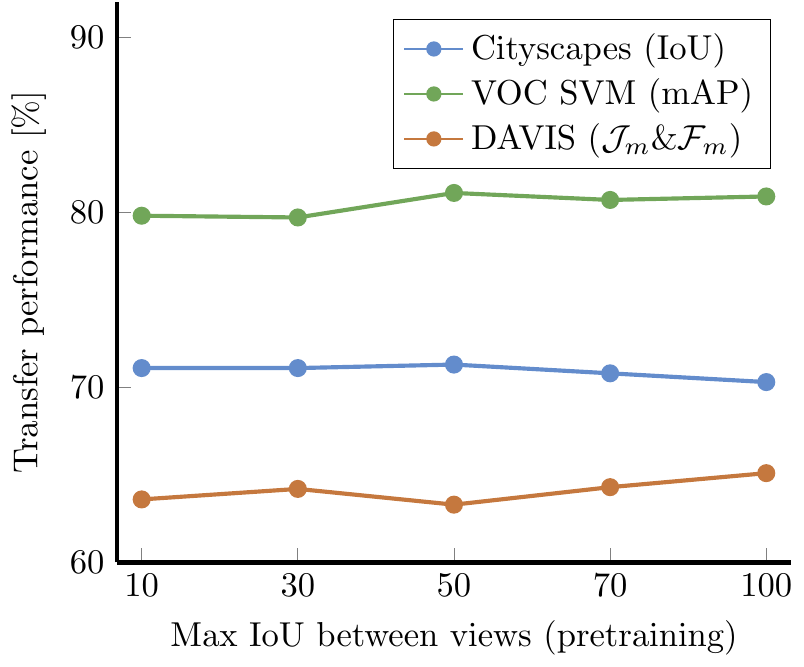}
\caption[Transfer results when limiting the IoU between crops in self-supervised contrastive frameworks]{Transfer results when thresholding the IoU between crops.}
\label{fig: ssl_transfer_iou}
\end{minipage} %
\hfill
\begin{minipage}[b]{.32\linewidth} 
\centering
\includegraphics[width=\textwidth]{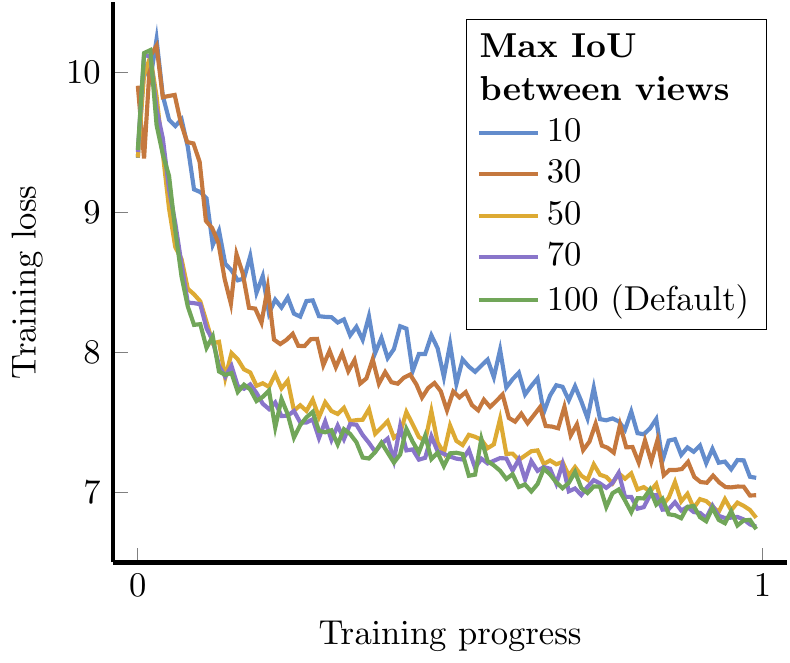}
\caption[Training curves when limiting the IoU between crops in self-supervised contrastive frameworks]{Training curves when thresholding the IoU between crops.}
\label{fig: ssl_loss_iou}
\end{minipage} %
\end{figure}

\paragraph{Discussion.} In contrast to prior belief~\cite{purushwalkam2020demystifying,selvaraju2020casting}, our results indicate that contrastive self-supervised methods do not suffer from pretraining on scene-centric datasets. How can we explain this inconsistency with earlier studies? First, the experimental setup in~\cite{purushwalkam2020demystifying} only considered the linear evaluation protocol for an object-centric dataset (i.e. PASCAL cropped boxes). This analysis~\cite{purushwalkam2020demystifying} does not show us the full picture. Second, the authors conclude that existing methods suffer from using scene-centric datasets due to the augmentation strategy, which involves random cropping. They argue that aggressive cropping could yield non-overlapping views which contain different objects. In this case, maximizing the feature similarity would be detrimental for object recognition tasks. However, the default cropping strategy barely yields non-overlapping views\footnote{We use the \texttt{RandomResizedCrop} in PyTorch with scaling $s=(0.2,1.0)$ and output size $224\times224$.}. This is verified by plotting the intersection over union (IoU) between pairs of crops (see Figure~\ref{fig: ssl_area_histogram}). 
We conclude that the used example of non-overlapping crops~\cite{purushwalkam2020demystifying,selvaraju2020casting} seldom occurs. 

The above observation motivates us to reconsider the importance of using overlapping views. We pretrain on COCO while forcing the IoU between views to be smaller than a predefined threshold. Figures~\ref{fig: ssl_transfer_iou}-\ref{fig: ssl_loss_iou} show the transfer performance and training curves for different values of the threshold. The optimization objective is harder to satisfy when applying more aggressive cropping (i.e. the training loss increases when lowering the IoU). However, Figure~\ref{fig: ssl_transfer_iou} shows that the transfer performance remains stable. Patches within the same image were observed at the same point in time and space, which means that they will share information like the camera viewpoint, color, shape, etc. As a result, the learning signal is still meaningful, even when less overlapping crops are used as positives.

\subsection{Uniform Versus Long-tailed}
\label{subsec: ssl_contrastive_wild_uniform}
Next, we study whether MoCo benefits from using a uniform (IN-118K) versus long-tailed (IN-118K-LT) dataset. We adopt the sampling strategy from~\cite{liu2019large} to construct a long-tailed version of ImageNet. The classes follow the Pareto distribution with power value $\alpha=6$. Tables~\ref{tab: ssl_linear_classification}-\ref{tab: ssl_transfer_learning} indicate that MoCo is robust to changes in the class distribution of the dataset. In particular, the IN-118K-LT model performs on par or better compared to its IN-118K counterpart across all tasks. We conclude that it is not essential to use a uniformly distributed dataset for pretraining. 

\subsection{Domain-Specific Datasets}
\label{subsec: ssl_domain_specific}
Images from ImageNet have discriminative visual features, e.g. the included bird species can be recognized from their beak or plumage. In this case, the model could achieve high performance on the instance discrimination task by solely focusing on the most discriminative feature in the image. Differently, in urban scene understanding, we deal with more monotonous scenery - i.e. all images contain parked cars, lane markings, pedestrians, etc. We try to measure the impact of using less discriminative images for representation learning. 

We pretrain on BDD100K - a dataset for urban scene understanding. Table~\ref{tab: ssl_transfer_learning} compares the representations for the semantic segmentation task on Cityscapes when pretraining on BDD100K versus IN-118K. The BDD100K model performs on par with the IN-118K pretrained baseline (BDD100K: $+0.00 \%$). This shows that MoCo can be applied to domain-specific data as well. Two models trained on very different types of data perform equally well. A recent study~\cite{zhao2020makes} showed that it is mostly the low- and mid-level visual features that are retained under the transfer learning setup. This effect explains how two different representations produce similar results. Finally, we expect that further gains can be achieved by finetuning the augmentation strategy using domain knowledge. 

\section{Learning Invariances}
\label{sec: ssl_invariances}
In Section~\ref{sec: ssl_contrastive_wild}, we showed that MoCo works well for a large variety of pretraining datasets: scene-centric, long-tailed and domain-specific images. This result shows that there is no obvious need to use a more advanced pretext task~\cite{selvaraju2020casting} or use an alternative source of data like video~\cite{purushwalkam2020demystifying} to improve the representations. Therefore, instead of developing a new framework, we try to realize further gains while sticking with the approach from MoCo. We realize this objective by learning additional invariances. We study three such mechanisms, i.e. multi-scale constrained cropping, stronger augmentations and the use of nearest neighbors. We concentrate on the implementation details and uncover several interesting qualities of the learned representations. Table~\ref{tab: overview_invariances} shows an overview of the results. 

\subsection{Multi-Scale Constrained Cropping}
\label{subsec: multi_crop}
The employed cropping strategy proves crucial to obtain powerful visual representations. So far, we used the default two-crop transform, which samples two image patches to produce an anchor and a positive. The analysis in Section~\ref{sec: ssl_contrastive_wild} showed that this transformation yields highly-overlapping views. As a result, the model can match the anchor with its positive by attending only to the most discriminative image component, while ignoring other possibly relevant regions. This observation leads us to revisit the \texttt{multi-crop} augmentation strategy from~\cite{caron2020unsupervised}.

\begin{landscape}
\definecolor{darkF7E0D5}{RGB}{0,0,0}
\newcommand{\rownumber}[1]{\textcolor{darkF7E0D5}{#1}}
\newcommand{\demph}[1]{\textcolor{Gray}{#1}}

\newcommand{\hl}[1]{\textcolor{Highlight}{#1}}

\newcommand{\res}[3]{
\tablestyle{1pt}{1}
\begin{tabular}{z{20}y{22}}
{#1} &
\fontsize{7.0pt}{1em}\selectfont{~(${#2}${#3})}
\end{tabular}}

\newcommand{\reshl}[3]{
\tablestyle{1pt}{1} 
\begin{tabular}{z{20}y{22}}
{#1} &
\fontsize{7.0pt}{1em}\selectfont{~\hl{(${#2}$\textbf{#3})}}
\end{tabular}}

\newcommand{\resrand}[2]{\tablestyle{1pt}{1} \begin{tabular}{z{20}y{22}} \demph{#1} & {} \end{tabular}}
\newcommand{\ressup}[2]{\tablestyle{1pt}{1} 
\begin{tabular}{z{20}y{22}} {#1} & {} \end{tabular}}

\begin{table}
\tablestyle{2pt}{1.1}
\caption[Ablation study of learning additional invariances]{\textbf{Ablation of different components.} Models are pretrained for 200 epochs on COCO using the settings from Section~\ref{sec: ssl_contrastive_wild}. We indicate the differences with MoCo. The best model is trained with additional invariances.}
\label{tab: ssl_overview_invariances}
\begin{tabular}{ry{29}|x{18}x{18}x{18}x{18}x{18}|x{44}|x{44}|x{44}|x{44}|x{44}|x{44} c}
\toprule
\multicolumn{2}{c|}{} &
\multicolumn{5}{c|}{\fontsize{7.5pt}{1em}\selectfont \textbf{Setup} } &
\multicolumn{3}{c|}{\fontsize{7.5pt}{1em}\selectfont \textbf{Semantic seg. } (mIoU)} & 
\multicolumn{3}{c}{\fontsize{7.5pt}{1em}\selectfont \textbf{Classification} (mAP / Acc. / Acc.)}  
\\
\multicolumn{2}{l|}{Method} & 
{\fontsize{7.5pt}{1em}\selectfont \textbf{MC}} &
{\fontsize{7.5pt}{1em}\selectfont \textbf{CC}} &
{\fontsize{7.5pt}{1em}\selectfont \textbf{m$\downarrow$}} &
{\fontsize{7.5pt}{1em}\selectfont \textbf{A$^+$}} &
{\fontsize{7.5pt}{1em}\selectfont \textbf{NN}} &
{\fontsize{7.5pt}{1em}\selectfont \textbf{VOC}} &
{\fontsize{7.5pt}{1em}\selectfont \textbf{Cityscapes}} & 
{\fontsize{7.5pt}{1em}\selectfont \textbf{NYUD}} &
{\fontsize{7.5pt}{1em}\selectfont \textbf{VOC}} &
{\fontsize{7.5pt}{1em}\selectfont \textbf{ImageNet}} &
{\fontsize{7.5pt}{1em}\selectfont \textbf{Places}} &

\\
\hline
\multicolumn{2}{l|}{\demph{Rand. init.}}
& \demph{-} & \demph{-} & \demph{-} & \demph{-} & \demph{-}
& \resrand{39.2}{} 
& \resrand{65.0}{}  
& \resrand{24.4}{}  
& \resrand{-}{}  
& \resrand{-}{}  
& \resrand{-}{}  
& \\
\multicolumn{2}{l|}{MoCo}
& \xmark & \xmark & \xmark & \xmark & \xmark
& \ressup{66.2}{} 
& \ressup{70.3}{}  
& \ressup{38.2}{}  
& \ressup{76.1}{}  
& \ressup{49.3}{}  
& \ressup{45.1}{}  
& \\

\multicolumn{2}{l|}{\textcolor{red}{Sec.~\ref{subsec: multi_crop}}}
& \checkmark & \xmark & \xmark & \xmark & \xmark 
& \res{69.9}{+}{3.7}  
& \res{70.9}{+}{0.6}  
& \res{39.4}{+}{1.2}  
& \res{81.3}{+}{5.2}  
& \res{53.4}{+}{4.1}  
& \res{47.7}{+}{2.6}  
& \\
\multicolumn{2}{l|}{}
& \checkmark & \checkmark & \xmark & \xmark & \xmark
& \res{70.2}{+}{4.0} 
& \res{70.9}{+}{0.6}  
& \res{39.5}{+}{1.3}  
& \res{82.1}{+}{6.0}  
& \res{54.0}{+}{4.7}  
& \res{47.9}{+}{2.8}  
& \\
\multicolumn{2}{l|}{}
& \checkmark & \checkmark & \checkmark & \xmark & \xmark
& \res{70.9}{+}{4.7} 
& \res{71.3}{+}{1.0}  
& \res{39.9}{+}{1.7}  
& \res{82.8}{+}{6.7}  
& \res{54.8}{+}{5.5}  
& \res{48.1}{+}{3.0}  
& \\
\multicolumn{2}{l|}{\textcolor{red}{Sec.~\ref{subsec: auto_augment}}}
& \checkmark & \checkmark & \checkmark & \checkmark & \xmark
& \res{71.4}{+}{5.2} 
& \res{72.0}{+}{1.7}  
& \res{40.0}{+}{1.8}  
& \res{83.7}{+}{7.6}  
& \res{55.5}{+}{6.2}  
& \res{48.2}{+}{3.1}  
& \\
\multicolumn{2}{l|}{\textcolor{red}{Sec.~\ref{subsec: neighbors}}}
& \checkmark & \checkmark & \checkmark & \checkmark & \checkmark
& \reshl{71.9}{+}{5.7} 
& \reshl{72.2}{+}{1.9}  
& \reshl{40.9}{+}{2.7}  
& \reshl{85.1}{+}{9.0}  
& \reshl{55.9}{+}{6.6}  
& \reshl{48.5}{+}{3.4}  
& \\
\bottomrule
\multicolumn{14}{c}{\fontsize{7.5pt}{1em}\selectfont MC: Multi-crop, CC: Constrained multi-crop, m$\downarrow$: Lower momentum, A$^+$: Stronger augmentations, NN: Nearest neighbors}\\
\end{tabular}
\end{table}
\end{landscape}

\begin{figure}[!t]
    \begin{minipage}[t]{.62\textwidth}
    \includegraphics[width=\linewidth]{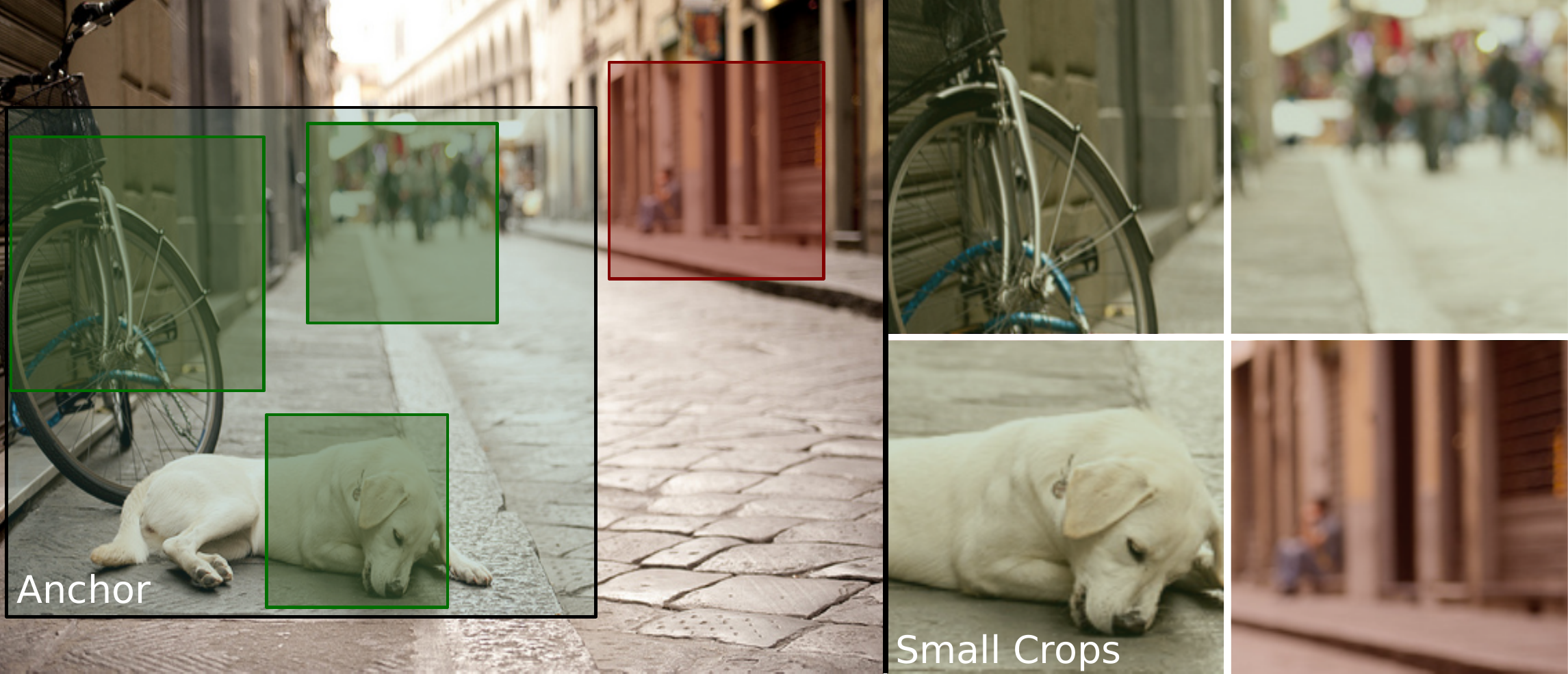}
    \caption[Visualization of the constrained multi-crop transform]{The constrained multi-crop. Smaller crops are forced to lie within the anchor image. Invalid/valid crops are colored in red/green.}
    \label{fig: ssl_multiple_crops}
    \end{minipage} 
    \hfill
    \begin{minipage}[t]{.36\textwidth}
    \centering
    \includegraphics[width=0.85\linewidth]{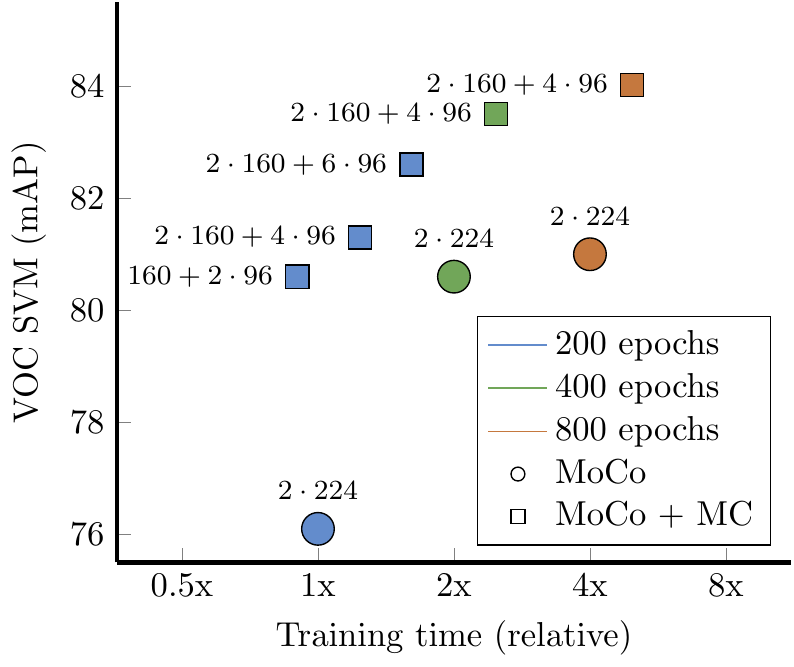}
    \caption[Transfer performance versus time for the multi-crop transform]{Transfer performance versus time for the multi-crop (MC) model.}
    \label{fig: ssl_multicrop_resource}
    \end{minipage}
\end{figure}

\paragraph{Setup} The \texttt{multi-crop} transform samples $N$ additional positive views for each anchor. Remember, the views are generated by randomly applying resized cropping, color distortions, horizontal flipping and Gaussian blur. We adjust the scaling parameters and resolution of the random crop transform to limit the computational overhead. The two default crops (i.e. the anchor and a positive) contain between $20\%$ and $100\%$ of the original image area, and are resized to $160\times160$ pixels. Differently, the $N$ additional views contain between $5\%$ and $14\%$ of the original image area, and are resized to $96\times96$ pixels. We maximize the agreement between all $N+1$ positives and the anchor. Note that the $N$ smaller, more aggressively scaled crops are not used as anchors, i.e. they are not shown to the momentum-updated encoder and are not used as negatives in the memory bank. 

Further, we consider two additional modifications. First, we enforce the smaller crops to overlap with the anchor image. In this way, all positives can be matched with their anchor as there is shared information. The new transformation is referred to as \texttt{constrained multi-cropping}. Figure~\ref{fig: ssl_multiple_crops} illustrates the concept. Second, since increasing the number of views facilitates faster training, we reduce the momentum hyperparameter $m$ from $0.999$ to $0.995$. We pretrain for 200 epochs on COCO.

\paragraph{Results} Table~\ref{tab: ssl_overview_invariances} benchmarks the  modifications. Both the \texttt{multi-crop} (MC) and \texttt{constrained multi-crop} (CC) significantly improve the transfer performance. Reducing the momentum (m$\downarrow$) yields further gains. Figure~\ref{fig: ssl_multicrop_resource} plots the training time versus performance for the \texttt{multi-crop} model. The \texttt{multi-crop} improves the computational efficiency w.r.t. the two-crop transform. 

\begin{figure}
    \centering
    \includegraphics[width=.9\linewidth]{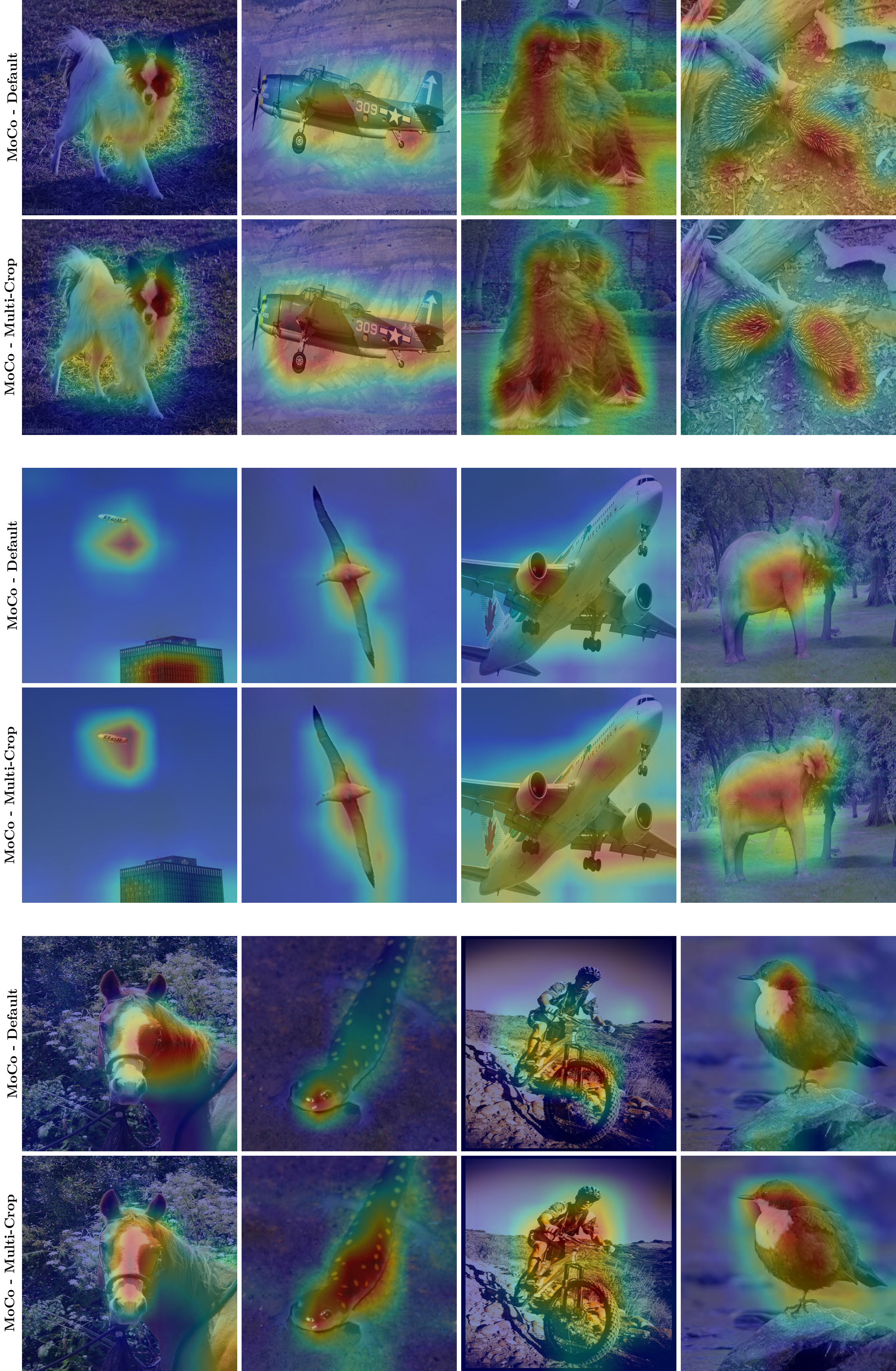}
    \caption{Class activation maps for the two-crop versus multi-crop model.}
    \label{fig: cams}
\end{figure}

\paragraph{Discussion} We investigate what renders the \texttt{multi-crop} and \texttt{constrained multi-crop} effective. The network matches the representations of the large anchor image with the small crops. Since the small crops show random subparts of the anchor image, the network is forced to encode as much information as possible, such that all image regions can be retrieved from the representation of the scene.
As a result, the representations will be informative of the spatial layout and different objects in the image - rather than attend only to the most discriminative image component. The latter is verified by visualizing the class activation maps (CAMs)~\cite{zhou2016learning} of different representations (see Figure~\ref{fig: cams}). The CAMs of the \texttt{multi-crop} model segment the complete object, while the \texttt{two-crop} model only looks at a few discriminative components.

Further, the representations are evaluated on two dense prediction tasks without any finetuning. We consider the tasks of video instance segmentation on DAVIS 2017~\cite{davis2017} and semantic segment retrieval on VOC~\cite{everingham2010pascal}. The \texttt{multi-crop} model is found superior at modeling dense correspondences for the video instance segmentation task in Table~\ref{table: davis} (MoCo vs. MoCo + MC). Similarly, Table~\ref{tab: voc} shows that the \texttt{multi-crop} model achieves higher performance on the semantic segment retrieval task compared to its two-crop counterpart. This indicates that the pixel embeddings are better disentangled according to the semantic classes. Finally, the MoCo \texttt{multi-crop} model is competitive with other methods that were specifically designed for the tasks. We conclude that the \texttt{multi-crop} setup provides a viable alternative to learn dense representations without supervision. Moreover, this setup does not rely on video~\cite{lai2020mast,jabri2020walk} or handcrafted priors~\cite{hwang2019segsort,zhang2020self,van2021unsupervised}.

\begin{table}
\caption[Results for the DAVIS 2017 video instance segmentation task]{\textbf{DAVIS 2017 video instance segmentation}. We use the publicly available code from~\cite{jabri2020walk} to evaluate our frozen representations. Qualitative results are shown for MoCo trained with the \texttt{multi-crop} (MC) transform.} 
\begin{minipage}{1.0\textwidth}
    \tablestyle{2pt}{1.1}
    \begin{tabular}{y{64} |x{44} | x{24} x{24} c}
    \toprule
    \fontsize{7.5pt}{1em}\selectfont \textbf{Method} &
    \fontsize{7.5pt}{1em}\selectfont \textbf{Data} &
    \fontsize{7.5pt}{1em}\selectfont $\mathcal{J}_m \uparrow$ & \fontsize{7.5pt}{1em}\selectfont $\mathcal{F}_m \uparrow$ & \\
    \hline
    \fontsize{7.5pt}{1em}\selectfont DenseCL~\cite{wang2020dense} &     \fontsize{7.5pt}{1em}\selectfont COCO &  60.6 & 63.9 & \\
    \fontsize{7.5pt}{1em}\selectfont MAST~\cite{lai2020mast} & \fontsize{7.5pt}{1em}\selectfont YT-VOS & 63.3 & 67.6 & \\
    \fontsize{7.5pt}{1em}\selectfont STC~\cite{jabri2020walk} & \fontsize{7.5pt}{1em}\selectfont Kinetics & 64.8 & 70.2 & \\
    \hline
    \fontsize{7.5pt}{1em}\selectfont MoCo & \fontsize{7.5pt}{1em}\selectfont COCO & 61.6 & 66.6 & \\
    \fontsize{7.5pt}{1em}\selectfont MoCo + MC & \fontsize{7.5pt}{1em}\selectfont COCO & 64.3 & 69.4 & \\
    \bottomrule
    \multicolumn{5}{c}{}\\
    \end{tabular}
\end{minipage} %
\vspace{5em}
\begin{minipage}{1.0\textwidth}
\centering
\includegraphics[width=\linewidth]{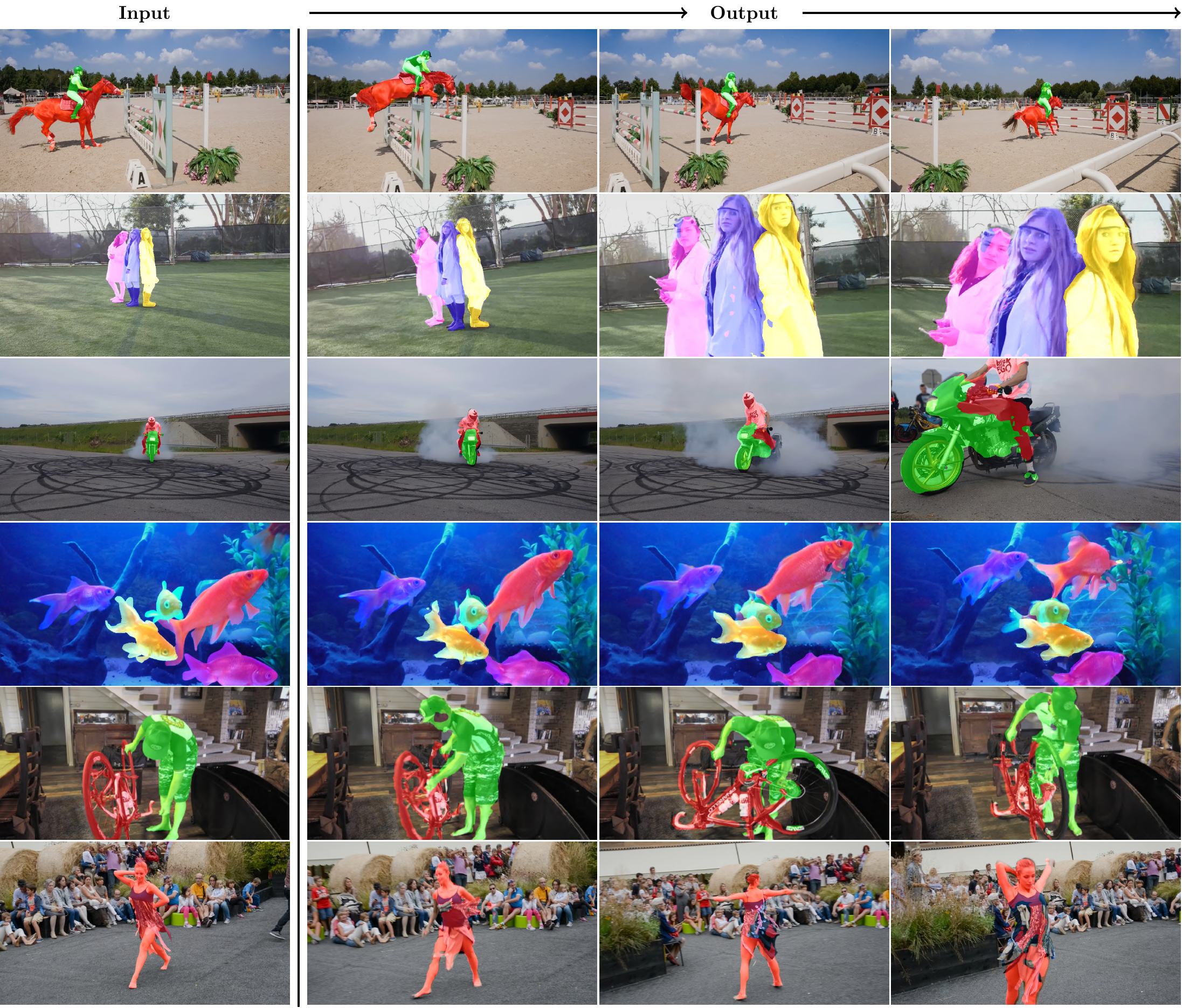}
\end{minipage}
\label{table: davis}
\end{table}

\begin{table}
\caption[Results for the PASCAL VOC semantic segment retrieval task]{\textbf{VOC semantic segment retrieval.} Each image is partitioned into a number of segments by running K-Means (K=15) on the spatial features. Then we adopt a region descriptor - computed by averaging the embeddings of all pixels within a segment - to obtain nearest neighbors of the validation regions from the train set. Qualitative results are shown for MoCo trained with the \texttt{multi-crop} (MC) augmentation strategy (K=5).}
\label{tab: voc}
\begin{minipage}{1.0\textwidth}
    \tablestyle{2pt}{1.1}
    \begin{tabular}{y{44}|x{52} x{54} c}
    \toprule
    & \multicolumn{2}{c}{\fontsize{7.5pt}{1em}\selectfont \textbf{Segm. Retr. VOC}} & \\
    \fontsize{7.5pt}{1em}\selectfont \textbf{Method} & \fontsize{7.5pt}{1em}\selectfont 7 classes (IoU)$\uparrow$ & \fontsize{7.5pt}{1em}\selectfont 21 classes (IoU)$\uparrow$ & \\
    \hline
    \fontsize{7.5pt}{1em}\selectfont SegSort~\cite{zhang2020self}
    & 10.2 & - & \\
    \fontsize{7.5pt}{1em}\selectfont SSL HG~\cite{zhang2020self}
    & 24.6 & - & \\
    \fontsize{7.5pt}{1em}\selectfont DenseCL~\cite{wang2020dense}
    & 48.4 & 35.1 & \\
    \hline
    \fontsize{7.5pt}{1em}\selectfont MoCo & 41.8 & 28.1 & \\
    \fontsize{7.5pt}{1em}\selectfont MoCo + MC & 48.1 & 35.1 & \\
    \bottomrule
    \multicolumn{4}{c}{}\\
    \end{tabular}
\end{minipage}
\begin{minipage}{1.0\textwidth}
\centering
\includegraphics[width=.8\linewidth]{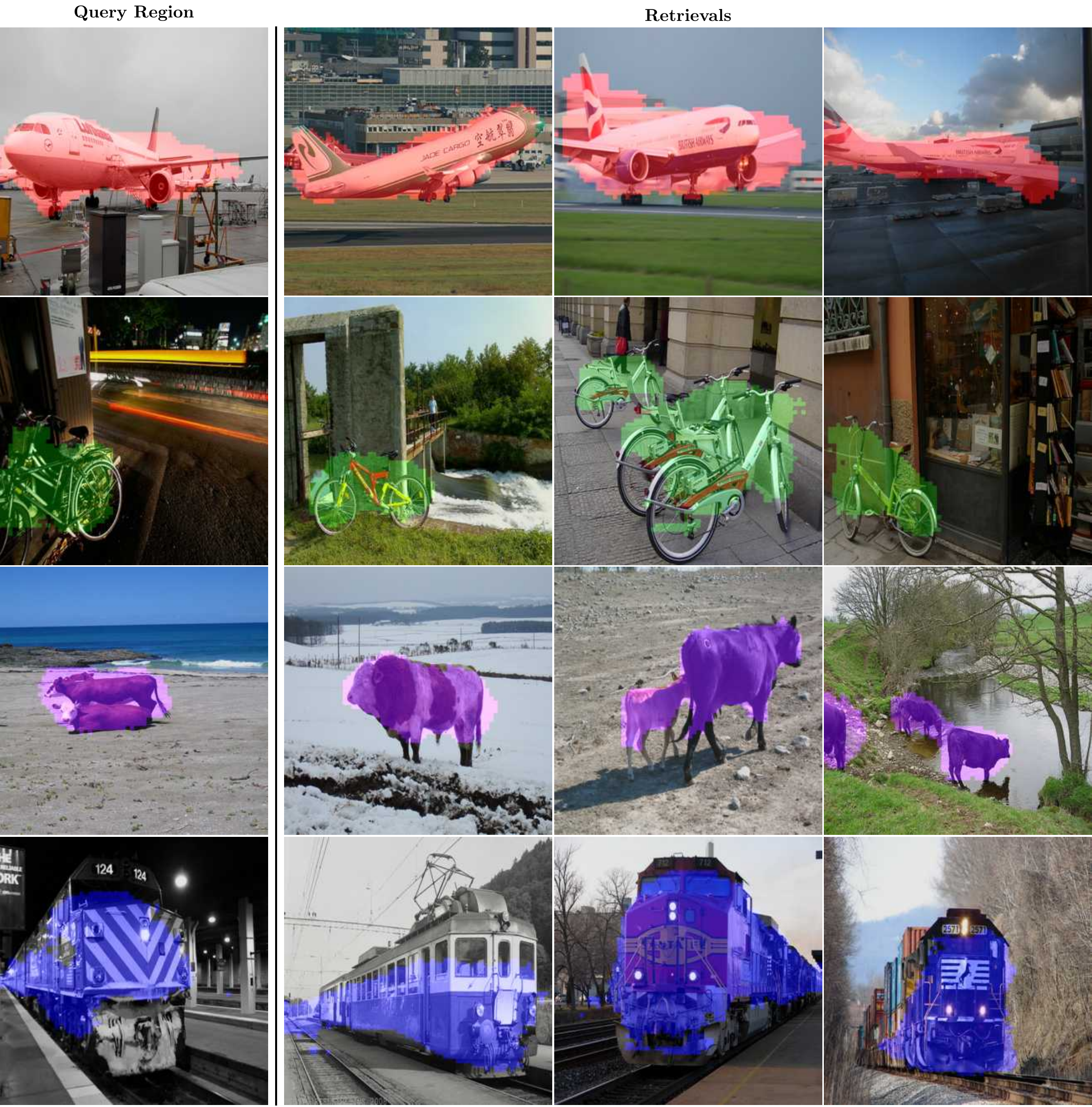}
\end{minipage}
\end{table}

\subsection{Stronger Augmentations}
\label{subsec: auto_augment}
As mentioned before, the image crops are augmented by random color distortions, horizontal flips and Gaussian blur. Can we do better by using stronger augmentations? We investigate the use of AutoAugment~\cite{cubuk2019autoaugment} - an advanced augmentation policy obtained using supervised learning on ImageNet. We consider three possible strategies to augment the positives: (1) standard augmentations, (2) AutoAugment and (3) randomly applying either (1) or (2). 

Table~\ref{tab: auto_augment_ablation} compares the representations under the linear evaluation protocol on PASCAL VOC. Replacing the standard augmentations with AutoAugment degrades the performance (from $82.8\%$ to $79.9\%$). However, randomly applying either the standard augmentations or AutoAugment does improve the result (from $82.8\%$ to $83.7\%$). Chen~\emph{et~al.}~\cite{chen2020simple} showed that contrastive SSL benefits from using strong color augmentations as two random crops from the same image will share a similar color distribution. AutoAugment applies fewer color distortions, resulting in lower performance. This is compensated when combining both augmentation strategies. Finally, Table~\ref{tab: ssl_overview_invariances} shows that combining our custom augmentation policy (A$^+$) with the model from Section~\ref{subsec: multi_crop} results in considerable improvements on all tasks. 

\begin{table}
    \tablestyle{2pt}{1.1}
    \caption{Ablation of augmentation policies applied to random crops.}
    \label{tab: auto_augment_ablation}
    \centering
    \begin{tabular}{y{80}|x{35}}
    \toprule
    \fontsize{7.5pt}{1em}\selectfont \textbf{Augmentation policy} & 
    \fontsize{7.5pt}{1em}\selectfont \textbf{VOC SVM} (mAP)\\
    \hline
    \fontsize{7.5pt}{1em}\selectfont Standard & 82.8 \\
    \fontsize{7.5pt}{1em}\selectfont AutoAugment & 79.9 \\
    \fontsize{7.5pt}{1em}\selectfont Standard or AutoAugment & 83.7 \\
    \bottomrule
    \end{tabular}
\end{table}

\subsection{Nearest Neighbors}
\label{subsec: neighbors}
Prior work~\cite{van2020scan} showed that the model learns to map visually similar images closer together than dissimilar ones when tackling the instance discrimination task. In this section, we build upon this insight by imposing additional invariances between neighboring samples. By leveraging other samples as positives, we can capture a rich set of deformations that are hard to model via handcrafted augmentations. However, a well-known problem with methods that group different samples together is preventing representation collapse. Therefore, we formulate our intuition as an auxiliary loss that regularizes the representations - keeping the instance discrimination task as the main force.

\paragraph{Setup} Recall that the encoder $f$ consists of a backbone $g$ and a projection head $h$. MoCo maintains a queue $\mathcal{Q}_h$ of encoded anchors $\{q^{h}_0~...~q^{h}_{K-1}\}$ processed by the momentum encoder $f'$. We now introduce a second - equally sized and aligned - queue $\mathcal{Q}_g$ which maintains the features from before the projection head  $\{q^{g}_0~...~q^{g}_{K-1}\}$. The queue of backbone features $\mathcal{Q}_g$ can be used to mine nearest neighbors on-the-fly during training. In particular, for a positive $x^+$ - processed by the encoder $f$ -  the set of its $k$ nearest neighbors~$\mathcal{N}_{x^+} = \{q^h_i~|~sim(g(x),~q_i^g)~\text{is top}~k\in\mathcal{Q}_g\}$ is computed w.r.t. the queue $\mathcal{Q}_g$. The cosine similarity measure is denoted by $sim$. Finally, we use the contrastive loss to maximize the agreement between $x^+$ and its nearest neighbors $\mathcal{N}_{x^+}$ after the projection head:
\begin{equation}
    \label{eq: nn_loss}
    \mathcal{L}_{\text{nn}} = -\sum_{x^+ \in \mathcal{X^+}}  \frac{1}{k}\sum_{q \in \mathcal{N}_{x^+}} \frac{\exp \left[q^T \cdot f(x^+) / \tau\right]}{\exp\left[ q^T \cdot f(x^+) / \tau \right] + \sum_{x^- \in \mathcal{X^-}} \exp\left[q^T \cdot f(x^-) / \tau\right]}.
\end{equation}
The total loss is the sum of the instance discrimination loss $\mathcal{L}_\text{inst}$ and the nearest neighbors loss $\mathcal{L}_\text{nn}$ from Eq.~\ref{eq: nn_loss}: $\mathcal{L_\text{inst}} + \lambda\mathcal{L}_\text{nn}$. Figure~\ref{fig: topk} shows a schematic overview of our k-Nearest Neighbors based Momentum Contrast setup (kNN-MoCo). Algorithm~\ref{alg: ssl} contains the pseudocode.

\definecolor{bluefig}{RGB}{0,80,149}
\definecolor{greenfig}{RGB}{0,79,0}
\newcommand{\blue}[1]{\textcolor{bluefig}{#1}}
\newcommand{\green}[1]{\textcolor{greenfig}{#1}}

\begin{figure}
\includegraphics[width=\linewidth]{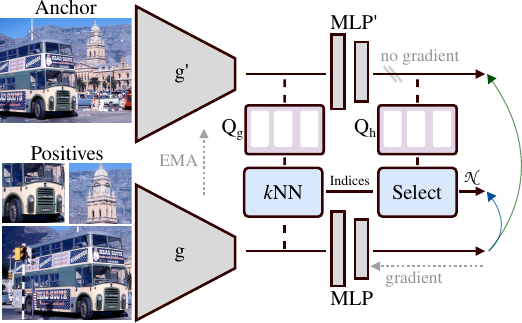}
\caption[Overview of the kNN-MoCo setup]{\textbf{kNN-MoCo setup.} $Q_g$ and $Q_h$ maintain aligned queues of backbone and output features. As before, the encoder $f$ needs to match positives with their anchor (\green{green} arrow). Additionally, we also match the positives with their $k$ nearest neighbors obtained from the queue $Q_g$ (\blue{blue} arrow).}
\label{fig: topk}
\end{figure}

\begin{algorithm}
\newcommand{\hlp}{\makebox[0pt][l]{\color{Mulberry!15}\rule[-3pt]{0.72\linewidth}{9pt}}}
\newcommand{\hlb}{\makebox[0pt][l]{\color{NavyBlue!15}\rule[-3pt]{0.72\linewidth}{9pt}}}
\algcomment{\fontsize{7.2pt}{0em}\selectfont \texttt{bmm}: batch matrix mult.; \texttt{mm}: matrix mult.; \texttt{cat}: concatenate; \texttt{topk}: indices of top-k largest elements; \texttt{CE}: cross-entropy loss \\  Important differences with MoCo are highlighted.}
\definecolor{codeblue}{rgb}{0.25,0.5,0.5}
\lstset{
  backgroundcolor=\color{white},
  basicstyle=\fontsize{7.2pt}{7.2pt}\ttfamily\selectfont,
  columns=fullflexible,
  breaklines=true,
  captionpos=b,
  commentstyle=\fontsize{7.2pt}{7.2pt}\color{codeblue},
  keywordstyle=\fontsize{7.2pt}{7.2pt},
}
\begin{lstlisting}[language=python, escapechar=@]
@\vspace{-2.0em}@
# g, g_m: the backbone g and momentum updated backbone g_m
# h, h_m: the projection head h and momentum updated projection head h_m (MLP)
# anchors: batch of anchor images (B x 3 x 224 x 224)
# positives: batch of global positive views (B x 3 x 224 x 224)
# positives_small: batch of N local positive views (B*N x 3 x 96 x 96)
# queue: dictionary as a queue (C x K)
# queue_g: dictionary as a queue (C_g x K)
# m: momentum
# t: temperature
# k: number of nearest neighbors
# lambda: weight in range [0, 1]

g_m.params = g.params # initialize momentum updated backbone
h_m.params = h.params # initialize momentum updated head

for batch in loader: # load a minibatch with B samples
   
    # randomly augment batch
    anchors @~~@= aug(batch)
    positives = aug(batch)
    @\hlb@positives_small = aug_small(batch, anchors) # constrained multi-crop @~~~~~~~~~\textcolor{red}{Sec.~4.1}@
    
    # forward pass
    anchors_g, positives_g = g_m(anchors), g(positives) # B x C_g
    anchors, positives = h_m(anchors_g), h(positives_g) # B x C
    @\hlb@positives_small_g = g(positives_small) # B*N x C_g 
    @\hlb@positives_small = h(positives_small_g) # B*N x C 
    
    # concatenate positive views
    @\hlb@positives_g = cat([positives_g, positives_small_g], dim=0) # B*(N+1) x C_g
    @\hlb@positives = cat([positives, positives_small], dim=0) # B*(N+1) x C

    # compute logits
    l_pos = bmm(positives.view(B,N+1,C), anchors.view(B,C,1)) # B x N+1 x 1
    l_neg = mm(positives, queue.view(C,K)).view(B,N+1,K) # B x N+1 x K
    logits = cat([l_pos, l_neg], dim=2) # B x N+1 x K+1
    
    # determine indices of the nearest neighbors
    @\hlp@indices = topk(mm(positives_g, queue_g).view(B,N+1,K),dim=2) # B x N+1 x k @~~~~\textcolor{red}{Sec.~4.3}@

    # loss: (1) sharpen with temperature t, (2) apply cross-entropy loss
    loss_inst = CE(logits/t, zeros((B,N+1))
    @\hlp@loss_nn = multi_label_CE(l_neg/t, indices)
    @\hlp@loss = loss_inst + lambda * loss_nn 
    
    # SGD update: g, h
    loss.backward()
    update(g.params, h.params)
    
    # momentum update: g_m, h_m
    g_m.params = m * g_m.params + (1-m) * g.params
    h_m.params = m * h_m.params + (1-m) * h.params

    # update dictionaries: queue & queue_g
    enqueue_dequeue(queue, anchors) 
    @\hlp@enqueue_dequeue(queue_g, anchors_g) 
\end{lstlisting}
\captionof{algorithm}{Pseudocode of kNN-MoCo in a PyTorch-like style.}
\label{alg: ssl}
\end{algorithm}

\begin{table}
    \caption[Ablation study of the number of neighbors $k$ and weight $lambda$ for kNN-MoCo]{Ablation study of the number of neighbors $k$ and weight $\lambda$ for a linear classifier on VOC. $\lambda=0$ represents the multi-crop model from Section~\ref{subsec: multi_crop}. Models are trained for 200 epochs on COCO.}
    \label{tab: ablation_knn}
    \begin{minipage}{0.50\textwidth}
    \begin{center}
    \tablestyle{2pt}{1.1}
    \begin{tabular}{c |x{25} x{25} x{25} x{25} x{25}}
    $k$ & 1 & 5 & 10 & 20 & 50 \\
    \shline
    mAP (\%) & 83.6 &  84.0 & 84.1 & 84.2 & 84.3 \\
    \end{tabular}
    \end{center}
    \end{minipage}
    \hfill
    \begin{minipage}{0.50\textwidth}
    \begin{center}
    \tablestyle{2pt}{1.1}
    \begin{tabular}{c |x{25} x{25} x{25} x{25} x{25}}
    $\lambda$ & 0.0 & 0.1 & 0.2 & 0.4 & 0.8 \\
    \shline
    mAP (\%) & 82.8 &  83.8 & 84.1 & 84.2 & 80.7 \\
    \end{tabular}
    \end{center}
    \end{minipage}
\end{table}

\newcommand{\demph}[1]{\textcolor{Gray}{#1}}

\begin{landscape}
\begin{table}
\tablestyle{2pt}{1.1}
\caption[State-of-the-art comparison between self-supervised pretraining methods on COCO]{\textbf{State-of-the-art comparison.} MoCo and DenseCL are trained for 800 epochs on COCO. VirTex is trained on COCO captions~\cite{chen2015microsoft}. MoCo is trained while imposing various additional invariances.\vspace{0.5em}}
\label{tab: ssl_sota}
\begin{tabular}{ry{55}|x{35}|x{35}|x{35}|x{35}|x{35}|x{35}|x{38}|x{38} c}
\toprule
\multicolumn{2}{c|}{} &
\multicolumn{3}{c|}{\fontsize{7.5pt}{1em}\selectfont \textbf{Semantic seg. } (mIoU)} & 
\multicolumn{3}{c|}{\fontsize{7.5pt}{1em}\selectfont \textbf{Classification} (mAP / Acc. / Acc.)} &
\fontsize{7.5pt}{1em}\selectfont \textbf{Vid. Seg.} ($\mathcal{J}\&\mathcal{F}$) & 
\fontsize{7.5pt}{1em}\selectfont \textbf{Depth} (rmse) & 
\\
\multicolumn{2}{l|}{{\fontsize{7.5pt}{1em}\selectfont \textbf{Method}}} & 
{\fontsize{7.5pt}{1em}\selectfont \textbf{VOC}} &
{\fontsize{7.5pt}{1em}\selectfont \textbf{Cityscapes}} & 
{\fontsize{7.5pt}{1em}\selectfont \textbf{NYUD}} &
{\fontsize{7.5pt}{1em}\selectfont \textbf{VOC}} &
{\fontsize{7.5pt}{1em}\selectfont \textbf{ImageNet}} &
{\fontsize{7.5pt}{1em}\selectfont \textbf{Places}} & 
{\fontsize{7.5pt}{1em}\selectfont \textbf{DAVIS}} &
{\fontsize{7.5pt}{1em}\selectfont \textbf{NYUD}} &
\\
\hline
\multicolumn{2}{l|}{\demph{Rand. init.}}
& \demph{39.2} 
& \demph{65.0} 
& \demph{24.4}  
& \demph{-}      
& \demph{-}      
& \demph{-}      
& \demph{40.8} 
& \demph{0.867}
& \\
\multicolumn{2}{l|}{DenseCL~\cite{wang2020dense}}
& 73.2  
& \textbf{73.5}  
& \textbf{42.1}  
& 83.5  
& 49.9  
& 45.8  
& 61.8
& 0.589
& \\
\multicolumn{2}{l|}{VirTex~\cite{desai2020virtex}}
& 72.7
& 72.5
& 40.3
& \textbf{87.4}
& 53.8
& 40.8
& 61.3
& 0.613
& \\
\hline
\multicolumn{2}{l|}{MoCo}
& 71.1  
& 71.3  
& 40.0 
& 81.0  
& 49.8  
& 44.7  
& 63.3 
& 0.606
& \\
\multicolumn{2}{l|}{+ CC}
& 72.2  
& 71.6  
& 40.4  
& 84.0  
& 54.6  
& 46.1  
& 65.5 
& 0.595
& \\
\multicolumn{2}{l|}{+ CC + A$^+$}
& 72.7 
& 71.8 
& 40.7 
& 85.0
& 56.0  
& 47.0  
& 65.7 
& 0.590
& \\
\multicolumn{2}{l|}{+ CC + A$^+$ + kNN}
& \textbf{73.5} 
& 72.3
& 41.3
& 85.9
& \textbf{56.1}
& \textbf{48.6}
& \textbf{66.2}
& \textbf{0.580}
& \\
\bottomrule
\multicolumn{11}{c}{\fontsize{7.5pt}{1em}\selectfont CC: Constrained multi-crop (Sec.~\ref{subsec: multi_crop}), A$^+$: Stronger augmentations (Sec.~\ref{subsec: auto_augment}), kNN: nearest neighbors (Sec.~\ref{subsec: neighbors}).}\\
\end{tabular}
\end{table}
\end{landscape}

\paragraph{Results \& Discussion.} Table~\ref{tab: ablation_knn} contains an ablation study of the number of neighbors $k$ and the weight $\lambda$. The performance remains stable for a large range of neighbors $k$ ($\lambda$ is fixed at 0.4). However, increasing the number of neighbors $k$ positively impacts the accuracy. We use $k=20$ for the remainder of our experiments. Further, the representation quality degrades when using a large weight (e.g. $\lambda=0.8$). This shows the importance of using the instance discrimination task as our main objective. Also, not shown in the table, we found that it is important to mine the neighbors using the features before the projection head (84.2\% vs 82.8\%). Finally, Table~\ref{tab: ssl_overview_invariances} shows improved results on all tasks when combining the nearest neighbors loss with our other modifications. In conclusion, we have successfully explored the data manifold to learn additional invariances. The proposed implementation can be seen as a simple alternative to clustering-based methods~\cite{asano20self,caron2020unsupervised,li2020prototypical,wang2020unsupervised}.

\subsection{Discussion}
\label{subsec: discussion}
We retrain our final model for 800 epochs on COCO and compare with two other methods, i.e. DenseCL~\cite{wang2020dense} and Virtex~\cite{desai2020virtex}. We draw the following conclusions from the results in Table~\ref{tab: ssl_sota}. First, our model improves over the MoCo baseline. The proposed modifications force the model to learn useful features that can not be learned through standard data augmentations, even when increasing the training time. Second, our representations outperform other works on several downstream tasks. These frameworks used more advanced schemes~\cite{wang2020dense} or caption annotations~\cite{desai2020virtex}. Interestingly, DenseCL reports better results for the segmentation tasks on Cityscapes and NYUD, but performs worse on other tasks. In contrast, the performance of our representations is better balanced across tasks. We conclude that generic pretraining is still an unsolved problem. 

\section{Related Work}
\label{sec: related_work}
\paragraph{Contrastive learning} The idea of contrastive learning~\cite{gutmann2010noise,oord2018representation} is to attract positive sample pairs and repel negative sample pairs. Self-supervised methods~\cite{chen2020simple,grill2020bootstrap,he2019momentum,misra2020self,tian2019contrastive,tian2020infomin,wu2018unsupervised,ye2019unsupervised,chen2020exploring} have used the contrastive loss to learn visual representations from unlabeled images. Augmentations of the same image are used as positives, while other images are considered as negatives. 

A number of extensions were proposed to boost the performance. For example, a group of works~\cite{wang2020dense,pinheiro2020unsupervised,van2021unsupervised,henaff2021efficient} applied the contrastive loss at the pixel-level to learn dense representations. Others improved the representations for object recognition tasks by re-identifying patches~\cite{yang2021instance} or by maximizing the similarity of corresponding image regions in the intermediate network layers~\cite{xiao2021region}. Finally, Selvaraju~\emph{et~al.}~\cite{selvaraju2020casting} employed an attention mechanism to improve the visual grounding abilities of the model. In contrast to these works, we do not employ a more advanced pretext task to learn spatially structured representations. Instead, we adopt a standard framework~\cite{he2019momentum} and find that the learned representations exhibit similar properties when modifying the cropping strategy. Further, we expect that our findings can be relevant for other contrastive learning frameworks too. 

\paragraph{Clustering} Several works combined clustering with self-labeling~\cite{caron2018deep,asano20self} or contrastive learning~\cite{caron2020unsupervised,li2020prototypical,wang2020unsupervised} to learn representations in a self-supervised way. Similar to the nearest neighbors loss (Eq.~\ref{eq: nn_loss}), these frameworks explore the data manifold to learn invariances. Differently, we avoid the use of a clustering criterion like K-Means by computing nearest neighbors on-the-fly w.r.t. a memory bank. A few other works~\cite{huang2019unsupervised,van2020scan} also used nearest neighbors as positive pairs in an auxiliary loss. However, the neighbors had to be computed off-line at fixed intervals during training. Concurrent to our work, Dwibedi~\emph{et~al.}~\cite{dwibedi2021little} adopted nearest neighbors from a memory bank under the BYOL~\cite{grill2020bootstrap} framework. The authors focus on image classification datasets. In conclusion, we propose a simple, yet effective alternative to existing clustering methods. 

\paragraph{Other} Contrastive SSL has been the subject of several recent surveys~\cite{purushwalkam2020demystifying,zhao2020makes,ericsson2020well}. We list the most relevant ones. Similar to our work, Zhao~\emph{et~al.}~\cite{zhao2020makes} pretrain on multiple datasets. They investigate what information is retained under the transfer learning setup, which differs from the focus of this paper. Purushwalkam and Gupta~\cite{purushwalkam2020demystifying} study the influence of the object-centric dataset bias, but their experimental scope is rather limited, and their conclusions diverge from the ones in this work. Ericsson~\emph{et~al.}~\cite{ericsson2020well} compare several ImageNet pretrained models under the transfer learning setup. In conclusion, we believe our study can complement these works.

\section{Application: Indoor Scene Understanding}
\label{sec: ssl_application}
In this section, we apply self-supervised pretraining to improve a model for indoor scene understanding. We consider the semantic segmentation and depth estimation tasks on the NYUD dataset~\cite{silberman2012indoor}. The model consists of a ResNet-50 backbone with ASPP head. Table~\ref{tab: ssl_application} shows the results obtained via single-task learning and a multi-task learning baseline. In both cases, we obtain better results when using MoCo pretrained weights versus supervised pretraining. We conclude that the gains obtained with contrastive self-supervised learning translate to multi-task learning models as well. 

\begin{table}
    \tablestyle{2pt}{1.1}
    \caption{Comparison of pretraining strategies under a single- and multi-task learning setup on NYUD. We consider the tasks of semantic segmentation and monocular depth estimation. The multi-task learning model with MoCo pretrained weights reports the best results on both tasks.}
    \label{tab: ssl_application}
    \centering
    \begin{tabular}{y{80}|y{80}|x{40}x{40}}
    \toprule
    \fontsize{7.5pt}{1em}\selectfont \textbf{Setup} &
    \fontsize{7.5pt}{1em}\selectfont \textbf{Pretraining (ImageNet)} & 
    \fontsize{7.5pt}{1em}\selectfont \textbf{Sem. Seg.} (IoU) &  \fontsize{7.5pt}{1em}\selectfont \textbf{Depth} (rmse) \\
    \hline
    \multirow{2}{*}{Single-Task Learning} & Supervised & 43.6 & 0.599 \\
    & MoCo & 44.7 & 0.590 \\
    \hline
    \multirow{2}{*}{Multi-Task Learning} & Supervised & 43.7 & 0.593 \\
    & MoCo & \textbf{45.2} & \textbf{0.581} \\
    \bottomrule
    \end{tabular}
\end{table}

\section{Conclusion}
\label{sec: ssl_conclusion}
In this chapter, we showed that we can find a generic set of augmentations/invariances that allows us to learn effective representations across different types of datasets (i.e., scene-centric, non-uniform, domain-specific). We provide empirical evidence to support this claim. First, in Section~\ref{sec: ssl_contrastive_wild}, we show that the standard SimCLR augmentations can be applied across several datasets. Then, Section~\ref{sec: ssl_invariances} studies the use of additional invariances to improve the results for a generic dataset (i.e., MS-COCO). In this way, we reduce the need for dataset or domain-specific expertise to learn useful representations in a self-supervised way. Instead, the results show that simple contrastive frameworks apply to a wide range of datasets. We believe this is an encouraging result. Finally, our overall conclusion differs from a few recent works~\cite{purushwalkam2020demystifying,selvaraju2020casting}, which investigated the use of a more advanced pretext task or video to learn visual representations.

Our results also yields a few interesting follow-up questions. \textbf{Modalities.} Can we reach similar conclusions for other modalities like video, text, audio, etc.? \textbf{Invariances.} What other invariances can be applied? How do we bias the representations to focus more on texture, shape or other specific properties? \textbf{Compositionality.} Can we combine different datasets to learn better representations?

\textbf{Limitations:} We performed extensive experiments to measure the influence of different dataset biases on the representations. The experimental setup covered the use of object-centric vs. scene-centric data, uniform vs. long-tailed class distributions and general vs. domain-specific data. Further, we explored various ways of imposing additional invariances to improve the representations. Undoubtedly, there are several components that fall outside the scope of our study. We briefly discuss some of the limitations below. 

\paragraph{Dataset size.} The experiments were conducted on datasets of moderate size (e.g. MS-COCO contains 118K images). It remains an open question whether the same results will be observed for larger datasets. We believe it would be useful to further investigate the behavior of existing methods on large-scale datasets.

\paragraph{Data augmentations.} Imposing invariances to different data transformations proves crucial to learn useful representations. In this work, we used the same data augmentation strategy for all our experiments. However, different types of data could benefit from a different set of augmentations. We hypothesize that this is particularly true for domain-specific datasets. In this case, we expect that specialized data transformations, based upon domain-knowledge, could further boost the results. Further, we observe that the performance of existing methods still strongly depends on handcrafted augmentations, which can limit their applicability. We tried to partially alleviate this problem by adopting nearest neighbors - which only relies on the data manifold itself. Still, this problem could benefit from further investigation. 

\paragraph{Inductive biases.} All experiments were performed with a ResNet architecture. It would be interesting to study how architectural design choices influence the representation quality. In particular, one could study more general models like transformers~\cite{dosovitskiy2020image,vaswani2017attention} which incorporate different biases. 

\cleardoublepage%

  \graphicspath{{\chaptersdir/conclusion/image/}}%
\chapter{Conclusion}
\label{ch: conclusion}
In this manuscript, we have proposed several techniques for improving deep multi-task learning models in computer vision. This chapter gives an overview of the contributions of each chapter. Further, we also address the limitations of our work and discuss possible directions for future research. 

\section{Summary of Contributions}
In Chapter~\ref{ch: related_work}, we reviewed recent methods for MTL within the scope of deep neural networks. We presented an extensive overview of both architectural and optimization based strategies for MTL. For each method, we described its key aspects, discussed the commonalities and differences with related works, and presented the possible advantages or disadvantages. Further, we introduced a novel taxonomy for describing deep multi-task learning architectures. The proposed taxonomy discriminates between encoder- and decoder-focused models based on where the task interactions are modeled. Encoder-focused models only share information during the encoding stage, while the decoding process happens in an isolated task-specific manner. Differently, decoder-focused models first make initial task predictions, and then leverage features from these initial predictions to further improve each task output - in a one-off or recursive manner. Finally, starting from our literature review, we motivated the various lenses through which we studied the multi-task learning problem in this manuscript.

In Chapter~\ref{ch: branched_mtl}, we considered the use of branched multi-task networks in which deeper layers gradually grow more task-specific. Additionally, we introduced a principled method to automate the construction of such branched multi-task networks. The construction process groups related tasks together while making a trade-off between task similarity and network complexity. The task relatedness scores are based on the premise that similar tasks can be solved with a similar set of features. The proposed approach was evaluated on three different benchmarks including CelebA, Cityscapes and Taskonomy. The results show that, for a given parameter or FLOPS budget, our method consistently yields models with the highest performance. 

Chapter~\ref{ch: mti_net} continued on improving the architecture. This time, we focused on building better decoder-focused models for tackling dense prediction tasks. First, we argued about the importance of modeling task interactions at multiple scales when distilling the task information under a multi-task learning setup. Guided by this principle, we then proposed a novel architecture named MTI-Net. Experiments on two multi-task dense labeling datasets, i.e. NYUD and PASCAL, showed that our model delivers on the full potential of multi-task learning, that is, smaller memory footprint, reduced number of calculations, and better performance w.r.t single-task learning. 

Chapter~\ref{ch: optimization} studied the problem of balancing the learning of tasks under a MTL setup. In particular, we need to avoid that one or more tasks dominate the weight update, thereby hindering the model from learning the other tasks. We analyzed multiple existing task balancing strategies and talked about their commonalities and differences. Additionally, we carried out an ablation study to isolate the elements that are most effective. Based upon our results, we proposed a set of heuristics to help practitioners determine the task-specific weights. Applying the heuristics results in more robust performance across different datasets. This differs from existing methods, which seem tuned for specific scenarios. 

Chapter~\ref{ch: ssl} considered the use of self-supervised pretraining to improve models for scene understanding. First, we showed that an existing technique like MoCo can be used for pretraining across a variety of datasets including scene-centric, long-tailed and domain-specific data. Second, we showed that the representations can be improved by imposing additional invariances. This result directly benefits many downstream tasks like semantic segmentation, detection, etc. Finally, we showed that the improvements obtained with self-supervised pretraining also translate to multi-task learning models. 

We can also draw more general conclusions from this work. First, it requires some effort to get the full benefits from multi-task learning: reduced memory footprint, fewer computations and improved performance. In particular, the improvements obtained with a simple multi-task learning baseline seem limited to a few specific, isolated cases like multi-attribute learning, or specific combinations of tasks like semantic segmentation and depth. It pays off to use more specialized models like MTI-Net when it comes to tackling multiple dense prediction tasks. Second, balancing the learning of tasks still requires a significant amount of trial-and-error. Many existing methods were designed for specific cases, e.g. uncertainty weighting works well with noisy data. We have addressed this problem by proposing a set of heuristics, which can be useful for practitioners working on applications. Third, contrastive self-supervised learning emerges as a powerful technique for improving scene understanding models. We believe such pretraining techniques will quickly get adopted, as they have the potential to benefit applications were annotated data is scarce, like medical imaging, or where large amounts of unlabeled data are readily available, like autonomous driving. 

\section{Limitations and Future Work}
Undoubtedly, there are several aspects of the problem that fall outside the scope of this manuscript. The limitations tied to the proposed methods were already discussed at the end of every chapter. Therefore, we focus on some of the more general limitations below. 

First, the software systems that implement state-of-the-art algorithms for tackling tasks like object detection, semantic segmentation, depth estimation, etc. are well established. Examples include Detectron~\cite{wu2019detectron2}, TorchVision~\cite{paszke2019pytorch}, MMLab~\cite{mmdetection}, etc. The availability of these frameworks allows rapid progress and a thorough evaluation of novel research contributions. Unfortunately, at this time, there are no such software frameworks for training multi-task learning models. As a result, different groups of researchers have implemented their own dataloaders, evaluation criteria, baselines, etc., which makes it hard to draw apples-to-apples comparisons between works. We partially addressed this problem in our survey paper~\cite{vandenhende2021multi}. Still, we believe the community could benefit from the development of a software framework with more explicit support for multi-task learning. This would not only allow to benchmark progress in the field more easily, but also be useful for engineers that are building applications. 

Second, very few works have considered MTL under the semi-supervised or active learning setting. Nonetheless, we believe that these are interesting directions for future research. For example, a major limitation of the fully-supervised MTL setting that we considered is the requirement for all samples to be annotated for every task. Although a few works~\cite{kim2018disjoint,nekrasov2019real} have already touched upon this problem, we believe that this is an interesting direction for future work.

Third, in this work we have focused on tackling multiple visual recognition tasks using images. It would be interesting to also include other modalities like text, video, audio, etc. A few works~\cite{lu202012,hu2021unit} have already shown promising results for such multi-modal learning problems. Further research along this direction could help to build more general purpose intelligence agents capable of handling larger task dictionaries across multiple domains. 

Finally, multi-task learning was recently shown to improve robustness. For example, in~\cite{maomultitask} a multi-task learning strategy showed robustness against adversarial attacks, while~\cite{zamir2020robust} found that applying cross-task consistency in MTL improves generalization, and allows for domain shift detection. Similarly, the performance profiles in Chapter~\ref{ch: optimization} showed that MTL algorithms are more robust against changes in the hyperparameter settings than others. It could be useful to further investigate this behavior. 

\cleardoublepage%

\backmatter

\bibliographystyle{acm}
\bibliography{allpapers}

\includecv{curriculum}

\includepublications{publications}

\makebackcoverXII

\end{document}